\documentclass{article}

\usepackage{iclr2027_conference,times}
\iclrfinalcopy
\usepackage{tikz}
\usepackage{xcolor}
\usetikzlibrary{arrows.meta,calc,positioning}
\usepackage[T1]{fontenc}
\usepackage[utf8]{inputenc}
\usepackage{amsmath,amssymb,mathtools,amsthm}
\usepackage{microtype}
\usepackage{booktabs,longtable,array,calc}
\usepackage{graphicx}
\usepackage{xcolor}
\usepackage{algorithm}
\usepackage{algpseudocode}
\usepackage{hyperref}
\usepackage{xurl}
\usepackage{cleveref}
\hypersetup{colorlinks=true,linkcolor=black,citecolor=blue!55!black,urlcolor=blue!55!black}
\newtheorem{theorem}{Theorem}
\newtheorem{proposition}{Proposition}
\newtheorem{corollary}{Corollary}
\newtheorem{assumption}{Assumption}
\newtheorem{lemma}{Lemma}
\theoremstyle{remark}
\newtheorem{remark}{Remark}

\newcommand{\E}{\mathbb E}

\newcommand{\1}{\mathbf 1}
\newcommand{\epsbar}{\bar\varepsilon}

\title{SGD Selects Initialization-Dependent\\
Representation in Overparameterized\\
ReLU Networks}
\author{%
\textbf{Nikos Tsikouras}\textsuperscript{\textmd{1,2}}\quad
\textbf{Yorgos Pantis}\textsuperscript{\textmd{1,2}}\quad
\textbf{Ahmed Mehdi Inane}\textsuperscript{\textmd{3,4}}\\
\textbf{Mehrab Hamidi}\textsuperscript{\textmd{3,4}}\quad
\textbf{Ioannis Mitliagkas}\textsuperscript{\textmd{2,3,4}}\\[1ex]
\normalfont\small\textsuperscript{1} National and Kapodistrian University of Athens, Greece\\
\normalfont\small\textsuperscript{2} Archimedes, Athena Research Center, Greece\\
\normalfont\small\textsuperscript{3} Universit\'e de Montr\'eal, Canada\\
\normalfont\small\textsuperscript{4} Mila - Quebec Artificial Intelligence Institute, Canada
}

\begin{document}
\maketitle
\lhead{Preprint}

% Unnumbered first-page code footnote, following the derandomization paper.
\begingroup
\renewcommand{\thefootnote}{}
\footnotetext{Code available at \href{https://github.com/NikosTsikouras/sgd-initialization-selection}{https://github.com/TPIHM26/JointSGD}.}
\endgroup

\begin{abstract}
    Overparameterized neural networks can represent the same function through many different hidden configurations, raising the question of which representation is favored by a given training algorithm. We study how stochastic gradient descent (SGD), jointly training both layers, resolves this ambiguity in finite-width, bias-free, two-layer ReLU networks trained to learn a linear teacher under Gaussian inputs. At width two, exact fitting leaves no choice: the neurons must align with the teacher directions. At every width $m \ge 4$, however, infinitely many non-aligned exact fits are possible. Despite this ambiguity, we prove that joint batch-size-one SGD from sufficiently small initialization drives every neuron arbitrarily close to one of the two teacher directions while reaching arbitrarily small population error in finite time. This occurs without explicit regularization, layerwise training, gradient clipping, or a learning-rate schedule. The dynamics additionally select an initialization-dependent allocation of the fitted function across neurons, providing a finer characterization of the representation selected by SGD. Numerical experiments reproduce the predicted alignment, fitting, and initialization-dependent allocation. Our results identify sufficient conditions under which standard SGD exhibits a strong implicit bias toward a highly structured internal representation, even when overparameterization creates infinitely many qualitatively different exact representations.
\end{abstract}

\section{Introduction}
 
Overparameterized neural networks (NNs) admit many solutions that fit the data, yet gradient-based training often exhibits a strong preference for particular solutions and representations \citep{brutzkus2017sgd, gunasekar2018implicit, chizat2020implicit}. Existing theoretical frameworks explain this phenomenon by introducing simplifying assumptions, such as explicit regularization, modified first-order optimization updates, or infinite-width limits \citep{jacot2018neural,mei2018mean,chizat2019lazy,rotskoff2022trainability}. While these assumptions make the evolution of learned representations tractable to analyze, standard SGD dynamics can produce similarly structured representations even in the absence of such constraints.

In this work, we study the representations of SGD that are favored without the aforementioned assumptions.  We take a
finite, overparameterized two-layer ReLU network and train both layers at the
same time with plain stochastic gradient descent (SGD) without regularization. We attempt to answer the following question
\begin{center}
    \emph{When many representations fit perfectly, which one does SGD prefer?}
\end{center}
To isolate this representation-selection question, we deliberately choose a setting in which the statistical problem is simple but the optimization dynamics are not. Following the teacher--student paradigm used to study neural-network training dynamics \citep{goldt2019dynamics, akiyama2021learnability, ba2022high, damian2022neural}, we consider Gaussian inputs and a linear teacher
$y(x)=\beta\theta_\star^\top x$ learned by an overparameterized two-layer ReLU student network
\begin{equation}\label{eq:first}
  f(x)=\frac1{\sqrt m}\sum_{j=1}^m a_j\,\sigma(w_j^\top x),
  \qquad \sigma(t)=\mathrm{ReLU}(t).
\end{equation}
Least squares could recover the target directly; our question is which of the network's many hidden representations the training dynamics select. The difficulty lies in the coupled nonlinear dynamics created by training both layers simultaneously. Each hidden weight \(w_j\) determines which inputs activate its ReLU through the gate \(\mathbf 1\{w_j^\top x>0\}\), while its output weight \(a_j\) controls how strongly that neuron influences the loss and hence the update of \(w_j\). As the hidden weights move, these activation regions change as well, so the geometry of the updates evolves along the optimization trajectory itself. This difficulty is reflected in recent feature-learning analyses, which typically use structured proof strategies such as staged training, different gradient algorithms, or explicit regularization \citep{mousavi2023lowdim, tsikouras2026derandomization,zhang2026neural}, or establish alignment during an early phase before the later interactions are studied
\citep{maennel2018quantizes,min2024alignment,boursier2025simplicity, boursier2025twoedged}.

The closest result to our setting is the recent work of \citep{tinaz2026both}, who analyze the same Gaussian linear teacher under joint training of both layers, but only for the minimal-width case \(m = 2\). They leave open the overparameterized regime, where exact representations are no longer unique. Here overparameterization refers to width beyond the two neurons needed to represent the teacher. Moving to this regime changes the question itself, as convergence to zero loss no longer determines which representation the dynamics will select.

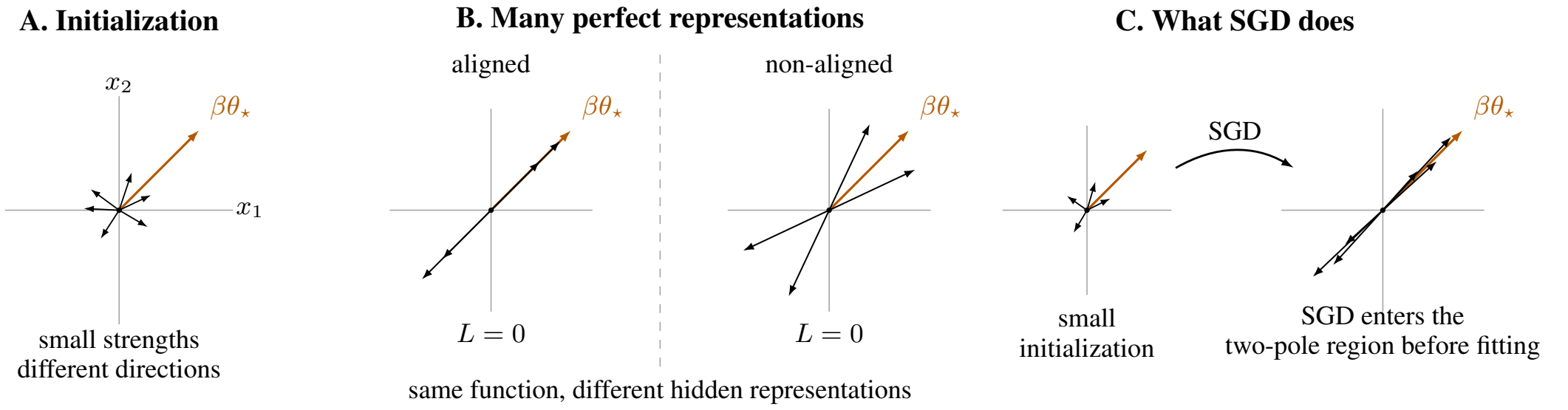
\begin{figure}[!hbt]
    \centering
    \resizebox{\linewidth}{!}{%
        \begin{tikzpicture}[
    x=1cm,y=1cm,
    neuron/.style={
        -{Latex[length=1.2mm,width=0.8mm]},
        line width=0.45pt,
        black
    },
    strongneuron/.style={
        -{Latex[length=1.3mm,width=0.9mm]},
        line width=0.5pt,
        black
    },
    teach/.style={
        -{Latex[length=1.5mm,width=1.0mm]},
        line width=0.8pt,
        orange!70!black
    },
    flow/.style={
        -{Latex[length=1.5mm,width=1.0mm]},
        line width=0.7pt,
        black
    },
    axisline/.style={line width=0.4pt, gray!70},
    paneltitle/.style={font=\bfseries},
    lab/.style={font=\small},
    note/.style={font=\small, align=center}
]

% ------------------------------------------------------------
% Helper macro: 2D axes
% ------------------------------------------------------------
\newcommand{\drawaxes}[1]{
    \draw[axisline] (-#1,0) -- (#1,0);
    \draw[axisline] (0,-#1) -- (0,#1);
}

% Teacher direction = diagonal
% (think theta_* = (1,1)/sqrt(2), teacher vector beta theta_*)
\def\tx{0.95}
\def\ty{0.95}

% ============================================================
% PANEL A
% ============================================================
\begin{scope}[shift={(0,0)}]
    \node[paneltitle] at (0,2.25) {A. Initialization};

    \drawaxes{1.35}
    \node[lab] at (1.55,0.00) {$x_1$};
    \node[lab] at (-0.00,1.48) {$x_2$};

    % Teacher vector
    \draw[teach] (0,0) -- (\tx,\ty);
    \node[lab, orange!70!black, above right] at (\tx,\ty) {$\beta\theta_\star$};

    % Small random neurons
    \draw[neuron] (0,0) -- (0.38,0.18);
    \draw[neuron] (0,0) -- (0.15,0.45);
    \draw[neuron] (0,0) -- (-0.34,0.25);
    \draw[neuron] (0,0) -- (-0.22,-0.34);
    \draw[neuron] (0,0) -- (0.34,-0.20);
    \draw[neuron] (0,0) -- (-0.42,0.02);

    \fill (0,0) circle (0.9pt);

    \node[note] at (0,-1.7) {small strengths\\different directions};
\end{scope}

% ============================================================
% PANEL B
% ============================================================
\begin{scope}[shift={(6.4,0)}]
    \node[paneltitle] at (0,2.25) {B. Many perfect representations};

    \draw[dashed, gray!70] (0,-1.85) -- (0,1.9);

    % -------- aligned subpanel --------
    \begin{scope}[shift={(-2.0,0)}]
        \node[lab] at (0,1.7) {aligned};
        \drawaxes{1.2}

        \draw[teach] (0,0) -- (\tx,\ty);
        \node[lab, orange!70!black, above right] at (\tx,\ty) {$\beta\theta_\star$};

        % aligned neurons along +/- teacher direction
        \draw[strongneuron] (0,0) -- (0.82,0.82);
        \draw[strongneuron] (0,0) -- (0.58,0.58);
        \draw[strongneuron] (0,0) -- (-0.82,-0.82);
        \draw[strongneuron] (0,0) -- (-0.58,-0.58);

        \fill (0,0) circle (0.9pt);

        \node[note] at (0,-1.45) {$L=0$};
    \end{scope}

    % -------- non-aligned subpanel --------
    \begin{scope}[shift={(2.0,0)}]
        \node[lab] at (0,1.7) {non-aligned};
        \drawaxes{1.2}

        \draw[teach] (0,0) -- (\tx,\ty);
        \node[lab, orange!70!black, above right] at (\tx,\ty) {$\beta\theta_\star$};

        % four neurons, symmetric around the teacher direction
        % pair 1
        \draw[strongneuron] (0,0) -- (0.48,1.02);
        \draw[strongneuron] (0,0) -- (-0.48,-1.02);

        % pair 2
        \draw[strongneuron] (0,0) -- (1.02,0.48);
        \draw[strongneuron] (0,0) -- (-1.02,-0.48);

        \fill (0,0) circle (0.9pt);

        \node[note] at (0,-1.45) {$L=0$};
    \end{scope}

    \node[note] at (0,-2.15) {same function, different hidden representations};
\end{scope}

% ============================================================
% PANEL C
% ============================================================
\begin{scope}[shift={(13.2,0)}]
    \node[paneltitle] at (0,2.25) {C. What SGD does};

    % left mini-system: initialization
    \begin{scope}[shift={(-1.75,0)}]
        \drawaxes{1.0}
        \draw[teach] (0,0) -- (0.72,0.72);

        \draw[neuron] (0,0) -- (0.28,0.14);
        \draw[neuron] (0,0) -- (0.10,0.34);
        \draw[neuron] (0,0) -- (-0.24,0.17);
        \draw[neuron] (0,0) -- (-0.16,-0.27);

        \fill (0,0) circle (0.9pt);

        \node[note] at (0,-1.45) {small\\initialization};
    \end{scope}

    % middle arrow: smooth curve
    \draw[flow] (-0.7,0.5) to[bend left=30] node[lab, above] {SGD} (0.7,0.5);

    % right mini-system: aligned outcome
    \begin{scope}[shift={(1.75,0)}]
        \drawaxes{1.2}
        \draw[teach] (0,0) -- (\tx,\ty);
        \node[lab, orange!70!black, above right] at (\tx,\ty) {$\beta\theta_\star$};

        \draw[strongneuron] (0,0) -- (0.81,0.87);
        \draw[strongneuron] (0,0) -- (0.64,0.58);
        \draw[strongneuron] (0,0) -- (0.42,0.47);
        \draw[strongneuron] (0,0) -- (-0.83,-0.79);
        \draw[strongneuron] (0,0) -- (-0.59,-0.65);
        \draw[strongneuron] (0,0) -- (-0.45,-0.41);

        \fill (0,0) circle (0.9pt);

        \node[note] at (0,-1.45) {SGD enters the \\ two-pole region before fitting};
    \end{scope}
\end{scope}

\end{tikzpicture}
    }
    \caption{
    Overparameterization creates many exact hidden representations, while small-initialization SGD selects the aligned family.
    (A) At initialization, neurons have small strength and unrelated directions.
    (B) At width two, exact fitting is rigid, but at width $m\ge4$ there are many
    perfect representations, including non-aligned ones.
    (C) From small initialization, SGD drives all neurons toward the two teacher poles
    $\pm \theta_\star$.
    }
    \label{fig:intro}
\end{figure}

\paragraph{Width changes the question.} With exactly two hidden units, the representation of a linear function is rigid:
zero loss forces one neuron to point along $+\theta_\star$ and the other along
$-\theta_\star$, with uniquely determined effective masses, modulo permutation
and the positive-homogeneity rescaling of ReLU (see left side of Panel B in
Figure~\ref{fig:intro}). This is precisely the regime analyzed by
\citep{tinaz2026both}.

At every integer width $m\ge4$, this rigidity disappears.
Additional neurons can form antipodal pairs away from the teacher: their
nonlinear parts cancel, while their linear contributions combine to represent
the same target (see right side of Panel B in
Figure~\ref{fig:intro}). We make this precise by characterizing the complete
nondegenerate zero-loss set (Proposition \ref{prop:zeroset}). It contains a continuum of exact representations
whose neurons are not aligned with the teacher poles. These solutions are not
saddles that SGD must escape. Their residual vanishes on every input, so every gradient is exactly zero and SGD initialized at any one of them never
moves.

From this, we understand that the loss is indifferent between the aligned representation and a
continuum of non-aligned alternatives. This distinguishes our question from
low-dimensional representation-learning results that use explicit weight decay
as part of the training dynamics
\citep{damian2022neural,mousavi2023lowdim,zhang2026neural}; for instance
\citep{mousavi2023lowdim} use regularization to rule out critical points away
from the teacher subspace, and report that removing it can change the
alignment behavior. Here no penalty resolves the ambiguity among exact fits, so
any preference must arise from the trajectory. Our main theorem shows that it does.

\vspace{0.2cm}
\textbf{Informal version of Theorem \ref{thm:main-sgd}.}
Consider the two-layer ReLU network in Equation~\eqref{eq:first}, trained to
learn a linear teacher. For every integer width $m\ge2$, joint SGD with batch
size one, sampling with replacement from a fixed training set, brings every neuron close to either $+\theta_\star$ or $-\theta_\star$, provided the network is
initialized at a sufficiently small scale. At the same time, the network
learns the linear teacher to arbitrarily small population error.
\vspace{0.2cm}

Figure~\ref{fig:intro} summarizes this contrast: panel A shows the tiny, initially unaligned network, panel B shows the non-uniqueness of perfect representations created by overparameterization, and panel C shows the narrow two-pole region selected by SGD.

The surprising part is not that the network can learn a linear teacher. It is
that it does so through a representation that the loss does not force. At width
$m\ge4$, non-aligned exact representations are abundant and stationary, yet
small-initialization SGD drives \emph{every hidden neuron} toward one of only two
directions and preserves this alignment while the network grows to the fitting
scale. The conclusion concerns every neuron, rather than only their span, their
average, or a subset. It is therefore stronger than recovery of the teacher
subspace: the entire hidden layer concentrates neuronwise around the two teacher
poles.

Alignment is only the first level of selection. When a sign class contains more than one neuron, exact alignment still leaves its mass free to be divided among those neurons. Our second theorem shows that SGD remembers enough of the initialization to
choose among them. While the network output is still negligible, the teacher
amplifies a growing component of each neuron depending on both its initial size
and how favorably it points at its selected pole, converting small initial
differences into different shares. Once aligned, same-class neurons grow at
nearly the same multiplicative rate, so those shares stay almost fixed while the
total mass rises. Theorem~\ref{thm:explicit-selection-main} makes this precise:
SGD selects not only the two-pole geometry but an explicit,
initialization-dependent division of labor within it.

\paragraph{Why does SGD align?}
Think of each neuron as a vector pointing in some direction, with an associated strength. At initialization, the neurons point in different directions and can be only weakly aligned, or even nearly orthogonal, to the teacher. Their strengths are tiny, however, so that
$f\approx 0$. This creates an early regime in which the teacher $y$ dominates the dynamics and can organize the representation before the network is strong enough to fit it, a mechanism that also appears in small-initialization analyses of feature learning \citep{min2024alignment,boursier2025twoedged, zhang2026neural}. For a linear teacher under Gaussian inputs, a key identity implies that the teacher term pulls each neuron toward its sign-selected teacher pole, independently of its current angle (Lemma ~\ref{lem:gaussian}). As a result, the part of each neuron pointing toward a teacher pole grows, while the component orthogonal to the teacher changes little, causing the neurons to align while the network is still weak. In the second phase, their strengths grow toward the fitting scale. The main difficulty is showing that this later growth does not undo the alignment: we prove that interactions among neurons near the same pole are stabilizing, while interactions across the two poles remain weak.

\paragraph{Why is the aligned family special?}
The family selected by SGD is not just visually simpler than the non-aligned
alternatives. It is exactly extremal among all perfect
representations: an exact fit uses the least possible total neuron strength precisely when every active neuron sits at its corresponding teacher pole; among parameterizations achieving this minimum strength, the squared parameter norm is minimized precisely when each neuron is balanced, \(|a_j|=\|w_j\|\). The variational result (Proposition ~\ref{prop:variational-characterization}) singles out the aligned family from the
much larger zero-loss set; the dynamics resolve the ambiguity remaining inside
it. More specifically, minimum norm still permits many allocations, from
two-neuron solutions to dense ones. Theorem~\ref{thm:explicit-selection-main} identifies a dense balanced
exact representation from the initialization and bounds the finite-time
SGD endpoint's distance to it.

\paragraph{Summary of our contributions.} Our key contributions can be summarized as follows:
\begin{itemize}
\item \emph{Width creates genuine representation ambiguity.}
  Width two is rigid, whereas every integer width $m\ge4$ admits infinitely many non-aligned exact fits, all of which are stationary under SGD on any dataset.
\item \emph{Joint SGD selects the aligned family.}
  From small initialization, vanilla SGD on one fixed Gaussian dataset aligns every neuron with its associated teacher pole, preserves this alignment through growth, and fits the teacher in \emph{population} at every integer width.
  \item \emph{The selected family is variationally distinguished.} Among exact fits, aligned solutions minimize effective mass, and balanced aligned solutions minimize parameter norm.
\item \emph{Initialization selects within that family.}
  The initial growing modes determine the relative neuron masses, which are set during alignment and remain nearly fixed through the fitting phase. This identifies an explicit dense representation, rather than only its two-pole geometry.
\end{itemize}\label{sec:introduction}

\section{Setup}\label{sec:setup}

\paragraph{Teacher, model, and loss.} Let data be generated from $X\sim\gamma_d=N(0,I_d)$, of dimension $d\ge2$, and let the teacher model be
$y(X)=\beta\theta_\star^\top X$, where $\|\theta_\star\|=1$ and $\beta>0$.
We train both layers of a bias-free ReLU network and evaluate its population square loss:
\[
f_\Psi(x)=\frac1{\sqrt m}\sum_{j=1}^m a_j\sigma(w_j^\top x), \qquad \sigma(t) = \text{ReLU}(t) \qquad \mathcal{L}(\Psi) = \frac{1}{2}\mathbb{E}\left[\left(f_\Psi(X) - y(X)\right)^2\right]
\]
Here $\Psi=((a_j,w_j))_{j=1}^m$ and $m \ge 2$ is an integer. Our guarantees concern population loss; SGD updates use the empirical loss on the fixed dataset.

\paragraph{The training set.}
Fix a failure probability $\delta\in(0,1/2)$ and an accuracy parameter
$\varepsilon$. Before training, we draw once
\[
 \mathcal{D}=\{X_1,\dots,X_M\}, \qquad X_1,\dots,X_M \stackrel{\text{i.i.d.}}\sim \gamma_d
 \qquad
 M=\widetilde O\!\left(\frac{d+\log(1/\delta)}{\varepsilon^{2}}\right),
\]
with labels $y_i=y(X_i)$. Throughout, $\widetilde O$ hides factors polylogarithmic in $1/\varepsilon$, and $a\asymp b$ means $c\le a/b\le C$ for universal constants $c,C>0$.
Write $\ell_i(\Psi)=\tfrac12(f_\Psi(X_i)-y_i)^2$  and
$\widehat{\mathcal L}(\Psi)=M^{-1}\sum_{i\le M}\ell_i(\Psi)$, for the single-example and empirical losses, respectively. No fresh data is ever drawn again.

\paragraph{The algorithm.}
At every iteration $n\ge0$ we draw an index
$I_n\sim\mathrm{Unif}\{1,\dots,M\}$, independently of everything else,
and update \emph{both layers simultaneously} on that single example:
\begin{equation}\label{eq:alg}
a_{j,n+1}=a_{j,n}-\eta\,\nabla_{a_j}\ell_{I_n}(\Psi_n),
\qquad
w_{j,n+1}=w_{j,n}-\eta\,\nabla_{w_j}\ell_{I_n}(\Psi_n).
\end{equation}
Both layers are updated after every sampled example, using a constant learning
rate and no explicit regularization. We use the convention \(\sigma'(0)=0\).

\paragraph{Polar notation and initialization scale.}

Let \(s_j:=\operatorname{sign}(a_j(0))\in\{-1,+1\}\)
denote the initial sign of the output weight of neuron \(j\); this sign selects the corresponding teacher pole \(s_j\theta_\star\). Write \(S_s=\{j:s_j=s\}\) and any neuron
with \(a_j\neq0\) and \(w_j\neq0\), put
\(p_j=|a_j||w_j|\) (effective mass), \(u_j=w_j/|w_j|\) (direction),
\(P_s=m^{-1/2}\sum_{j\in S_s}p_j\) (normalized sign-class mass), and
\(\theta_j=\angle(u_j,s_j\theta_\star)\) (angle to the selected pole).
The parameter \(\varepsilon\) sets the initialization scale and the accuracy
guarantee; it is not a tolerance chosen after training.

\begin{assumption}[Initialization]\label{assumption:initialization}
Let $p_0=c_0\beta\varepsilon^4/\sqrt m$, where the universal constant
$c_0>0$ is fixed in Theorem~\ref{thm:main-sgd}. Fix $s_j\in\{-1,+1\}$ and
$S_s=\{j:s_j=s\}$ with $m/4\le |S_s|\le3m/4$. Initialize
\[
 a_j(0)=s_jb_j^0,\quad w_j(0)=r_j^0u_j^0,\quad
 b_j^0,r_j^0\in[\sqrt{p_0},\tfrac54\sqrt{p_0}],\quad
 \|u_j^0\|=1,\quad s_j\langle u_j^0,\theta_\star\rangle\ge-\tfrac12.
\]
\end{assumption}

The two layer magnitudes may vary independently, so the initialization need not be balanced. The initialization is independent of the dataset and sampled indices. Corollary~\ref{cor:random-init} gives a sampling rule that does not use the teacher direction and satisfies Assumption~\ref{assumption:initialization} with probability at least $1-\gamma$ (for any $\gamma\in(0,1)$) when $d\gtrsim\log(m/\gamma)$ and $m\gtrsim\log(1/\gamma)$.

\section{Joint Ordinary SGD Aligns then Fits}\label{sec:sgd}
We now present our main finite-time guarantee for joint SGD. The statement uses four scales:
\[
L=\log(1/\varepsilon),
\qquad
\epsbar=\varepsilon L,
\qquad
\tau=\frac{\eta\beta}{\sqrt m},
\qquad
D=d+\log\frac{16m}{\delta\varepsilon}.
\]
Here $L$ is the logarithmic training time scale, $\bar\varepsilon$ is the target scale for the angles \(\theta_j\), $\tau$ is the normalized learning rate, and $D$ is an effective dimension parameter.

\begin{theorem}[SGD aligns then fits]
\label{thm:main-sgd}
There are universal constants
\(c_0, C_M,\varepsilon_0,C>0\) such that the following holds.
Let \(0<\varepsilon\le\varepsilon_0\), \(0<\delta<1/2\), and let
Assumption~\ref{assumption:initialization} hold. Draw
\[
M=\Bigl\lceil C_M\bigl(d+\log(16/\delta)\bigr)
 \frac{L^2}{\varepsilon^2}\log^2(eL/\varepsilon)\Bigr\rceil
\]
i.i.d.\ Gaussian examples once, and run the SGD update~\eqref{eq:alg}
with the constant learning rate $\eta=\tau\sqrt m/\beta$, where $\tau\asymp\varepsilon^3/(D^2L^3)$. Let $T_1<T_\star$ be the iteration counts given in Appendix~\ref{app:single-example}; they depend only on $(\varepsilon,\delta,d,m)$ and satisfy $T_\star=O(D^2L^4/\varepsilon^3)$. Then, with probability at least
\(1-\delta\) over the dataset and the sampled indices, every output sign remains fixed at its initial value and every neuron stays nonzero through \(T_\star\), and
\[
\max_{T_1\le n\le T_\star}
\max_{1\le j\le m}
\angle(u_{j,n},s_j\theta_\star)
\le
C\bar\varepsilon,
\qquad
\|f_{\Psi_{T_\star}}-y\|_{L^2(\gamma_d)}
\le
C\beta\varepsilon .
\]
\end{theorem}

Here $T_1$ marks the end of the alignment phase (Phase I) and $T_\star$ the fitting time. The full proof can be found in
Appendices~\ref{app:technical}--~\ref{app:single-example}. 

Directional alignment holds for every neuron at every iteration between $T_1$ and $T_\star$ (Phase II), while the fitting guarantee bounds the true population $L^2(\gamma_d)$ error even though SGD only sees the $M$ training examples. We next sketch the two-phase dynamics underlying Theorem~\ref{thm:main-sgd}.

\subsection{Proof sketch}

The dynamics have two phases: the neurons first turn toward their poles while
the network is still negligible, and then grow to the fitting
scale without ruining alignment. We first describe the population-gradient reference dynamics. The final part of the sketch explains how blocks of single-example SGD satisfy the same estimates with controlled errors.

\paragraph{Phase I: why the neurons align.}
Until the end of Phase~I the network output satisfies $\|f_{\Psi_n}\|_{L^2(\gamma_d)}\le\beta\varepsilon^2$, so the residual is essentially $-y$ and each neuron is driven by the teacher alone. One might expect this force to favor only the neurons that already point the right way, since the ReLU gate depends on $w_j$. It does not: the gate integrates out, because $\mathbb{E}[(\theta_\star^\top X)\mathbf 1\{w^\top X>0\}X]=\tfrac12\theta_\star$ for every $w\neq 0$ (Lemma~\ref{lem:gaussian}). The teacher therefore pulls every hidden weight $w_j$ along its sign selected direction $s_j\theta_\star$, with a strength proportional to $|a_j|$ and independent of the angle that $w_j$ currently makes.

Writing \(A_j=|a_j|\) and \(B_j=s_j\theta_\star^\top w_j,
\) for the second-layer magnitude and the component of $w_j$ toward its selected
pole, the teacher-dominated dynamics take the approximate form 
$$
A_j^+\approx A_j+\frac{\tau}{2}B_j,
\qquad
B_j^+\approx B_j+\frac{\tau}{2}A_j.
$$
Thus the update of each quantity is proportional to the other: a larger
second-layer weight strengthens the pull of $w_j$ toward the teacher, while a
larger teacher-aligned component strengthens the growth of $|a_j|$. The two
therefore reinforce one another. Equivalently, in the coordinates
\[
G_j=\frac{A_j+B_j}{2},
\qquad
H_j=\frac{A_j-B_j}{2},
\]
the dynamics approximately decouple: $G_j$ grows like
$(1+\tau/2)^n$, while $H_j$ decays like $(1-\tau/2)^n$. Thus the growing mode
is amplified while the difference $H_j$ becomes negligible.

At the same time, the component of $w_j$ orthogonal to $\theta_\star$ (nuisance component)
receives no teacher force and stays of the order of its initial size
$\sqrt{p_0}$. Hence the growing teacher-aligned component eventually dominates:
$w_j$ aligns with $s_j\theta_\star$, while $|a_j|$ becomes comparable to
$s_j\theta_\star^\top w_j$.

The growth rate is common to all neurons; what differs between them is the initial growing mode
\[
G_j^0=\frac{|a_j(0)|+s_j\theta_\star^\top w_j(0)}{2}.
\]
By the end of
Phase~I, $p_j = |a_j|\|w_j\|$ is proportional, up to small relative error, to $(G_j^0)^2$, so the same mechanism
that aligns the neurons also determines their relative masses.

\paragraph{Phase II: why growth does not undo alignment.}
At the end of Phase~I, the neurons point close to their sign-selected poles, but the network is still too small to fit the teacher. The class masses satisfy the approximate logistic recursion
\(P_s^+\approx P_s+\frac{\tau}{\beta}P_s(\beta-P_s)
\)
(Lemma~\ref{lem:phase-two-mass}), so both class masses approach \(\beta\) within \(O(L/\tau)\) further steps.

The main difficulty is showing that this growth does not destroy alignment. Three facts control the angular dynamics. First, the net restoring coefficient is governed by the class deficit \(\beta-P_s\). Second, the leading same-class interaction has a favorable sign when the angular dynamics are averaged with the neuron masses. Third, opposite-class neurons are nearly antipodal, so the overlap of their active regions, and hence their interaction, is small.

While $P_s \le \beta/2$, the restoring coefficient is uniformly positive and keeps every neuron within \(O(\varepsilon)\) of its selected pole. After $P_s$ crosses $\beta/2$, that coefficient weakens and may become slightly negative if the class overshoots the target, so the proof no longer relies on contraction. Instead, it shows directly that the accumulated adverse drift and error terms remain small, keeping the mass-weighted angular error \(O(\varepsilon)\). The influence of the class on any individual neuron is controlled by this weighted error. Accumulating the resulting \(O(\varepsilon)\) forcing over \(O(L)\) normalized time gives the worst-neuron bound \(O(\varepsilon L)\). Hence every neuron remains within \(O(\varepsilon\log(1/\varepsilon))\) of its pole throughout Phase~II. At \(T_\star\), \(|P_s-\beta|=O(\beta\varepsilon)\) and the mass-weighted angular error is \(O(\varepsilon)\), which together yield
\(\|f_{\Psi_{T_\star}}-y\|_{L^2(\gamma_d)}
=O(\beta\varepsilon).
\)
\paragraph{Why one example at a time is enough.}
The preceding two phases describe the population dynamics, whereas SGD uses one training example per update. A single stochastic gradient need not approximate the population gradient, so the proof groups consecutive updates into blocks of \(K\) steps. These blocks are used only in the analysis; the algorithm updates both layers after every sampled example. Each block must contain enough steps,
\(
K\asymp \frac{D^2L^2}{\varepsilon^2},
\)
for the sampling fluctuations to average out, while its normalized duration,
\(K\tau\asymp\frac{\varepsilon}{L},
\)
must be small enough that the parameters move little within the block. These requirements give
\(\tau\asymp\frac{\varepsilon^3}{D^2L^3}.
\)
The total update over a block can then be compared with one controlled population-gradient step.

The estimates must hold relative to each neuron's current size: each effective mass is an \(O(\varepsilon^4)\) fraction of its fitting-scale value at initialization and still an \(O(\varepsilon^2)\) fraction at \(T_1\), so an absolute error bound would not protect every neuron. A neuron-relative regularity estimate and uniform concentration of the Gaussian gated covariances provide this control throughout the trajectory, even though the same dataset is reused (Lemmas~\ref{se:lem:relative} and~\ref{se:lem:tinaz-uniform}).

\subsection{What function-level geometry misses}
\label{sec:egop}

A common way to identify the features learned by a predictor is through its
expected gradient outer product (EGOP)
\[
G_f=\E[\nabla f(X)\nabla f(X)^\top].
\]
Its leading eigenspace records the input directions along which the predictor
varies most. This viewpoint underlies active-subspace methods and recent work
on feature learning in NNs
\citep{constantine2015active, beaglehole2024average,radhakrishnan2024agop}.

Theorem~\ref{thm:main-sgd} implies that the input-gradient energy outside
the teacher direction is small. Let
$Q_\star=I-\theta_\star\theta_\star^\top$.

\begin{corollary}[Near-rank-one gradient geometry]
\label{cor:gradient}
Under the assumptions of Theorem~\ref{thm:main-sgd}, on the same event of probability at least \(1-\delta\), and with \(\lambda_1\ge\cdots\ge\lambda_d\) the eigenvalues of \(G_{f}\) for \(f=f_{\Psi_{T_\star}}\),
\[
\E\|Q_\star\nabla f(X)\|^2
\le C\beta^2\varepsilon^2,
\qquad
\sum_{i\ge2}\lambda_i(G_{f})
\le C\beta^2\varepsilon^2.
\]
\end{corollary}
\noindent\emph{Proof.}
See Appendices~\ref{app:sharp-fit} and~\ref{app:single-example}.

This function-level description does not identify the hidden representation.
Every exact fit, including the non-aligned representations constructed in Section \ref{sec:zeroset},
has $G_f=\beta^2\theta_\star\theta_\star^\top$. Neuronwise alignment therefore
reveals a preference of the dynamics that the EGOP cannot distinguish.

\section{Exact zero-set geometry}
\label{sec:zeroset}
\vspace{-0.15cm}
Theorem~\ref{thm:main-sgd} shows that SGD reaches the fitting scale with every
neuron close to a teacher pole. We now ask whether this geometry is forced by
exact fitting. At width two it is; from width four onward, even the hidden lines
are no longer determined by exact fitting.

The starting point is the decomposition
\(
\sigma(t)=\frac{t+|t|}{2}.
\)
Each ReLU neuron therefore contributes a linear part and an even nonlinear
part. Since the teacher \(y(x)=\beta\theta_\star^\top x\) is linear and odd,
an exact representation must reproduce its linear part while canceling all
even contributions.
\begin{proposition}[Exact zero set]
\label{prop:zeroset}
Assume \(a_j\neq0\neq w_j\) for every \(j\), and set
\(s_j=\operatorname{sign}(a_j)\). Then
\(
f_\Psi(x)=\beta\theta_\star^\top x
\quad\text{in }L^2(\gamma_d)
\)
if and only if
\[
\frac1{\sqrt m}\sum_{j=1}^m s_jp_ju_j
=
2\beta\theta_\star
\tag{Z1}
\]
and, for every unit vector \(u\),
\[
\sum_{j:\,u_j\in\{u,-u\}}s_jp_j=0.
\tag{Z2}
\]
\end{proposition}
\noindent\emph{Proof.}
See Appendix~\ref{app:local-geometry}.

The two conditions are the linear and the nonlinear parts. (Z1) says the
weighted directions must sum to the teacher direction with the correct
magnitude; individual neurons may point elsewhere, only their sum is fixed.
(Z2) says the even terms must cancel in every direction; $u$ and $-u$ are
grouped because $|u^\top x|=|(-u)^\top x|$, so among neurons on that line the
positive and negative output contributions must have equal total mass. The
simplest example is an antipodal pair,
$p\,\sigma(u^\top x)-p\,\sigma(-u^\top x)=p\,u^\top x$: the nonlinear parts
cancel and a linear contribution in direction $u$ remains. A linear teacher can therefore be represented by combining several such pairs
whose linear contributions sum to the teacher. Using more than one hidden line requires at least four neurons.

\begin{corollary}[Width dichotomy]
\label{cor:width}
Let \(m=2\). Then \(f_\Psi=y\) if and only if, after relabeling the neurons, \(a_1>0>a_2\), \(u_1=\theta_\star\), \(u_2=-\theta_\star\), and \(p_1=p_2=\sqrt2\,\beta\). In particular both neurons are active, and only the split of each \(p_j\) between \(|a_j|\) and \(\|w_j\|\) is free.

For every integer \(m\ge4\) and every prescribed angle
\(\gamma\in(0,\pi/2)\), there exist parameters with \(f_\Psi=y\), \(|a_j|=\|w_j\|>0\) for every \(j\), \(\max_j\angle(u_j,s_j\theta_\star)=\gamma\), and at least two distinct hidden lines.
\end{corollary}

\noindent\emph{Proof.}
See Appendix~\ref{app:local-geometry}.

Corollary~\ref{cor:width} shows when alignment is forced by fitting and
when it is not. At width two, every exact fit is aligned. From width four
on, exact fits can place every neuron at any angle
$\gamma\in(0,\pi/2)$ from its pole, even with perfectly balanced layers,
so neither zero loss nor balance forces alignment.

At width three, exact fitting still forces the teacher line, but need
not place each neuron at its output-sign-selected pole;
Appendix~\ref{app:local-geometry} gives an example.

The non-aligned exact fits are not states that SGD could later escape:
their residual vanishes on every input, so every single-example gradient
is zero. Thus every exact representation, aligned or not, is a fixed
point of the algorithm~\eqref{eq:alg}, for every dataset and every
sequence of sampled indices. Training on a finite dataset does not
resolve this ambiguity: every exact representation satisfies
$f_\Psi(X_i)=y_i$ for all $i$, and therefore has zero training loss as
well. Nevertheless, Theorem~\ref{thm:main-sgd} shows that, from small
initialization, SGD reaches the fitting scale near the aligned family.
This preference is therefore determined by the training trajectory
rather than by zero loss or stationarity.

All exact representations are also indistinguishable at the level of the
learned function: each represents the same linear teacher and hence, as
noted in Section~\ref{sec:egop}, has
\(G_f=\beta^2\theta_\star\theta_\star^\top\). The EGOP recovers the
teacher direction but cannot reveal whether the hidden neurons themselves
are aligned. Theorem~\ref{thm:main-sgd} therefore identifies a property
of the learned representation that is invisible both from the loss and
from the input-gradient geometry.

The aligned representations are nevertheless distinguished inside the zero set by a simple size criterion. Define the effective mass and the raw
squared parameter norm by
\[
\Pi(\Psi)
=
\frac1{\sqrt m}\sum_{j=1}^m |a_j|\|w_j\|,
\qquad
S(\Psi)
=
\sum_{j=1}^m\bigl(a_j^2+\|w_j\|^2\bigr).
\]

\begin{proposition}[Variational characterization of aligned representations]
\label{prop:variational-characterization}
Among all parameterizations \(\Psi\) that represent the teacher exactly,
the minimum effective mass and minimum raw squared parameter norm are
\vspace{-0.15cm}
\[
% \vspace{-0.15cm}
\min_{f_\Psi=y}\Pi(\Psi)=2\beta,
\qquad
\min_{f_\Psi=y}S(\Psi)=4\beta\sqrt m,
\]
respectively. A representation minimizes \(\Pi\) if and only if every active
neuron lies at its output-sign teacher pole and the two sign classes each have
effective mass \(\beta\). Among these representations, it minimizes \(S\) if
and only if
\(
w_j=a_j\theta_\star \text{ for every }j.
\)
\end{proposition}
\noindent\emph{Proof.}
See Appendix~\ref{app:selection-principle}.

The characterization has a simple interpretation. Any mass placed away from
the two teacher poles is partly wasted: only its projection onto
\(\theta_\star\) contributes to the teacher, while the remaining components
must cancel elsewhere. Alignment removes this excess. For a fixed effective
mass, the raw parameter norm is then smallest when the two layers are
balanced.

Proposition~\ref{prop:variational-characterization} identifies the
aligned minimum-mass family and its balanced minimum-norm members.
It leaves the allocation of mass within each sign class undetermined; the next section shows that the initialization determines it.

\vspace{-0.15cm}
\section{Initialization selects the neuronwise masses}
\label{sec:mass-selection}
\vspace{-0.15cm}
Within the aligned family, exact fitting determines each sign-class mass
but leaves its allocation among neurons free; the initialization predicts
that allocation. Recall the initial growing mode 
\(
G_j^0=\frac{1}{2}\left(|a_j(0)|+s_j\theta_\star^\top w_j(0)\right).
\)
It combines the second-layer weight magnitude with the first-layer component toward
the sign-selected teacher pole. Define the predicted and observed shares by
\[
\lambda_j=\frac{(G_j^0)^2}{\sum_{i\in S_{s_j}}(G_i^0)^2},
\qquad
v_{j,n}=\frac{p_{j,n}}{\sum_{i\in S_{s_j}}p_{i,n}}.
\]
Both sum to one within each sign class. For the fixed teacher, \(\lambda_j\) is determined from the initialization and $\theta_\star$ before training, while $v_{j,n}$ measures the allocation along training.
A larger initial growing mode predicts a larger final share. The corresponding balanced aligned representation is
\[
p_j^{\rm sel}=\beta\sqrt m\,\lambda_j,
\qquad
a_j^{\rm sel}=s_j\sqrt{p_j^{\rm sel}},
\qquad
w_j^{\rm sel}=a_j^{\rm sel}\theta_\star.
\]
We write \(\Psi^{\rm sel}=((a_j^{\rm sel},w_j^{\rm sel}))_{j=1}^m\).

\begin{theorem}[Initialization-dependent mass selection]
\label{thm:explicit-selection-main}
Under the assumptions of Theorem~\ref{thm:main-sgd}, on the same event of probability at least
\(1-\delta\), for every neuron \(j\),
\[
\vspace{-0.15cm}
e^{-C\varepsilon}\lambda_j\le v_{j,n}\le
e^{C\varepsilon}\lambda_j
\quad (T_1\le n\le T_\star),
\qquad
\left|
\frac{p_{j,T_\star}}{\beta\sqrt m\,\lambda_j}-1
\right|\le C\varepsilon,
\]
and
\[
\vspace{-0.15cm}
\frac{\|\Psi_{T_\star}-\Psi^{\rm sel}\|}
     {\|\Psi^{\rm sel}\|}
\le C\varepsilon.
\vspace{-0.15cm}
\]
\end{theorem}

\noindent\emph{Proof.}
See Appendices~\ref{app:selection-principle} and~\ref{app:single-example}.

Thus the initialization predicts both the relative share and the fitting-time
effective mass of every neuron. It also identifies the balanced aligned
representation approached at the fitting time. Phase I establishes these shares up to the stated error, and they change little during Phase II.

\vspace{-0.1cm}
\section{Experiments}
\label{sec:experiments}
% \vspace{-0.0cm}

We test the two predictions of our theory: that alignment precedes
fitting, and that the initialization determines the neuronwise
allocation. In the two experiments below, SGD updates both layers
after each example sampled uniformly with replacement from one
fixed dataset of $M=2000$ Gaussian examples per run, without
regularization. Performance is measured by the Gaussian population
loss, computed in closed form. Details and additional experiments appear in
Appendices~\ref{app:two-phase-experiment}--\ref{app:additional-experiments}.

\begin{figure}[t]
    \centering
    \begin{minipage}[t]{0.5\linewidth}
        \centering
        \includegraphics[width=\linewidth]{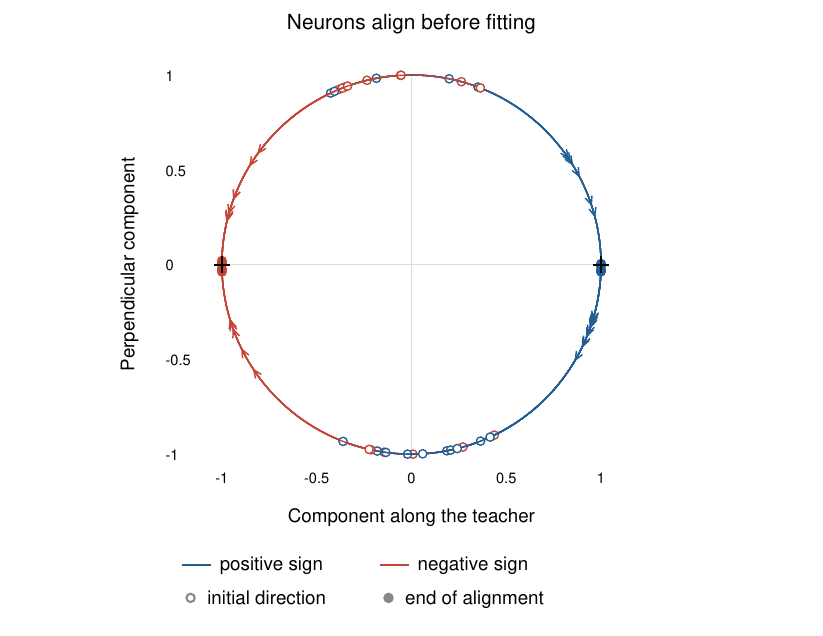}
    \end{minipage}\hfill
    \begin{minipage}[t]{0.5\linewidth}
        \centering
        \includegraphics[width=\linewidth]{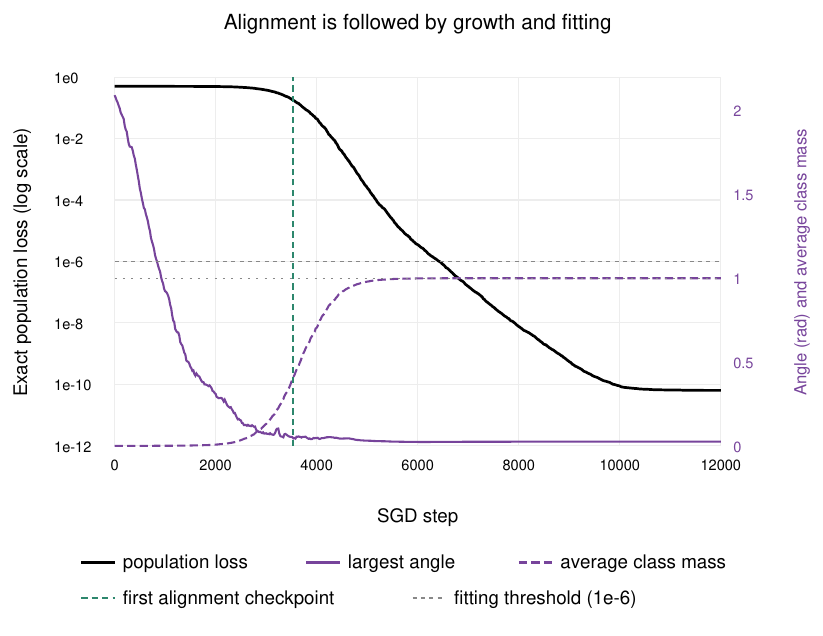}
    \end{minipage}
    \caption{\textbf{The two phases of joint SGD.}
    \textbf{Left:} Normalized hidden directions for a random $3\%$ of
    neurons initialized away from $\pm\theta_\star$. Open circles mark
    initialization, filled circles the first alignment checkpoint, and
    arrows the direction of time. Each neuron moves toward the teacher
    pole selected by the sign of its output weight.
    \textbf{Right:} Population loss (black, logarithmic left axis),
    maximum angle to the selected pole (solid purple), and average
    sign-class mass (dashed purple). After the first alignment checkpoint
    (green), the class mass continues to grow toward its exact-fit value
    and the loss falls below $10^{-8}$. Training uses batch-size-one SGD
    sampled uniformly with replacement from one fixed dataset of
    $M=2000$ Gaussian examples.}
    \label{fig:two-phase}
\end{figure}

Figure~\ref{fig:two-phase} shows the two-phase dynamics in a run
with $d=2$, $m=1000$, and $\beta=1$. The hidden directions first
move toward the teacher pole selected by the sign of their
second-layer weight. Initially, the class masses and population
loss change little. As alignment improves, the class masses grow toward their exact-fit
values and the loss rapidly decreases.

% \vspace{-0.2cm}

The two panels separate directional learning from growth.
The left panel plots $u_{j,n}=w_{j,n}/\|w_{j,n}\|$, so movement
reflects changes in direction rather than magnitude.
In the right panel, the maximum angle first falls below $0.05$
radians at step $3{,}540$, well before fitting. After small
fluctuations, it stays below this threshold at every monitored
iterate from step $4{,}340$ through fitting. The average class mass
grows toward $\beta=1$, and the population loss crosses $10^{-6}$
at step $6{,}440$. This illustrates the ordering used in the proof:
directional alignment precedes fitting.

We next test the neuronwise prediction of
Theorem~\ref{thm:explicit-selection-main}. In dimension $d=20$,
we run $20$ trials at each width $m\in\{4,16,64\}$ and compare
the neuronwise shares at fitting with the shares $\lambda_j$
predicted from initialization. Although all neurons start with
the same effective mass, their fitting-time shares are nonuniform
and closely match the growing-mode prediction: the pooled median
absolute log-share error is at most $0.086$ and the log-share
correlation is at least $0.948$ at every width.
All $20$ runs at each width align before fitting.
Figure~\ref{fig:mass-selection-experiment} compares predicted and
observed shares for all neurons across these runs, illustrating
how the nonuniform allocation reached by SGD is predicted before
training.

\vspace{-0.2cm}
\section{Conclusion and limitations}
\label{sec:discussion}
\vspace{-0.2cm}

Joint SGD from small initialization approaches an aligned exact
representation whose neuronwise masses are predicted by the initialization.
The aligned exact family minimizes effective mass, and its balanced members
minimize raw parameter norm; the dynamics identify a particular member
without explicit regularization. Our experiments illustrate alignment
before fitting and the predicted mass allocation. The guarantee concerns a finite fitting time on a fixed dataset,
not asymptotic convergence. Proposition~\ref{prop:cubic-degeneracy} (Appendix~\ref{app:local-geometry})
exhibits cubic flatness of the population loss near an aligned fit;
it does not determine the convergence rate of the SGD trajectory.
Convergence of the same run, larger initialization scales, sharper
learning-rate bounds, and richer targets remain open.

\vspace{1cm}

\section*{Acknowledgments and Disclosure of Funding}
Nikos Tsikouras and Yorgos Pantis did this work in Archimedes, Athena Research Center supported by a PhD fellowship.
Ioannis Mitliagkas acknowledges support by Archimedes, Athena Research Center, Greece and a Canada
CIFAR AI chair. 

\section*{Reproducibility statement}
The appendices give the assumptions and proofs for alignment,
population fitting, gradient geometry, variational characterization,
and neuronwise mass selection. Appendix \ref{app:single-example} establishes the
fixed-dataset, single-example SGD guarantees, and Appendix \ref{app:local-geometry} proves
the exact zero-set characterization and cubic local obstruction. The experimental appendix specifies the initialization, training
procedure, and evaluation formulas.

\section*{Ethics statement}
This paper is theoretical and studies optimization mechanisms in stylized NN models.
It uses no human-subject data and introduces no direct high-risk application.

\section*{AI use statement}
AI tools were used during the development and preparation of this work to assist with the writing of the text and improving the organization. Additionally, we used AI tools to improve certain math statements.

All mathematical statements and proofs were independently checked by the authors, and cited literature was verified against the original sources. The authors take full responsibility for the correctness of the mathematical results, the accuracy of the citations, the novelty claims, and the final content of the paper.

\bibliographystyle{iclr2027_conference}
\bibliography{references}

\begin{thebibliography}{43}
\providecommand{\natexlab}[1]{#1}
\providecommand{\url}[1]{\texttt{#1}}
\expandafter\ifx\csname urlstyle\endcsname\relax
  \providecommand{\doi}[1]{doi: #1}\else
  \providecommand{\doi}{doi: \begingroup \urlstyle{rm}\Url}\fi

\bibitem[Akiyama \& Suzuki(2021)Akiyama and Suzuki]{akiyama2021learnability}
Shunta Akiyama and Taiji Suzuki.
\newblock On learnability via gradient method for two-layer relu neural networks in teacher-student setting.
\newblock In \emph{International Conference on Machine Learning}, pp.\  152--162. PMLR, 2021.

\bibitem[Ba et~al.(2022)Ba, Erdogdu, Suzuki, Wang, Wu, and Yang]{ba2022high}
Jimmy Ba, Murat~A Erdogdu, Taiji Suzuki, Zhichao Wang, Denny Wu, and Greg Yang.
\newblock High-dimensional asymptotics of feature learning: How one gradient step improves the representation.
\newblock \emph{Advances in Neural Information Processing Systems}, 35:\penalty0 37932--37946, 2022.

\bibitem[Beaglehole et~al.(2024{\natexlab{a}})Beaglehole, Mitliagkas, and Agarwala]{beaglehole2024alignment}
Daniel Beaglehole, Ioannis Mitliagkas, and Atish Agarwala.
\newblock Feature learning as alignment: a structural property of gradient descent in non-linear neural networks.
\newblock \emph{arXiv preprint arXiv:2402.05271}, 2024{\natexlab{a}}.

\bibitem[Beaglehole et~al.(2024{\natexlab{b}})Beaglehole, S{\'u}ken{\'\i}k, Mondelli, and Belkin]{beaglehole2024average}
Daniel Beaglehole, Peter S{\'u}ken{\'\i}k, Marco Mondelli, and Mikhail Belkin.
\newblock Average gradient outer product as a mechanism for deep neural collapse.
\newblock \emph{Advances in Neural Information Processing Systems}, 37:\penalty0 130764--130796, 2024{\natexlab{b}}.

\bibitem[Bietti et~al.(2022)Bietti, Bruna, Sanford, and Song]{bietti2022learning}
Alberto Bietti, Joan Bruna, Clayton Sanford, and Min~Jae Song.
\newblock Learning single-index models with shallow neural networks.
\newblock \emph{Advances in neural information processing systems}, 35:\penalty0 9768--9783, 2022.

\bibitem[Boursier \& Flammarion(2025{\natexlab{a}})Boursier and Flammarion]{boursier2025simplicity}
Etienne Boursier and Nicolas Flammarion.
\newblock Simplicity bias and optimization threshold in two-layer relu networks.
\newblock In \emph{Proceedings of the 42nd International Conference on Machine Learning}, volume 267 of \emph{Proceedings of Machine Learning Research}, pp.\  5241--5275, 2025{\natexlab{a}}.

\bibitem[Boursier \& Flammarion(2025{\natexlab{b}})Boursier and Flammarion]{boursier2025twoedged}
Etienne Boursier and Nicolas Flammarion.
\newblock Early alignment in two-layer networks training is a two-edged sword.
\newblock \emph{Journal of Machine Learning Research}, 26\penalty0 (183):\penalty0 1--75, 2025{\natexlab{b}}.

\bibitem[Boursier et~al.(2022)Boursier, Pillaud-Vivien, and Flammarion]{boursier2022gradient}
Etienne Boursier, Loucas Pillaud-Vivien, and Nicolas Flammarion.
\newblock Gradient flow dynamics of shallow relu networks for square loss and orthogonal inputs.
\newblock \emph{Advances in Neural Information Processing Systems}, 35:\penalty0 20105--20118, 2022.

\bibitem[Brutzkus et~al.(2017)Brutzkus, Globerson, Malach, and Shalev-Shwartz]{brutzkus2017sgd}
Alon Brutzkus, Amir Globerson, Eran Malach, and Shai Shalev-Shwartz.
\newblock Sgd learns over-parameterized networks that provably generalize on linearly separable data.
\newblock \emph{arXiv preprint arXiv:1710.10174}, 2017.

\bibitem[Chistikov et~al.(2023)Chistikov, Englert, and Lazic]{chistikov2023learning}
Dmitry Chistikov, Matthias Englert, and Ranko Lazic.
\newblock Learning a neuron by a shallow relu network: Dynamics and implicit bias for correlated inputs.
\newblock \emph{Advances in Neural Information Processing Systems}, 36:\penalty0 23748--23760, 2023.

\bibitem[Chizat \& Bach(2020)Chizat and Bach]{chizat2020implicit}
Lenaic Chizat and Francis Bach.
\newblock Implicit bias of gradient descent for wide two-layer neural networks trained with the logistic loss.
\newblock In \emph{Conference on learning theory}, pp.\  1305--1338. PMLR, 2020.

\bibitem[Chizat et~al.(2019)Chizat, Oyallon, and Bach]{chizat2019lazy}
Lenaic Chizat, Edouard Oyallon, and Francis Bach.
\newblock On lazy training in differentiable programming.
\newblock \emph{Advances in neural information processing systems}, 32, 2019.

\bibitem[Constantine(2015)]{constantine2015active}
Paul~G Constantine.
\newblock \emph{Active subspaces: Emerging ideas for dimension reduction in parameter studies}.
\newblock SIAM, 2015.

\bibitem[Constantine et~al.(2017)Constantine, Eftekhari, Hokanson, and Ward]{constantine2017ridge}
Paul~G Constantine, Armin Eftekhari, Jeffrey Hokanson, and Rachel~A Ward.
\newblock A near-stationary subspace for ridge approximation.
\newblock \emph{Computer Methods in Applied Mechanics and Engineering}, 326:\penalty0 402--421, 2017.

\bibitem[Damian et~al.(2022)Damian, Lee, and Soltanolkotabi]{damian2022neural}
Alexandru Damian, Jason Lee, and Mahdi Soltanolkotabi.
\newblock Neural networks can learn representations with gradient descent.
\newblock In \emph{Conference on learning theory}, pp.\  5413--5452. PMLR, 2022.

\bibitem[Du et~al.(2018)Du, Hu, and Lee]{du2018balanced}
Simon~S Du, Wei Hu, and Jason~D Lee.
\newblock Algorithmic regularization in learning deep homogeneous models: Layers are automatically balanced.
\newblock volume~31, 2018.

\bibitem[Glasgow(2024)]{glasgow2024finds}
Margalit Glasgow.
\newblock Sgd finds then tunes features in two-layer neural networks with near-optimal sample complexity: A case study in the xor problem.
\newblock In \emph{International Conference on Learning Representations}, volume 2024, pp.\  52419--52430, 2024.

\bibitem[Goldt et~al.(2019)Goldt, Advani, Saxe, Krzakala, and Zdeborov{\'a}]{goldt2019dynamics}
Sebastian Goldt, Madhu Advani, Andrew~M Saxe, Florent Krzakala, and Lenka Zdeborov{\'a}.
\newblock Dynamics of stochastic gradient descent for two-layer neural networks in the teacher-student setup.
\newblock \emph{Advances in neural information processing systems}, 32, 2019.

\bibitem[Gunasekar et~al.(2018)Gunasekar, Lee, Soudry, and Srebro]{gunasekar2018implicit}
Suriya Gunasekar, Jason~D Lee, Daniel Soudry, and Nati Srebro.
\newblock Implicit bias of gradient descent on linear convolutional networks.
\newblock \emph{Advances in neural information processing systems}, 31, 2018.

\bibitem[Jacot et~al.(2018)Jacot, Gabriel, and Hongler]{jacot2018neural}
Arthur Jacot, Franck Gabriel, and Cl{\'e}ment Hongler.
\newblock Neural tangent kernel: Convergence and generalization in neural networks.
\newblock \emph{Advances in neural information processing systems}, 31, 2018.

\bibitem[Kumar \& Haupt(2024)Kumar and Haupt]{kumar2024directional}
Akshay Kumar and Jarvis Haupt.
\newblock Directional convergence near small initializations and saddles in two-homogeneous neural networks.
\newblock \emph{arXiv preprint arXiv:2402.09226}, 2024.

\bibitem[Kumar \& Haupt(2025)Kumar and Haupt]{kumar2025towards}
Akshay Kumar and Jarvis Haupt.
\newblock Towards understanding gradient flow dynamics of homogeneous neural networks beyond the origin.
\newblock \emph{Journal of Machine Learning Research}, 26\penalty0 (269):\penalty0 1--83, 2025.

\bibitem[Liu et~al.(2022)Liu, Zhu, and Belkin]{liu2022landscapes}
Chaoyue Liu, Libin Zhu, and Mikhail Belkin.
\newblock Loss landscapes and optimization in over-parameterized non-linear systems and neural networks.
\newblock \emph{Applied and Computational Harmonic Analysis}, 59:\penalty0 85--116, 2022.

\bibitem[Maennel et~al.(2018)Maennel, Bousquet, and Gelly]{maennel2018quantizes}
Hartmut Maennel, Olivier Bousquet, and Sylvain Gelly.
\newblock Gradient descent quantizes relu network features.
\newblock \emph{arXiv preprint arXiv:1803.08367}, 2018.

\bibitem[Mei et~al.(2018)Mei, Montanari, and Nguyen]{mei2018mean}
Song Mei, Andrea Montanari, and Phan-Minh Nguyen.
\newblock A mean field view of the landscape of two-layer neural networks.
\newblock \emph{Proceedings of the National Academy of Sciences}, 115\penalty0 (33):\penalty0 E7665--E7671, 2018.

\bibitem[Min et~al.(2024)Min, Mallada, and Vidal]{min2024alignment}
Hancheng Min, Enrique Mallada, and Rene Vidal.
\newblock Early neuron alignment in two-layer relu networks with small initialization.
\newblock In \emph{International Conference on Learning Representations}, volume 2024, pp.\  3431--3473, 2024.

\bibitem[Mousavi-Hosseini et~al.(2022)Mousavi-Hosseini, Park, Girotti, Mitliagkas, and Erdogdu]{mousavi2023lowdim}
Alireza Mousavi-Hosseini, Sejun Park, Manuela Girotti, Ioannis Mitliagkas, and Murat~A Erdogdu.
\newblock Neural networks efficiently learn low-dimensional representations with sgd.
\newblock \emph{arXiv preprint arXiv:2209.14863}, 2022.

\bibitem[Ongie et~al.(2019)Ongie, Willett, Soudry, and Srebro]{ongie2020function}
Greg Ongie, Rebecca Willett, Daniel Soudry, and Nathan Srebro.
\newblock A function space view of bounded norm infinite width relu nets: The multivariate case.
\newblock 2019.

\bibitem[Phuong \& Lampert(2020)Phuong and Lampert]{phuong2020inductive}
Mary Phuong and Christoph~H Lampert.
\newblock The inductive bias of relu networks on orthogonally separable data.
\newblock In \emph{International Conference on Learning Representations}, 2020.

\bibitem[Radhakrishnan et~al.(2024)Radhakrishnan, Beaglehole, Pandit, and Belkin]{radhakrishnan2024agop}
Adityanarayanan Radhakrishnan, Daniel Beaglehole, Parthe Pandit, and Mikhail Belkin.
\newblock Mechanism for feature learning in neural networks and backpropagation-free machine learning models.
\newblock \emph{Science}, 383\penalty0 (6690):\penalty0 1461--1467, 2024.

\bibitem[Radhakrishnan et~al.(2025)Radhakrishnan, Belkin, and Drusvyatskiy]{radhakrishnan2025linear}
Adityanarayanan Radhakrishnan, Mikhail Belkin, and Dmitriy Drusvyatskiy.
\newblock Linear recursive feature machines provably recover low-rank matrices.
\newblock \emph{Proceedings of the National Academy of Sciences}, 122\penalty0 (13):\penalty0 e2411325122, 2025.

\bibitem[Ronen et~al.(2019)Ronen, Jacobs, Kasten, and Kritchman]{ronen2019convergence}
Basri Ronen, David Jacobs, Yoni Kasten, and Shira Kritchman.
\newblock The convergence rate of neural networks for learned functions of different frequencies.
\newblock \emph{Advances in Neural Information Processing Systems}, 32, 2019.

\bibitem[Rotskoff \& Vanden-Eijnden(2022)Rotskoff and Vanden-Eijnden]{rotskoff2022trainability}
Grant Rotskoff and Eric Vanden-Eijnden.
\newblock Trainability and accuracy of artificial neural networks: An interacting particle system approach.
\newblock \emph{Communications on Pure and Applied Mathematics}, 75\penalty0 (9):\penalty0 1889--1935, 2022.

\bibitem[Savarese et~al.(2019)Savarese, Evron, Soudry, and Srebro]{savarese2019function}
Pedro Savarese, Itay Evron, Daniel Soudry, and Nathan Srebro.
\newblock How do infinite width bounded norm networks look in function space?
\newblock In \emph{Conference on Learning Theory}, pp.\  2667--2690. PMLR, 2019.

\bibitem[Timor et~al.(2023)Timor, Vardi, and Shamir]{timor2023rank}
Nadav Timor, Gal Vardi, and Ohad Shamir.
\newblock Implicit regularization towards rank minimization in relu networks.
\newblock In \emph{International Conference on Algorithmic Learning Theory}, pp.\  1429--1459. PMLR, 2023.

\bibitem[Tinaz et~al.(2026)Tinaz, Xie, and Soltanolkotabi]{tinaz2026both}
Berk Tinaz, Changzhi Xie, and Mahdi Soltanolkotabi.
\newblock When both layers learn: Training dynamics of representing linear models via relu networks.
\newblock \emph{arXiv preprint arXiv:2606.04476}, 2026.

\bibitem[Tsikouras et~al.(2026)Tsikouras, Pantis, Mitliagkas, and Tzamos]{tsikouras2026derandomization}
Nikos Tsikouras, Yorgos Pantis, Ioannis Mitliagkas, and Christos Tzamos.
\newblock A derandomization framework for structure discovery: Applications in neural networks and beyond.
\newblock In \emph{Proceedings of the Fourteenth Annual International Conference on Learning Representations}, 2026.

\bibitem[Xu \& Du(2023)Xu and Du]{xu2023over}
Weihang Xu and Simon Du.
\newblock Over-parameterization exponentially slows down gradient descent for learning a single neuron.
\newblock In \emph{The Thirty Sixth Annual Conference on Learning Theory}, pp.\  1155--1198. PMLR, 2023.

\bibitem[Zahm et~al.(2020)Zahm, Constantine, Prieur, and Marzouk]{zahm2020dimension}
Olivier Zahm, Paul~G Constantine, Cl{\'e}mentine Prieur, and Youssef~M Marzouk.
\newblock Gradient-based dimension reduction of multivariate vector-valued functions.
\newblock \emph{SIAM Journal on Scientific Computing}, 42\penalty0 (1):\penalty0 A534--A558, 2020.

\bibitem[Zhang et~al.(2026)Zhang, Wang, Fu, and Lee]{zhang2026neural}
Bohan Zhang, Zihao Wang, Hengyu Fu, and Jason Lee.
\newblock Neural networks learn generic multi-index models near information-theoretic limit.
\newblock In \emph{International Conference on Learning Representations}, volume 2026, pp.\  70833--70895, 2026.

\bibitem[Zhang et~al.(2024)Zhang, Saxe, and Latham]{zhang2024bias}
Yedi Zhang, Andrew Saxe, and Peter~E Latham.
\newblock When are bias-free relu networks effectively linear networks?
\newblock \emph{arXiv preprint arXiv:2406.12615}, 2024.

\bibitem[Zhu et~al.(2026)Zhu, Davis, Drusvyatskiy, and Fazel]{zhu2026agop}
Libin Zhu, Damek Davis, Dmitriy Drusvyatskiy, and Maryam Fazel.
\newblock Average gradient outer product in kernel regression provably recovers the central subspace for multi-index models.
\newblock \emph{arXiv preprint arXiv:2605.15082}, 2026.

\bibitem[Zhu et~al.(2025)Zhu, Liu, and Cevher]{zhu2025balances}
Zhenyu Zhu, Fanghui Liu, and Volkan Cevher.
\newblock How gradient descent balances features: A dynamical analysis for two-layer neural networks.
\newblock In \emph{International Conference on Learning Representations}, volume 2025, pp.\  9342--9385, 2025.

\end{thebibliography}

\appendix
\numberwithin{theorem}{section}
\numberwithin{theorem}{section}
\numberwithin{proposition}{section}
\numberwithin{corollary}{section}
\numberwithin{lemma}{section}
\numberwithin{assumption}{section}
\numberwithin{remark}{section}

\section{Related work}\label{sec:related}

\paragraph{Feature learning and low-dimensional targets.}
A broad literature proves that neural training can recover low-dimensional structure beyond kernel regimes.
~\citep{damian2022neural} show representation learning for low-dimensional polynomial targets, while
~\citep{ba2022high} identify a rank-one spike aligned with the linear teacher component after one gradient step.
~\citep{bietti2022learning} study single-index learning with a shallow architecture, and
~\citep{mousavi2023lowdim} prove arbitrary-width subspace learning for Gaussian multi-index targets using online SGD with weight decay.
Those results cover substantially richer target classes than our linear-teacher theorem.
Our focus is simultaneous training of both layers without explicit weight decay, and all-neuron representation selection through fitting.

\paragraph{Joint training and alignment dynamics.}
The closest result to our end-to-end dynamics is ~\citep{tinaz2026both}, who study the Gaussian linear teacher with exactly two hidden ReLU units and jointly train both layers by finite-data gradient descent. They prove global linear convergence from moderately small random initialization with order-wise optimal sample complexity, through alignment, growth, and local-refinement phases. Their result is substantially stronger on initialization, asymptotic convergence, and sample efficiency. The width-two restriction, however, removes the within-class interactions and non-identifiable wide zero set that are central here. We instead analyze every integer width $m\ge2$ using single-example
SGD sampled with replacement from one fixed Gaussian dataset, without
an explicit regularizer, at the price of a more structured small
initialization and conservative sample complexity.

\citet{chistikov2023learning} analyze arbitrary-width ReLU networks
trained by gradient flow from small initialization to learn a single
ReLU teacher on correlated training inputs. Their neuronwise analysis
establishes convergence and an implicit bias toward minimum rank.
Our setting instead concerns a linear teacher, fixed-dataset SGD,
and initialization-dependent neuronwise allocation through fitting.

Joint-layer dynamics have also been analyzed for XOR
~\citep{glasgow2024finds}. In a different parameterization,
~\citep{zhu2025balances} study feature grouping under gradient descent with
fixed positive output coefficients; their theorem does not jointly train both
layers. General results for two-homogeneous models show that near-origin
gradient flow drives nonvanishing parameter blocks toward stationary
directions of a neural correlation function ~\citep{kumar2024directional}, and
subsequent work follows related dynamics beyond the origin
~\citep{kumar2025towards}. For orthogonal regression data,
~\citep{boursier2022gradient} prove convergence to zero loss and an implicit
bias toward minimum variation norm. Related alignment results appear for
orthogonally separable classification
~\citep{phuong2020inductive,min2024alignment}. Our result instead treats single-example SGD on a fixed dataset,
preserves all-neuron alignment through fitting, and tracks the
growing-mode magnitudes needed for neuronwise mass selection.
Earlier and complementary perspectives include directional quantization
~\citep{maennel2018quantizes} and the simplifying and obstructive effects of
early alignment ~\citep{boursier2025twoedged,boursier2025simplicity}. The
balancing phenomenon for homogeneous models goes back to
~\citep{du2018balanced}; here homogeneity is used samplewise to show that the
first-order balance drift cancels exactly under SGD.

\paragraph{Expressivity of bias-free ReLU networks.}
The linear teacher is a natural realizable odd target in our model.
\citep{ronen2019convergence} identified restrictions on the odd components
learnable by shallow bias-free ReLU networks on spherical data, and
\citep{zhang2024bias} proved that every odd function represented by a
two-layer bias-free ReLU network is linear. These results concern the
functions representable by the architecture. Our result instead concerns
which hidden representation joint SGD selects among the many
parameterizations of the same linear function.

\paragraph{AGOP/EGOP and active subspaces.}
The gradient second-moment matrix is classical in active-subspace and ridge-approximation theory ~\citep{constantine2015active,constantine2017ridge,zahm2020dimension}; in particular, its spectral tail controls approximation by functions of the leading subspace under suitable input measures. In neural feature learning, 
average gradient outer product (AGOP) / EGOP has emerged as a descriptor and proposed mechanism for learned feature geometry ~\citep{radhakrishnan2024agop,beaglehole2024alignment,beaglehole2024average}, and ~\citep{radhakrishnan2025linear} prove low-rank recovery properties for linear Recursive Feature Machines. Very recently, ~\citep{zhu2026agop} show that the AGOP of a kernel-ridge predictor can recover the central subspace of multi-index polynomial targets even before accurate prediction.

The generic implication from a small EGOP tail to ridge approximation is established theory. Our principal statement concerns hidden-neuron directions; the EGOP bound records a function-level consequence and cannot distinguish aligned from non-aligned exact representations.

\paragraph{Function-space versus parameter-space geometry.}
Function-space descriptions of two-layer ReLU networks and non-uniqueness of parameterizations are classical themes
~\citep{savarese2019function,ongie2020function}.
~\citep{timor2023rank} give positive and negative results on implicit rank minimization in ReLU networks.
Our exact zero-set characterization isolates this distinction in the present linear-teacher model and shows a sharp \(m=2\) versus \(m\ge4\) change.
Overparameterization-induced slow convergence also appears for a single ReLU teacher with fixed output coefficients ~\citep{xu2023over}. For local optimization geometry, ~\citep{liu2022landscapes} develop PL-type conditions for over-parameterized nonlinear systems; our cubic family gives an explicit failure of any state-independent local PL constant around the two-pole solution at width \(m\ge3\).

\section{Warm-up trajectory analysis}\label{app:technical}

We first develop the trajectory analysis in a simplified setting with
exactly balanced sign classes, equal-magnitude initialization, and fresh
batches. This lets us focus on the main dynamics without carrying the
technical complications of a more general initialization and reused data.
The phase analysis itself only requires small neuron-relative gradient
errors and controlled balance drift, as stated after
Lemma~\ref{lem:balance-propagation}. The preceding concentration bounds
provide these controls for fresh batches, while
Appendix~\ref{app:single-example} establishes their blockwise counterparts
for fixed-dataset SGD.

Appendix~\ref{app:robust-init} extends the balanced, even-width,
equal-magnitude initialization used here to
Assumption~\ref{assumption:initialization} used in the actual SGD algorithm, and
Appendix~\ref{app:sharp-fit} sharpens the final error bound to the rate
stated in the main theorem.

\subsection{Setting, notation, standing assumptions}

\subsubsection{Model and dynamics}

\(d\ge2\), \(m\ge2\) even, \(\beta>0\),
\(\theta_\star\in\mathbb S^{d-1}\), \(\gamma_d=\mathcal N(0,I_d)\),
\(X\sim\gamma_d\). Teacher \(y(x)=\beta\theta_\star^\top x\). Parameter
state \(\Psi=(a_j,w_j)_{j\le m}\in\mathbb R^m\times(\mathbb R^d)^m\);
\[f_\Psi(x)=\frac1{\sqrt m}\sum_{j=1}^ma_j\sigma(w_j^\top x),\qquad\sigma(z)=z_+;\qquad\mathcal L(\Psi)=\tfrac12\mathbb E\bigl[(f_\Psi(X)-y(X))^2\bigr].\]
At iteration \(n\ge0\) draw \(X_{n,1},\dots,X_{n,B}\sim\gamma_d\)
i.i.d., independent of the past. With \(R_n:=f_{\Psi_n}-y\),
\[
\begin{aligned}
\widehat g^a_{j,n}
&=\frac1{B\sqrt m}\sum_{b=1}^B
R_n(X_{n,b})\,\sigma(w_{j,n}^\top X_{n,b}),\\
\widehat g^w_{j,n}
&=\frac1{B\sqrt m}\sum_{b=1}^B
R_n(X_{n,b})\,a_{j,n}
\mathbf 1_{\{w_{j,n}^\top X_{n,b}>0\}}X_{n,b}.
\end{aligned}
\]
\[a_{j,n+1}=a_{j,n}-\eta\widehat g^a_{j,n},\qquad w_{j,n+1}=w_{j,n}-\eta\widehat g^w_{j,n}.\tag{1.1}\label{app:1.1}\]

These are the gradients of
\(\widehat{\mathcal L}_n(\Psi)=\frac1{2B}\sum_b(f_\Psi(X_{n,b})-y(X_{n,b}))^2\)
with the convention $\sigma'(0)=0$. Away from gate boundaries they coincide with the classical sample gradients. Conditional on the past, every nonzero fixed hidden weight has zero probability of a gate-boundary event under the fresh Gaussian batch. For reused data, Appendix~\ref{app:single-example} handles strict gates, including zero preactivations, directly.

\(\mathcal F_n:=\sigma(\text{initialization},\ \text{batches }0,\dots,n-1)\),
so \(\Psi_n\) is \(\mathcal F_n\)-measurable and batch \(n\) is
independent of \(\mathcal F_n\). Population gradients
\(g^a_{j,n}:=\mathbb E[\widehat g^a_{j,n}\mid\mathcal F_n]\),
\(g^w_{j,n}:=\mathbb E[\widehat g^w_{j,n}\mid\mathcal F_n]\).

\subsubsection{Initialization}

Signs \(s_j\in\{\pm1\}\), \(S_\pm:=\{j:s_j=\pm1\}\),
\(|S_+|=|S_-|=m/2\);
\[p_0:=c_0\frac{\beta\varepsilon^4}{\sqrt m},\qquad a_{j,0}=s_j\sqrt{p_0},\qquad w_{j,0}=\sqrt{p_0}\,u^0_j,\ \ u^0_j\in\mathbb S^{d-1},\qquad\max_j\bigl|\langle u^0_j,\theta_\star\rangle\bigr|\le\tfrac12.\tag{1.2}\label{app:1.2}\]
So \(|a_{j,0}|=\|w_{j,0}\|=\sqrt{p_0}\): the initialization is exactly
balanced, \(p_{j,0}=p_0\).

\subsubsection{Polar variables}

The original parameters \((a_j,w_j)\) contain two radial magnitudes,
even though ReLU is positively homogeneous. It is therefore useful to
separate each neuron into its sign, its effective magnitude, and its
direction. When \(a_{j,n}\ne0\ne w_{j,n}\): \(b_{j,n}:=|a_{j,n}|\),
\(r_{j,n}:=\|w_{j,n}\|\), \(p_{j,n}:=b_{j,n}r_{j,n}\),
\(u_{j,n}:=w_{j,n}/r_{j,n}\). Identically
\(a_j\sigma(w_j^\top x)=\operatorname{sign}(a_j)\,p_j\,\sigma(u_j^\top x)\),
so under sign stability
\[f_n(x)=\frac1{\sqrt m}\sum_{j=1}^m s_jp_{j,n}\sigma(u_{j,n}^\top x).\tag{1.3}\label{app:1.3}\]
For each sign class, define the total effective mass
\(P_{s,n}:=m^{-1/2}\sum_{j\in S_s}p_{j,n}\) and its unnormalized version
\(D_{s,n}:=\sum_{j\in S_s}p_{j,n}=\sqrt m\,P_{s,n}\). We also write
\(\Pi_n:=P_{+,n}+P_{-,n}=m^{-1/2}\sum_jp_{j,n}\).

The angular error of neuron \(j\) is
\(\theta_{j,n}:=\angle(u_{j,n},s_j\theta_\star)\). Its component orthogonal
to the teacher direction is
\(\xi_{j,n}:=(I-\theta_\star\theta_\star^\top)u_{j,n}\), so that
\(u_{j,n}=s_j\cos\theta_{j,n}\,\theta_\star+\xi_{j,n}\) and
\(\|\xi_{j,n}\|=\sin\theta_{j,n}\).

We use three summaries of the angular error. The worst angle is
\(\Theta_{s,n}:=\max_{j\in S_s}\theta_{j,n}\), with
\(\Theta_n:=\max_j\theta_{j,n}\). The mass-weighted mean-square transverse
error is
\(V_{s,n}:=D_{s,n}^{-1}\sum_{j\in S_s}p_{j,n}\|\xi_{j,n}\|^2\), and we
write \(W_{s,n}:=\sqrt{V_{s,n}}\). Finally,
\(Z_{s,n}:=\max_{j\in S_s}\|\xi_{j,n}\|\) controls the worst transverse
component. The balance defect is
\(\Delta_{j,n}:=a_{j,n}^2-\|w_{j,n}\|^2\).

\subsubsection{Scales, kernels, working region}

\[L:=\log(1/\varepsilon),\quad\bar\varepsilon:=\varepsilon L,\quad\kappa:=\frac\beta{2\sqrt m},\quad\tau:=\frac{\eta\beta}{\sqrt m}=2\eta\kappa,\quad h:=\eta\kappa=\tau/2;\]
\[K(\psi):=\frac{\sin\psi+(\pi-\psi)\cos\psi}{2\pi},\qquad q(\psi):=\frac{\pi-\psi}{2\pi};\]
\[K_{ij,n}:=K(\psi_{ij,n}),\quad q_{ij,n}:=q(\psi_{ij,n}),\quad\psi_{ij,n}:=\angle(u_{i,n},u_{j,n});\]
\[\mathcal R:=\bigl\{\Psi:\ \Pi\le6\beta,\ \ \tfrac12\le|a_j|/\|w_j\|\le2\ \ \forall j\bigr\}.\]

\subsubsection{Standing assumptions}

\begin{itemize}
\item
  \textbf{(P1)} \(d\ge2\); \(m\ge2\) even; \(\beta>0\);
  \(\theta_\star\in\mathbb S^{d-1}\); \(\delta\in(0,\tfrac12)\).
\item
  \textbf{(P2)} \(0<\varepsilon\le\varepsilon_0\);
  \(\Theta_\star:=A\bar\varepsilon\).
\item
  \textbf{(P3)} Initialization \ref{app:1.2}.
\item
  \textbf{(P4)} \(\tau L\le c_\eta\varepsilon\),
  i.e.~\(\eta\le c_\eta\sqrt m\,\varepsilon/(\beta L)\).
\item
  \textbf{(P5)} \(N_1:=\lceil L/\log(1+\tau/2)\rceil\),
  \(N_2:=N_1+\lceil C_L\tau^{-1}L\rceil\), \(N_\star:=N_2\).
\item
  \textbf{(P6)} \(B\ge C_B\,\varepsilon^{-2}L^2D_\star\) with
  \(D_\star:=d+\log\frac{16mN_\star}\delta\).
\end{itemize}

The universal constants \(c_0,c_1,c_\eta,C_B,C_L,C_T,A,\varepsilon_0\)
are fixed in the balanced-case proof below and, finally, in
Appendix~\ref{app:robust-init}. \(C\ge1\) and \(c\le1\) denote universal constants
that may change between occurrences. Error parameters:
\[\iota:=\varepsilon^2+\rho_B\ \ (\text{Phase I}),\qquad\zeta:=\Theta_\star^2+\rho_B+\tau\ \ (\text{Phase II}),\]
with \(\rho_B\) defined in Lemma~\ref{lem:concentration}. 

\paragraph{Role of (P6).} Condition P6 is specific to the fresh-batch realization of the deterministic error bounds used below: together with Lemmas B.12--B.13, it ensures that \(\rho_B\) is sufficiently small. The subsequent phase analysis is deterministic conditional on these error bounds. Appendix~\ref{app:single-example} establishes the corresponding error bounds blockwise for SGD on a fixed dataset, allowing the same deterministic phase analysis to be reused in that setting.

\begin{lemma}[Smallness budget]\label{lem:smallness-budget}
{Let $A\ge1$. Suppose
\(\varepsilon_0\le10^{-2}\) and
\(A^2\varepsilon_0\log^3(1/\varepsilon_0)\le1\). Then for every
\(0<\varepsilon\le\varepsilon_0\),}
\[
\begin{gathered}
L\ge\log100>4,\qquad
A^2\varepsilon L^3\le1,\qquad
\Theta_\star^2L\le\varepsilon,\\
\Theta_\star^2\le\varepsilon\le10^{-2},\qquad
\Theta_\star\le\tfrac1{10},\qquad
\varepsilon L\le1.
\end{gathered}
\]
{If moreover \(\rho_B\le c_1\varepsilon/L\) and
\(\tau\le c_\eta\varepsilon/L\) with \(c_1,c_\eta\le1\) (Lemma~\ref{lem:batch-consequence},
(P4)), then}
\[\zeta\le3\varepsilon,\qquad\zeta L\le3\varepsilon,\qquad\iota L\le2\varepsilon,\qquad\zeta\le3\bar\varepsilon.\tag{1.4} \label{app:1.4}\]
\end{lemma}
\begin{proof} \(g(\varepsilon):=\varepsilon\log^3(1/\varepsilon)\) has
\(g'(\varepsilon)=L^2(L-3)>0\) for \(L>3\), i.e.~on \((0,e^{-3})\).
Since \(10^{-2}<e^{-3}\approx0.0498\), \(g\) is increasing on
\((0,\varepsilon_0]\), so
\(A^2g(\varepsilon)\le A^2g(\varepsilon_0)\le1\). Then
\(\Theta_\star^2L=A^2\varepsilon^2L^3=\varepsilon\cdot A^2\varepsilon L^3\le\varepsilon\);
and since \(L\ge1\),
\(\Theta_\star^2=A^2\varepsilon^2L^2\le A^2\varepsilon^2L^3\le\varepsilon\le10^{-2}\),
hence \(\Theta_\star\le10^{-1}\). Also
\(\varepsilon L\le\varepsilon L^3\le A^{-2}\le1\). For \ref{app:1.4}:
\(\zeta=\Theta_\star^2+\rho_B+\tau\le\varepsilon+c_1\varepsilon+c_\eta\varepsilon\le3\varepsilon\);
\(\zeta L\le\Theta_\star^2L+c_1\varepsilon+c_\eta\varepsilon\le3\varepsilon\);
\(\iota L=\varepsilon^2L+\rho_BL\le\varepsilon+c_1\varepsilon\le2\varepsilon\)
(using \(\varepsilon L\le1\));
\(\zeta\le3\varepsilon\le3\bar\varepsilon\) (as \(L\ge1\)). 
\end{proof}

\subsection{Preliminaries}

\subsubsection{Standard facts}

This appendix uses the following standard result:

\begin{itemize}
\item
  \textbf{(E1)} Gaussian concentration:
  \(\|h(X)-\mathbb Eh(X)\|_{\psi_2}\le C\operatorname{Lip}(h)\) for
  \(X\sim\gamma_d\).
\item
  \textbf{(E2)} Bernstein's inequality for sums of i.i.d. centered
  sub-exponential random variables.
\item
  \textbf{(E3)} Volumetric bound: \(\mathbb S^{d-1}\) has a
  \(\tfrac12\)-net of cardinality \(\le5^d\).
\item
  \textbf{(E4)} Gaussian integration by parts (Stein's identity):
  \(\mathbb E[(\theta^\top X)h(X)]=\mathbb E[\theta^\top\nabla h(X)]\)
  for \(h\in C^1\) of at most polynomial growth, \(X\sim\gamma_d\),
  \(\theta\in\mathbb R^d\).
\end{itemize}

\subsubsection{Gaussian identities}

\begin{lemma}[Gaussian identities]\label{lem:gaussian} {Let \(u,v,\theta\in\mathbb S^{d-1}\),
\(\psi:=\angle(u,v)\), \(X\sim\gamma_d\). Then}
\[\text{(i) }\ \mathbb E\bigl[(\theta^\top X)\sigma(v^\top X)\bigr]=\tfrac12\,\theta^\top v;\qquad\qquad\text{(ii) }\ \mathbb E\bigl[(\theta^\top X)\mathbf 1_{\{u^\top X>0\}}X\bigr]=\tfrac12\,\theta;\]
\[\text{(iii) }\ \mathbb E\bigl[\sigma(v^\top X)\mathbf 1_{\{u^\top X>0\}}X\bigr]=q(\psi)\,v+\frac{\sin\psi}{2\pi}\,u;\]
\[\text{(iv) }\ \mathbb E\bigl[\sigma(u^\top X)\sigma(v^\top X)\bigr]=K(\psi),\qquad\mathbb P(u^\top X>0,\ v^\top X>0)=q(\psi).\]
\end{lemma}
\begin{proof} Fix a unit \(u^\perp\perp u\) such that
\(\operatorname{span}(u,u^\perp)\) contains \(v\) (for (iii),(iv)) resp.
\(\theta\) (for (ii)); write \(X=t\,u+s\,u^\perp+X_0\) with
\(t,s\sim\mathcal N(0,1)\) independent and \(X_0\) independent of
\((t,s)\), mean zero, supported on
\(\operatorname{span}(u,u^\perp)^\perp\). Put
\(v=\cos\psi\,u+\sin\psi\,u^\perp\), so
\(v^\top X=\cos\psi\,t+\sin\psi\,s\).

\textbf{(iv)} Write \((t,s)=R(\cos\Phi,\sin\Phi)\). By rotational
invariance of \(\mathcal N(0,I_2)\), \(\Phi\sim\mathrm{Unif}[0,2\pi)\)
and \(R\) (Rayleigh, density \(re^{-r^2/2}\) on \((0,\infty)\),
\(\mathbb ER^2=2\)) are independent. Then \(u^\top X=R\cos\Phi\) and
\(v^\top X=R\cos(\Phi-\psi)\). Both are positive iff
\(\Phi\in(-\tfrac\pi2,\tfrac\pi2)\cap(\psi-\tfrac\pi2,\psi+\tfrac\pi2)=(\psi-\tfrac\pi2,\tfrac\pi2)\),
an arc of length \(\pi-\psi\); hence the probability equals
\(\frac{\pi-\psi}{2\pi}=q(\psi)\). Next,
\[
\begin{aligned}
\mathbb E[\sigma(u^\top X)\sigma(v^\top X)]
&=\mathbb E[R^2]\frac1{2\pi}
\int_{\psi-\pi/2}^{\pi/2}\cos\phi\,\cos(\phi-\psi)\,d\phi\\
&=\frac1{2\pi}\int_{\psi-\pi/2}^{\pi/2}
\bigl[\cos\psi+\cos(2\phi-\psi)\bigr]\,d\phi .
\end{aligned}
\]
using \(2\cos\phi\cos(\phi-\psi)=\cos\psi+\cos(2\phi-\psi)\) and
\(\mathbb ER^2=2\). Now
\(\int_{\psi-\pi/2}^{\pi/2}\cos\psi\,d\phi=(\pi-\psi)\cos\psi\), and
\[\int_{\psi-\pi/2}^{\pi/2}\cos(2\phi-\psi)\,d\phi=\tfrac12\bigl[\sin(2\phi-\psi)\bigr]_{\psi-\pi/2}^{\pi/2}=\tfrac12\bigl[\sin(\pi-\psi)-\sin(\psi-\pi)\bigr]=\sin\psi .\]
Hence \(K(\psi)=\frac{(\pi-\psi)\cos\psi+\sin\psi}{2\pi}\).

\textbf{(i)} Apply (E4) to the mollifications
\(h_\epsilon:=\sigma(v^\top\cdot)*\varphi_\epsilon\in C^\infty\):
\(\mathbb E[(\theta^\top X)h_\epsilon(X)]=\mathbb E[\theta^\top\nabla h_\epsilon(X)]\).
As \(\epsilon\downarrow0\), \(h_\epsilon\to\sigma(v^\top\cdot)\)
pointwise with \(|h_\epsilon(x)|\le|v^\top x|+\epsilon\), and
\(\nabla h_\epsilon\to v\,\mathbf 1_{\{v^\top x>0\}}\) for a.e. \(x\)
with \(\|\nabla h_\epsilon\|_\infty\le1\); dominated convergence gives
\(\mathbb E[(\theta^\top X)\sigma(v^\top X)]=(\theta^\top v)\mathbb P(v^\top X>0)=\tfrac12\theta^\top v\).

\textbf{(ii)} For any \(e\perp\operatorname{span}(u,\theta)\), the
random variable \(e^\top X\) is independent of \((\theta^\top X,t)\) and
has mean \(0\), so the \(e\)-component of the left side vanishes; and
\(\theta^\top e=0\). So it suffices to compute the components along
\(u\) and \(u^\perp\), with
\(\theta=\cos\phi_0\,u+\sin\phi_0\,u^\perp\):
\[u\text{-component}=\mathbb E[(\cos\phi_0\,t+\sin\phi_0\,s)\mathbf 1_{\{t>0\}}t]=\cos\phi_0\,\mathbb E[t^2\mathbf 1_{\{t>0\}}]=\tfrac12\cos\phi_0,\]
\[u^\perp\text{-component}=\mathbb E[(\cos\phi_0\,t+\sin\phi_0\,s)\mathbf 1_{\{t>0\}}s]=\sin\phi_0\,\mathbb E[s^2]\,\mathbb P(t>0)=\tfrac12\sin\phi_0 .\]

\textbf{(iii)} Components orthogonal to \(\operatorname{span}(u,v)\)
vanish by the argument in (ii), and the right-hand side lies in that
span. The \(u\)-component of the left side is
\(\mathbb E[\sigma(v^\top X)\mathbf 1_{\{t>0\}}t]=\mathbb E[\sigma(v^\top X)\sigma(u^\top X)]=K(\psi)\),
and the \(u\)-component of the right side is
\(q(\psi)\cos\psi+\frac{\sin\psi}{2\pi}=\frac{(\pi-\psi)\cos\psi+\sin\psi}{2\pi}=K(\psi)\).
For the \(u^\perp\)-component, condition on \(t\) and apply
one-dimensional (E4) in \(s\) (again via mollification):
\(\mathbb E_s[\sigma(\cos\psi\,t+\sin\psi\,s)\,s]=\sin\psi\,\mathbb P_s(v^\top X>0)\),
so
\[\mathbb E[\sigma(v^\top X)\mathbf 1_{\{t>0\}}s]=\sin\psi\,\mathbb P(t>0,\ v^\top X>0)=\sin\psi\,q(\psi),\]
matching the \(u^\perp\)-component \(q(\psi)\sin\psi\) of the right
side. 
\end{proof}

\begin{remark}[Boundary term]\label{rem:boundary-term}
\(\frac{\sin\psi}{2\pi}u=\phi(0)\,\mathbb E[\sigma(v^\top X)\mid u^\top X=0]\,u\)
with \(\phi(0)=(2\pi)^{-1/2}\). This term is parallel to \(u\), hence
annihilated by \(I-uu^\top\). We record for
later: \(K(\psi)-q(\psi)\cos\psi=\frac{\sin\psi}{2\pi}\).
\end{remark}

\begin{lemma}[Kernel bounds]\label{lem:kernel-bounds}{For \(\psi\in[0,\pi]\):
\(\ 0\le\tfrac12-K(\psi)\le\tfrac{\psi^2}4\);
\(\ 0\le K(\pi-\psi)\le\tfrac{\psi^3}{6\pi}\);
\(\ 0\le q(\psi)\le\tfrac12\); \(\ q(\pi-\psi)=\tfrac\psi{2\pi}\).}
\end{lemma}
\begin{proof} \(K'(\psi)=-\frac{(\pi-\psi)\sin\psi}{2\pi}\), so
\(K'\le0\) and \(|K'(\psi)|\le\frac{\pi\psi}{2\pi}=\frac\psi2\);
integrating from \(0\) and using \(K(0)=\tfrac12\) gives the first
claim. Next \(K(\pi-\psi)=\frac{\sin\psi-\psi\cos\psi}{2\pi}\) and
\(\sin\psi-\psi\cos\psi=\int_0^\psi t\sin t\,dt\in[0,\psi^3/3]\). The
statements for \(q\) are immediate. 
\end{proof}

\subsubsection{Elementary discrete lemmas}

\begin{lemma}[Normalization remainder]\label{lem:normalization-remainder} {For
\(u\in\mathbb S^{d-1}\) and \(\|\varpi\|\le\tfrac15\),}
\[\Bigl\|\frac{u+\varpi}{\|u+\varpi\|}-u-(I-uu^\top)\varpi\Bigr\|\ \le\ 14\|\varpi\|^2 .\]
\end{lemma}
\begin{proof} Let \(\sigma_0:=u^\top\varpi\),
\(\varpi^\perp:=(I-uu^\top)\varpi\), \(\|u+\varpi\|=:1+\mu\). Then
\((1+\mu)^2=1+2\sigma_0+\|\varpi\|^2=:1+x\) with
\(|x|\le2\|\varpi\|+\|\varpi\|^2\le0.44\le\tfrac12\), so
\(\mu=\sqrt{1+x}-1\). Since \(|\sqrt{1+y}-1-\tfrac y2|\le\tfrac{y^2}2\)
for \(|y|\le\tfrac12\),
\[|\mu-\sigma_0|\le\bigl|\mu-\tfrac x2\bigr|+\bigl|\tfrac x2-\sigma_0\bigr|\le\tfrac{x^2}2+\tfrac{\|\varpi\|^2}2\le\tfrac{(2.2\|\varpi\|)^2}2+\tfrac{\|\varpi\|^2}2\le3\|\varpi\|^2 .\]
Hence \(|\mu|\le|\sigma_0|+3\|\varpi\|^2\le1.6\|\varpi\|\le0.32\) and
\(1+\mu\ge\tfrac12\). Finally, using \(\varpi=\sigma_0u+\varpi^\perp\),
\[\frac{u+\varpi}{1+\mu}-u-\varpi^\perp=\frac{(u+\varpi)-(1+\mu)(u+\varpi^\perp)}{1+\mu}=\frac{(\sigma_0-\mu)u-\mu\varpi^\perp}{1+\mu},\]
whose norm is
\(\le2\bigl(3\|\varpi\|^2+1.6\|\varpi\|\cdot\|\varpi\|\bigr)\le14\|\varpi\|^2\).

\end{proof}

\begin{lemma}[Norm remainder]\label{lem:norm-remainder} {If \(w\ne0\), \(r=\|w\|\),
\(u=w/r\) and \(\eta\|V\|\le r/5\), then}
\[\bigl|\ \|w+\eta V\|-r-\eta\,u^\top V\ \bigr|\ \le\ \frac{3\eta^2\|V\|^2}{r}.\]
\end{lemma}
\begin{proof} \(\|w+\eta V\|=r\sqrt{1+x}\) with
\(x=2\eta u^\top V/r+\eta^2\|V\|^2/r^2\). With
\(\epsilon:=\eta\|V\|/r\le\tfrac15\),
\(|x|\le2\epsilon+\epsilon^2\le0.44\le\tfrac12\), so
\(r\sqrt{1+x}=r+\tfrac{rx}2+r\vartheta\) with
\(|\vartheta|\le\tfrac{x^2}2\). Now
\(\tfrac{rx}2=\eta u^\top V+\tfrac{\eta^2\|V\|^2}{2r}\) and
\(\tfrac{rx^2}2\le\tfrac{r(2.2\epsilon)^2}2=2.42\,\tfrac{\eta^2\|V\|^2}r\).
Summing, the error is \(\le3\eta^2\|V\|^2/r\). 
\end{proof}

\begin{lemma}[From squared to unsquared error]\label{lem:v-to-w} {Let \(x,x_+\ge0\),
\(a\in\mathbb R\) with \(\eta|a|\le\tfrac12\), \(b\ge0\), and suppose
\(x_+^2\le x^2-2\eta ax^2+2\eta bx+4\eta^2b^2\). Then
\(x_+\le(1-\eta a)x+3\eta b\).}
\end{lemma}
\begin{proof} Put \(y:=(1-\eta a)x+3\eta b\ge0\). Using
\((1-\eta a)^2\ge1-2\eta a\) and \(1-\eta a\ge\tfrac12\),
\[y^2\ge(1-2\eta a)x^2+6\eta b(1-\eta a)x+9\eta^2b^2\ge x^2-2\eta ax^2+3\eta bx+9\eta^2b^2\ge x_+^2 .\]
\end{proof}

\begin{lemma}[Scalar recursions]\label{lem:scalar-recursions} {Let \((x_n)_{n\ge n_0}\)
be nonnegative.} 1. {If \(x_{n+1}\le(1-\lambda_n)x_n+g\) for
\(n\in[n_0,N)\), with \(\lambda_n\in[\lambda,1]\), \(\lambda>0\),
\(g\ge0\), then \(x_n\le\max\{x_{n_0},\,g/\lambda\}\) for all
\(n\in[n_0,N]\).} 2. {If \(x_{n+1}\le(1+\mu)x_n+g\) for
\(n\in[n_0,N)\), with \(\mu,g\ge0\), then
\(x_n\le e^{\mu(n-n_0)}\bigl(x_{n_0}+g\,(n-n_0)\bigr)\).}
\end{lemma}
\begin{proof} (1) If \(x_n\le g/\lambda\) then
\(x_{n+1}\le(1-\lambda_n)\tfrac g\lambda+g=\tfrac g\lambda+g\bigl(1-\tfrac{\lambda_n}\lambda\bigr)\le\tfrac g\lambda\).
If \(x_n>g/\lambda\) then
\(x_{n+1}\le x_n-\lambda_nx_n+g<x_n-\lambda_n\tfrac g\lambda+g\le x_n\).
Induct. (2) Unrolling,
\(x_n\le(1+\mu)^{n-n_0}x_{n_0}+g\sum_{k<n-n_0}(1+\mu)^k\le(1+\mu)^{n-n_0}\bigl(x_{n_0}+g(n-n_0)\bigr)\)
and \((1+\mu)^k\le e^{\mu k}\). 
\end{proof}

\begin{lemma}[Balance ratio]\label{lem:balance-ratio}
{If \(b,r>0\) and
\(|b^2-r^2|\le\varrho\,br\) with \(0\le\varrho\le\tfrac12\), then
\(\bigl|\tfrac br-1\bigr|\le\varrho\) and
\(\bigl|\tfrac br+\tfrac rb-2\bigr|\le2\varrho^2\).}
\end{lemma}
\begin{proof} \(x:=b/r>0\) satisfies \(|x^2-1|\le\varrho x\). If
\(x\ge1\): \(x-1=\frac{x^2-1}{x+1}\le\frac{\varrho x}{x+1}\le\varrho\).
If \(x<1\): \(1-x=\frac{1-x^2}{1+x}\le\frac{\varrho x}{1+x}\le\varrho\).
Then \(x\ge1-\varrho\ge\tfrac12\) and
\(x+\tfrac1x-2=\tfrac{(x-1)^2}x\le2\varrho^2\). 
\end{proof}

\subsection{Sub-Gaussian bounds on the working region}

We use the standard facts
\(\|YZ\|_{\psi_1}\le\|Y\|_{\psi_2}\|Z\|_{\psi_2}\) and
\(\|v^\top X\|_{\psi_2}\le C\) for \(v\in\mathbb S^{d-1}\), plus:

\begin{lemma}[Centering]\label{lem:centering}
{\(\|Y-\mathbb EY\|_{\psi_1}\le C\|Y\|_{\psi_1}\).}
\end{lemma}
\begin{proof} \(\|\cdot\|_{\psi_1}\) is a norm and
\(\|c\|_{\psi_1}=|c|/\log2\) for constants. From \(t\le e^t\),
\(|\mathbb EY|\le\mathbb E|Y|\le\|Y\|_{\psi_1}\mathbb E\bigl[e^{|Y|/\|Y\|_{\psi_1}}\bigr]\le2\|Y\|_{\psi_1}\).
Hence \(\|Y-\mathbb EY\|_{\psi_1}\le(1+2/\log2)\|Y\|_{\psi_1}\).

\end{proof}

\begin{lemma}[Residual bounds]\label{lem:residual-bounds}{If \(\Psi\in\mathcal R\) and \(R=f_\Psi-y\),
then}
\[
\begin{aligned}
\operatorname{Lip}(f_\Psi)&\le\Pi\le6\beta,
&\operatorname{Lip}(R)&\le7\beta,\\
|\mathbb ER(X)|&\le\frac{\Pi}{\sqrt{2\pi}}\le3\beta,
&\|R(X)\|_{\psi_2}&\le C\beta,\\
\|R\|_{L^2}&\le\frac{\Pi}{\sqrt2}+\beta\le6\beta.
\end{aligned}
\]
\end{lemma}
\begin{proof}
\(\nabla f_\Psi(x)=\frac1{\sqrt m}\sum_ja_j\mathbf 1_{\{w_j^\top x>0\}}w_j\)
a.e., so \(\|\nabla f_\Psi\|\le\frac1{\sqrt m}\sum_j|a_j|\|w_j\|=\Pi\);
\(f_\Psi\) is Lipschitz as a finite sum of Lipschitz functions, and
\(\operatorname{Lip}(y)=\beta\). Since \(\mathbb Ey(X)=0\) and
\(\mathbb E\sigma(w^\top X)=\|w\|/\sqrt{2\pi}\),
\(|\mathbb ER|=|\mathbb Ef_\Psi|\le\Pi/\sqrt{2\pi}\). By (E1),
\(\|R-\mathbb ER\|_{\psi_2}\le C\operatorname{Lip}(R)\le7C\beta\), and
\(\|c\|_{\psi_2}\le C|c|\) for constants. Finally
\(\|\sigma(u^\top X)\|_{L^2}=1/\sqrt2\) for \(u\in\mathbb S^{d-1}\).

\end{proof}

\begin{lemma}[Population-gradient bounds]\label{lem:population-gradient-bounds}{On \(\mathcal R\):
\(|g^a_j|\le C\beta r_j/\sqrt m\) and
\(\|g^w_j\|\le C\beta b_j/\sqrt m\).}
\end{lemma}
\begin{proof}
By Cauchy--Schwarz,
\[
|g^a_j|
\le\frac{r_j}{\sqrt m}\|R\|_2\|\sigma(u_j^\top X)\|_2,
\qquad
|v^\top g^w_j|
\le\frac{b_j}{\sqrt m}\|R\|_2\|v^\top X\|_2
\]
for every unit vector \(v\). Apply Lemma~\ref{lem:residual-bounds}.
\end{proof}

\subsection{Adaptive mini-batch concentration}

\begin{lemma}[Uniform adaptive mini-batch concentration]\label{lem:concentration}
{There is a universal \(C_\star\ge1\) such that for every integer
\(N\ge1\) and every \(\delta\in(0,1)\), setting}
\[D_N:=d+\log\frac{8mN}\delta,\qquad\rho_B:=C_\star\Bigl(\sqrt{\tfrac{D_N}B}+\tfrac{D_N}B\Bigr),\]
{the event \(\mathcal E\) that for every \(0\le n<N\) with
\(\Psi_n\in\mathcal R\) and every \(j\le m\)}
\[\bigl|\widehat g^a_{j,n}-g^a_{j,n}\bigr|\le\frac{\beta r_{j,n}}{\sqrt m}\rho_B,\qquad\bigl\|\widehat g^w_{j,n}-g^w_{j,n}\bigr\|\le\frac{\beta b_{j,n}}{\sqrt m}\rho_B\tag{4.1}\label{app:4.1}\]
{satisfies \(\mathbb P(\mathcal E)\ge1-\delta\).}
\end{lemma}
\begin{proof} Fix \(n<N\) and \(j\le m\); condition on
\(\mathcal F_n\) and work on the \(\mathcal F_n\)-measurable event
\(\{\Psi_n\in\mathcal R\}\), on which
\(R_n,a_{j,n},w_{j,n},r_{j,n},b_{j,n}\) are constants and the batch
\((X_{n,b})_b\) is i.i.d. \(\gamma_d\) independent of \(\mathcal F_n\).

{Second layer.} Let
\(Y_b:=\frac1{\sqrt m}R_n(X_{n,b})\sigma(w_{j,n}^\top X_{n,b})\), so
\(\widehat g^a_{j,n}=\frac1B\sum_bY_b\) and
\(\mathbb E[Y_b\mid\mathcal F_n]=g^a_{j,n}\). By Lemma~\ref{lem:residual-bounds} and
\(\|\sigma(w_j^\top X)\|_{\psi_2}\le r_j\|u_j^\top X\|_{\psi_2}\le Cr_j\),
\[\|Y_b\|_{\psi_1}\le\frac1{\sqrt m}\|R_n(X)\|_{\psi_2}\|\sigma(w_{j,n}^\top X)\|_{\psi_2}\le\frac{C_1'\beta r_{j,n}}{\sqrt m},\]
and by Lemma~\ref{lem:centering} the centered summand satisfies
\(\|Y_b-\mathbb EY_b\|_{\psi_1}\le\frac{C_1\beta r_{j,n}}{\sqrt m}=:K_a\).
By (E2) there is a universal \(c_2\in(0,1]\) with
\[\mathbb P\Bigl(\bigl|\tfrac1B\textstyle\sum_b(Y_b-\mathbb EY_b)\bigr|>t\Bigr)\le2\exp\Bigl[-c_2B\min\bigl(\tfrac{t^2}{K_a^2},\tfrac t{K_a}\bigr)\Bigr].\]
Set \(t:=K_a\,c_2^{-1}\bigl(\sqrt{u/B}+u/B\bigr)\) and let
\(\theta_0:=t/K_a\ge c_2^{-1}\max\{\sqrt{u/B},\,u/B\}\). {Case
\(u\le B\):} \(\theta_0^2\ge c_2^{-2}u/B\ge c_2^{-1}u/B\) and
\(\theta_0\ge c_2^{-1}u/B\), so \(c_2B\min(\theta_0^2,\theta_0)\ge u\).
{Case \(u>B\):} \(\theta_0\ge c_2^{-1}u/B>c_2^{-1}\ge1\), hence
\(\theta_0^2\ge\theta_0\) and
\(\min(\theta_0^2,\theta_0)=\theta_0\ge c_2^{-1}u/B\), so again
\(c_2B\min(\cdot)\ge u\). Therefore
\[\mathbb P\Bigl(\bigl|\widehat g^a_{j,n}-g^a_{j,n}\bigr|>\tfrac{C_3\beta r_{j,n}}{\sqrt m}\bigl(\sqrt{u/B}+u/B\bigr)\ \Big|\ \mathcal F_n\Bigr)\le2e^{-u},\qquad C_3:=C_1/c_2 .\tag{4.2} \label{app:4.2}\]

{First layer.} For fixed \(v\in\mathbb S^{d-1}\) put
\(Y^v_b:=\frac1{\sqrt m}R_n(X_{n,b})a_{j,n}\mathbf 1_{\{w_{j,n}^\top X_{n,b}>0\}}(v^\top X_{n,b})\).
Since \(|\mathbf 1_{\{\cdot\}}v^\top X|\le|v^\top X|\) pointwise,
\(\|Y^v_b\|_{\psi_1}\le\frac{C_1'\beta b_{j,n}}{\sqrt m}\) uniformly in
\(v\), so \ref{app:4.2} holds for \(|v^\top(\widehat g^w_{j,n}-g^w_{j,n})|\)
with \(r_{j,n}\) replaced by \(b_{j,n}\). Take a \(\tfrac12\)-net
\(\mathcal N\) of \(\mathbb S^{d-1}\) with \(|\mathcal N|\le5^d\) (E3);
then \(\|z\|\le2\max_{v\in\mathcal N}v^\top z\) for every
\(z\in\mathbb R^d\). Applying the scalar bound with \(u\) replaced by
\(u+d\log5\) at each \(v\in\mathcal N\) and union bounding, the failure
probability is \(\le2\cdot5^de^{-u-d\log5}=2e^{-u}\), and on the
complement
\[\|\widehat g^w_{j,n}-g^w_{j,n}\|\le\frac{2C_3\beta b_{j,n}}{\sqrt m}\Bigl(\sqrt{\tfrac{u+d\log5}B}+\tfrac{u+d\log5}B\Bigr).\tag{4.3} \label{app:4.3}\]

{Union bound.} Take \(u:=\log\frac{8mN}\delta\); since
\(\log5\le2\), \(u+d\log5\le2D_N\), so with \(C_\star:=4C_3\) the right
sides of \ref{app:4.2}, \ref{app:4.3} are at most
\(\frac{\beta r_{j,n}}{\sqrt m}\rho_B\) and
\(\frac{\beta b_{j,n}}{\sqrt m}\rho_B\). Let \(\mathcal B_{n,j}\) be the
event that \ref{app:4.1} fails at \((n,j)\) \textbf{and}
\(\Psi_n\in\mathcal R\). As \(\{\Psi_n\in\mathcal R\}\in\mathcal F_n\),
\[\mathbb P(\mathcal B_{n,j})=\mathbb E\bigl[\mathbf 1_{\{\Psi_n\in\mathcal R\}}\,\mathbb P(\text{\ref{app:4.1} fails at }(n,j)\mid\mathcal F_n)\bigr]\le4e^{-u}=\frac\delta{2mN}.\]
Union bounding over \(j\le m\) and \(n<N\) gives
\(\mathbb P(\mathcal E^c)\le\delta/2\le\delta\). 
\end{proof}

\begin{remark}[Adaptivity]\label{rem:adaptivity}
Adaptivity is handled entirely by the
\(\mathcal F_n\)-measurability of \(\{\Psi_n\in\mathcal R\}\); no
stopping-time device is needed. On \(\mathcal E\), the bounds \ref{app:4.1} hold
at every iteration whose state lies in \(\mathcal R\), however the
trajectory arrived there. The subsequent phase estimates are deterministic on the event \(\mathcal E\).
\end{remark}

\begin{lemma}[Batch-size consequence]\label{lem:batch-consequence}{Under (P6) with \(N=N_\star\) and
\(C_B\ge(2C_\star/c_1)^2\), we have \(\rho_B\le c_1\varepsilon/L\le1\).}
\end{lemma}
\begin{proof} \(D_{N_\star}\le D_\star\). From
\(B\ge C_B\varepsilon^{-2}L^2D_\star\):
\(\sqrt{D_\star/B}\le\varepsilon/(L\sqrt{C_B})\), and
\(D_\star/B\le\varepsilon^2/(C_BL^2)\le\varepsilon/(L\sqrt{C_B})\) since
\(\varepsilon\le1\le L\) and \(C_B\ge1\). Hence
\(\rho_B\le2C_\star\varepsilon/(L\sqrt{C_B})\le c_1\varepsilon/L\).

\end{proof}

\subsection{Balance}

\begin{lemma}[First-order balance cancellation]\label{lem:balance-cancellation}
{For \ref{app:1.1}, for every \(j\), every \(n\), and every realization of
the batch,}
\[\Delta_{j,n+1}=\Delta_{j,n}+\eta^2\Bigl[(\widehat g^a_{j,n})^2-\|\widehat g^w_{j,n}\|^2\Bigr].\tag{5.1} \label{app:5.1}\]
\end{lemma}
\begin{proof} Samplewise,
\(w_j^\top\bigl[a_j\mathbf 1_{\{w_j^\top x>0\}}x\bigr]=a_j\mathbf 1_{\{w_j^\top x>0\}}(w_j^\top x)=a_j\sigma(w_j^\top x)\).
Averaging over the batch with the common weight
\(\frac1{B\sqrt m}R_n(x)\) gives the exact identity
\[a_{j,n}\,\widehat g^a_{j,n}=w_{j,n}^\top\widehat g^w_{j,n}.\tag{5.2} \label{app:5.2}\]
Expanding,
\[\Delta_{j,n+1}=(a_j-\eta\widehat g^a_j)^2-\|w_j-\eta\widehat g^w_j\|^2=\Delta_{j,n}-2\eta\bigl(a_j\widehat g^a_j-w_j^\top\widehat g^w_j\bigr)+\eta^2\bigl[(\widehat g^a_j)^2-\|\widehat g^w_j\|^2\bigr],\]
and the \(O(\eta)\) term vanishes by \ref{app:5.2}. 
\end{proof}

\begin{corollary}[One-step balance drift]\label{cor:balance-drift}
{On \(\mathcal E\), if
\(\Psi_n\in\mathcal R\) then
\(|\Delta_{j,n+1}-\Delta_{j,n}|\le C_\sharp\,\tau^2p_{j,n}\) for a
universal \(C_\sharp\).}
\end{corollary}
\begin{proof} By Lemma~\ref{lem:population-gradient-bounds} and \ref{app:4.1} with \(\rho_B\le1\):
\(|\widehat g^a_j|\le C\beta r_j/\sqrt m\),
\(\|\widehat g^w_j\|\le C\beta b_j/\sqrt m\). On \(\mathcal R\),
\(r_j^2+b_j^2=p_j\bigl(\tfrac{r_j}{b_j}+\tfrac{b_j}{r_j}\bigr)\le\tfrac52p_j\).
Insert into \ref{app:5.1} and use \(\eta^2\beta^2/m=\tau^2\). 
\end{proof}

\begin{lemma}[Accumulated balance drift]\label{lem:balance-propagation}{On \(\mathcal E\), if \(\Psi_k\in\mathcal R\)
for all \(k<n\), then
\(|\Delta_{j,n}|\le C_\sharp\tau^2\sum_{k<n}p_{j,k}\).}
\end{lemma}
\begin{proof} \(\Delta_{j,0}=0\) by \ref{app:1.2}; telescope Corollary~\ref{cor:balance-drift}.

\end{proof}
\paragraph{What the phase analysis needs.}
The phase analysis below only needs two properties of each update:
the stochastic gradient must remain close to the population gradient,
and the layer imbalance \(\Delta_j = b_j^2-r_j^2\) must change slowly . Specifically, for updates
of the form~(\ref{app:1.1}), suppose that on $\mathcal R$,
\[
|\widehat g_j^a-g_j^a|
\le \frac{\beta r_j}{\sqrt m}\rho,
\qquad
\|\widehat g_j^w-g_j^w\|
\le \frac{\beta b_j}{\sqrt m}\rho,
\qquad
\rho\le\varepsilon/L,
\]
and
\[
|\Delta_j^+-\Delta_j|
\le C_\sharp\tau^2p_j.
\]
Here $g_j^a,g_j^w$ denote the population gradients at the current
state. Under the same step-size and initialization conditions, the
phase analysis below applies with $\rho$ in place of $\rho_B$. The
second bound controls the accumulated layer imbalance.

The preceding lemmas establish these bounds in the fresh-batch setting.
For fixed-dataset SGD, Lemma~\ref{se:lem:interval} establishes their
blockwise counterparts, with normalized step $\omega=K\tau$ in place
of $\tau$.

\subsection{Phase I: perturbed hyperbolic dynamics}

The purpose of Phase I is to show that the teacher creates a uniform
longitudinal force before the network itself becomes large enough to
interfere. The pair \((a_j,\alpha_j)\) then behaves like a perturbed
two-dimensional hyperbolic system: one mode grows and the other decays.
The transverse component receives only the small network and sampling
errors.

Set \(\alpha_{j,n}:=\theta_\star^\top w_{j,n}\),
\(w^\perp_{j,n}:=(I-\theta_\star\theta_\star^\top)w_{j,n}\), and
\(\tau_1:=\inf\{n\ge0:\|f_n\|_{L^2(\gamma_d)}>\beta\varepsilon^2\}\).

\begin{lemma}[Phase-I dynamics]\label{lem:phase-one-dynamics}
{Work on \(\mathcal E\). If \(n<\min(\tau_1,N_\star)\) and
\(\Psi_n\in\mathcal R\), then for every \(j\)}
\[a_{j,n+1}=a_{j,n}+\eta\kappa\alpha_{j,n}+\eta e^a_{j,n},\qquad\alpha_{j,n+1}=\alpha_{j,n}+\eta\kappa a_{j,n}+\eta e^\alpha_{j,n},\tag{6.1} \label{app:6.1}\]
\[|e^a_{j,n}|\le C\kappa\iota\,r_{j,n},\qquad|e^\alpha_{j,n}|\le C\kappa\iota\,b_{j,n},\qquad\|w^\perp_{j,n+1}-w^\perp_{j,n}\|\le C\eta\kappa\iota\,b_{j,n}.\tag{6.2}\label{app:6.2}\]
{If in addition \(\Psi_k\in\mathcal R\) and
\(k<\min(\tau_1,N_\star)\) for all \(k\le n\), then}
\[b_{j,n+1}+r_{j,n+1}\le2\sqrt{p_0}\,(1+h)^{n+1}e^{C\iota h(n+1)}.\tag{6.3} \label{app:6.3}\]
\end{lemma}
\begin{proof} {Second layer.}
\(-g^a_j=\frac1{\sqrt m}\mathbb E[(y-f_n)\sigma(w_j^\top X)]\). By
Lemma~\ref{lem:gaussian}(i),
\[\frac1{\sqrt m}\mathbb E[y\,\sigma(w_j^\top X)]=\frac{\beta r_j}{\sqrt m}\mathbb E[(\theta_\star^\top X)\sigma(u_j^\top X)]=\frac{\beta r_j}{2\sqrt m}\theta_\star^\top u_j=\kappa\alpha_j .\]
By Cauchy--Schwarz and \(n<\tau_1\),
\(\frac1{\sqrt m}|\mathbb E[f_n\sigma(w_j^\top X)]|\le\frac{r_j}{\sqrt m}\|f_n\|_2\|\sigma(u_j^\top X)\|_2\le\sqrt2\,\kappa\varepsilon^2r_j\).
On \(\mathcal E\),
\(|\widehat g^a_j-g^a_j|\le\frac{\beta r_j}{\sqrt m}\rho_B=2\kappa\rho_Br_j\).
Since \(a_{j,n+1}=a_{j,n}-\eta\widehat g^a_j\), this gives \ref{app:6.1}--\ref{app:6.2}
for \(e^a\).

{Radial first layer.}
\(-g^w_j=\frac{a_j}{\sqrt m}\mathbb E[(y-f_n)\mathbf 1_{\{w_j^\top X>0\}}X]\).
By Lemma~\ref{lem:gaussian}(ii),
\(\mathbb E[y\mathbf 1_{\{u_j^\top X>0\}}X]=\frac\beta2\theta_\star\),
so
\[-\theta_\star^\top g^w_j=\kappa a_j-\frac{a_j}{\sqrt m}\mathbb E\bigl[f_n\mathbf 1_{\{u_j^\top X>0\}}\theta_\star^\top X\bigr],\qquad\bigl|\mathbb E[f_n\mathbf 1\,\theta_\star^\top X]\bigr|\le\|f_n\|_2\|\theta_\star^\top X\|_2\le\beta\varepsilon^2 .\]
With
\(|\theta_\star^\top(\widehat g^w_j-g^w_j)|\le\frac{\beta b_j}{\sqrt m}\rho_B\)
this gives \ref{app:6.1}--\ref{app:6.2} for \(e^\alpha\).

{Transverse.}
\((I-\theta_\star\theta_\star^\top)\frac\beta2\theta_\star=0\), and by
duality
\[\|(I-\theta_\star\theta_\star^\top)g^w_j\|\le\frac{b_j}{\sqrt m}\sup_{v\perp\theta_\star,\|v\|=1}\bigl|\mathbb E[f_n\mathbf 1_{\{u_j^\top X>0\}}v^\top X]\bigr|\le\frac{b_j}{\sqrt m}\|f_n\|_2\le2\kappa\varepsilon^2b_j ;\]
adding the noise \(2\kappa\rho_Bb_j\) gives the last bound in \ref{app:6.2}.

{Growth.} From \ref{app:6.1}--\ref{app:6.2} and \(|\alpha_j|\le r_j\):
\(b_{j,k+1}\le b_{j,k}+h(1+C\iota)r_{j,k}\). Also
\(\|g^w_j\|\le\frac{b_j}{\sqrt m}\bigl(\frac\beta2+\beta\varepsilon^2\bigr)\),
so with the noise \(\|\widehat g^w_j\|\le\kappa b_j(1+C\iota)\) and
\(r_{j,k+1}\le r_{j,k}+h(1+C\iota)b_{j,k}\). Adding, \(b+r\) grows per
step by a factor \(\le1+h(1+C\iota)\le(1+h)e^{C\iota h}\); iterate from
\(b_{j,0}+r_{j,0}=2\sqrt{p_0}\). 
\end{proof}

\subsection{End of Phase I}

\begin{lemma}[Horizon arithmetic]\label{lem:horizon-arithmetic} {For \(0<h\le\tfrac12\)
and \(L\ge1\):}
\[
\begin{aligned}
L&\le hN_1\le2L+h\le3L,\\
(1+h)^{N_1}&\in[\varepsilon^{-1},2\varepsilon^{-1}],\qquad
(1-h)^{N_1}\le\varepsilon,\\
N_1&\le\frac{2L}{h}+1=\frac{4L}{\tau}+1.
\end{aligned}
\]
\end{lemma}

\begin{proof} By definition \(N_1\log(1+h)\in[L,\ L+\log(1+h)]\). Since
\(\log(1+h)\le h\), \(hN_1\ge N_1\log(1+h)\ge L\). Since
\(\log(1+h)\ge h/2\) for \(h\in(0,1]\),
\(N_1\le\frac{L}{\log(1+h)}+1\le\frac{2L}h+1\), whence
\(hN_1\le2L+h\le3L\) (using \(h\le1\le L\)). Also
\((1+h)^{N_1}=e^{N_1\log(1+h)}\in[e^L,\ e^L(1+h)]\subseteq[\varepsilon^{-1},2\varepsilon^{-1}]\),
and \((1-h)^{N_1}\le e^{-hN_1}\le e^{-L}=\varepsilon\). 
\end{proof}

\begin{proposition}[End of Phase I]\label{prop:phase-one}
{Under (P1)--(P6), on the event \(\mathcal E\):}

\begin{enumerate}
\def\labelenumi{\arabic{enumi}.}
\item
  {\(N_1<\tau_1\) and \(\Psi_n\in\mathcal R\) for all
  \(0\le n\le N_1\);}
\item
  {\(\operatorname{sign}(a_{j,n})=s_j\) and \(b_{j,n},r_{j,n}>0\)
  for all \(j\) and \(0\le n\le N_1\);}
\item
  {\(|\Delta_{j,n}|/p_{j,n}\le C_4\tau\) for all \(j\) and
  \(0\le n\le N_1\);}
\item
  {\(\theta_{j,N_1}\le C_1\varepsilon\) for all \(j\); hence
  \(W_{s,N_1}\le Z_{s,N_1}\le C_1\varepsilon\) for \(s=\pm1\);}
\item
  {\(c_5\,p_0\varepsilon^{-2}\le p_{j,N_1}\le C_2'\,p_0\varepsilon^{-2}\);}
\item
  {\(c_6\,c_0\beta\varepsilon^2\le P_{s,N_1}\le C_6'\,c_0\beta\varepsilon^2\)
  for \(s=\pm1\).}
\end{enumerate}

{The constants \(C_1,C_2',C_4,c_5,c_6,C_6'\) do not depend on
\(c_0\); item 1 requires \(c_0\le c_0^{\max}\) for a universal
\(c_0^{\max}\).}
\end{proposition}
\begin{proof} Strong induction on \(n\) with hypothesis
\[\mathcal I_n:\qquad\forall k\le n:\ \ \Psi_k\in\mathcal R,\ \ \|f_k\|_2\le\beta\varepsilon^2,\ \ \operatorname{sign}(a_{j,k})=s_j,\ \ b_{j,k},r_{j,k}>0 .\]

{Base \(n=0\).}
\(\Pi_0=\frac m{\sqrt m}p_0=\sqrt m\,p_0=c_0\beta\varepsilon^4\le6\beta\);
\(b_{j,0}/r_{j,0}=1\);
\(\|f_0\|_2\le\Pi_0/\sqrt2\le\beta\varepsilon^2\); signs and positivity
by \ref{app:1.2}.

{Induction step.} Assume \(\mathcal I_n\) for some \(n<N_1\). Set
\(A_{j,k}:=s_ja_{j,k}\), \(B_{j,k}:=s_j\alpha_{j,k}\),
\(G_{j,k}:=\tfrac12(A+B)\), \(H_{j,k}:=\tfrac12(A-B)\). By \ref{app:1.2},
\(A_{j,0}=\sqrt{p_0}\) and
\(|B_{j,0}|=\sqrt{p_0}|\langle u_j^0,\theta_\star\rangle|\le\tfrac12\sqrt{p_0}\),
so
\[G_{j,0},H_{j,0}\in\bigl[\tfrac14\sqrt{p_0},\ \tfrac34\sqrt{p_0}\bigr].\tag{7.1} \label{app:7.1}\]
Lemma~\ref{lem:phase-one-dynamics} applies at every \(k\le n\) and gives
\[G_{j,k+1}=(1+h)G_{j,k}+\eta E^G_{j,k},\qquad H_{j,k+1}=(1-h)H_{j,k}+\eta E^H_{j,k},\]
\[|E^G_{j,k}|+|E^H_{j,k}|\le C\kappa\iota\,(b_{j,k}+r_{j,k})\le C\kappa\iota\sqrt{p_0}(1+h)^k,\tag{7.2}\label{app:7.2}\]
where we used \ref{app:6.3} together with \(hk\le hN_1\le3L\) (Lemma~\ref{lem:horizon-arithmetic}) and
\(e^{3C\iota L}\le2\), which holds by (S2) below.

\textbf{(a) \(G\) grows.} Dividing by \((1+h)^{k+1}\) and telescoping,
\[
\begin{aligned}
\bigl|(1+h)^{-k}G_{j,k}-G_{j,0}\bigr|
&\le\eta\sum_{l<k}(1+h)^{-(l+1)}|E^G_{j,l}|\\
&\le C\eta\kappa\iota\sqrt{p_0}\,k
=Ch\iota\sqrt{p_0}\,k\\
&\le3C\iota L\sqrt{p_0}
\le\tfrac18\sqrt{p_0}.
\end{aligned}
\]
the last step by (S2). With \ref{app:7.1}, for all \(k\le n+1\),
\[\tfrac18\sqrt{p_0}(1+h)^k\ \le\ G_{j,k}\ \le\ \sqrt{p_0}(1+h)^k.\tag{7.3} \label{app:7.3}\]

\textbf{(b) \(H\) decays.}
\(H_{j,k}=(1-h)^kH_{j,0}+\eta\sum_{l<k}(1-h)^{k-1-l}E^H_{j,l}\), and
\[\sum_{l<k}(1-h)^{k-1-l}(1+h)^l\le(1+h)^{k-1}\sum_{i\ge0}\Bigl(\tfrac{1-h}{1+h}\Bigr)^i=(1+h)^{k-1}\cdot\tfrac{1+h}{2h}=\tfrac{(1+h)^k}{2h},\]
so, using \(\eta\kappa=h\),
\[|H_{j,k}-(1-h)^kH_{j,0}|\le C\iota\sqrt{p_0}(1+h)^k,\qquad|H_{j,k}|\le\tfrac34\sqrt{p_0}(1-h)^k+C\iota\sqrt{p_0}(1+h)^k.\tag{7.4} \label{app:7.4}\]

\textbf{(c) Sign stability and non-vanishing at \(n+1\).} Since
\(H_{j,0}>0\), \ref{app:7.4} gives \(H_{j,k}\ge-C\iota\sqrt{p_0}(1+h)^k\), so
with \ref{app:7.3}
\[b_{j,k}=A_{j,k}=G_{j,k}+H_{j,k}\ \ge\ \bigl(\tfrac18-C\iota\bigr)\sqrt{p_0}(1+h)^k\ \ge\ \tfrac1{16}\sqrt{p_0}(1+h)^k>0\qquad(k\le n+1),\tag{7.5} \label{app:7.5}\]
using (S2). Also
\(r_{j,n+1}\ge r_{j,n}-\eta\|\widehat g^w_{j,n}\|\ge r_{j,n}\bigl(1-C\tau\tfrac{b_{j,n}}{r_{j,n}}\bigr)\ge r_{j,n}(1-2C\tau)>0\)
by (S1).

\textbf{(d) Transverse freeze.} By \ref{app:6.2}, \ref{app:6.3} and
\(\sum_{l<k}(1+h)^l\le(1+h)^k/h\),
\[\|w^\perp_{j,k}\|\le\|w^\perp_{j,0}\|+C\eta\kappa\iota\sum_{l<k}b_{j,l}\le\sqrt{p_0}\bigl(1+C\iota(1+h)^k\bigr)\qquad(k\le n+1).\tag{7.6} \label{app:7.6}\]

\textbf{(e) Upper bound on \(p\); closing the lazy bound and
\(\Pi\le6\beta\).} From \ref{app:7.3}--\ref{app:7.4}, \(b_{j,k}\le2\sqrt{p_0}(1+h)^k\);
and
\[r_{j,k}^2=\alpha_{j,k}^2+\|w^\perp_{j,k}\|^2\le(G_{j,k}+|H_{j,k}|)^2+\|w^\perp_{j,k}\|^2\le C_2\,p_0(1+h)^{2k},\]
so, by Lemma~\ref{lem:horizon-arithmetic},
\[p_{j,k}=b_{j,k}r_{j,k}\le C_2\,p_0(1+h)^{2k}\le4C_2\,p_0\varepsilon^{-2}\qquad(k\le n+1\le N_1),\tag{7.7} \label{app:7.7}\]
with \(C_2\) universal and independent of \(c_0\) (it is built only from
the constants in \ref{app:7.3}, \ref{app:7.4}, \ref{app:7.6}). Hence
\[\Pi_k\le\frac m{\sqrt m}\cdot4C_2p_0\varepsilon^{-2}=4C_2c_0\beta\varepsilon^2,\qquad\|f_k\|_2\le\Pi_k/\sqrt2 .\]
\textbf{Set \(c_0^{\max}:=1/(8C_2)\).} For \(c_0\le c_0^{\max}\) this
gives \(\Pi_{n+1}\le\tfrac12\beta\varepsilon^2\le\beta\le6\beta\) and
\(\|f_{n+1}\|_2\le\tfrac1{2\sqrt2}\beta\varepsilon^2<\beta\varepsilon^2\).

\textbf{(f) Lower bound on \(r\); balance; closing
\(b/r\in[\tfrac12,2]\).} By Lemma~\ref{lem:balance-propagation} (applicable:
\(\Psi_k\in\mathcal R\) for \(k\le n\)), \ref{app:7.7}, and
\(\sum_{k\le n}(1+h)^{2k}\le\frac{(1+h)^{2(n+1)}}{(1+h)^2-1}\le\frac{(1+h)^{2(n+1)}}{2h}\),
together with \(\tau^2/h=2\tau\),
\[|\Delta_{j,n+1}|\ \le\ C_\sharp\tau^2\sum_{k\le n}C_2p_0(1+h)^{2k}\ \le\ C_3'\,\tau\,p_0(1+h)^{2(n+1)},\tag{7.8} \label{app:7.8}\]
with \(C_3'\) universal and \(c_0\)-free. Since \(r^2=b^2-\Delta\),
\ref{app:7.5} and \ref{app:7.8} give
\[r_{j,n+1}^2\ \ge\ \bigl(\tfrac1{256}-C_3'\tau\bigr)p_0(1+h)^{2(n+1)}\ \ge\ \tfrac1{512}\,p_0(1+h)^{2(n+1)}\]
by (S1). Hence \(p_{j,n+1}\ge c_5\,p_0(1+h)^{2(n+1)}\) with
\(c_5:=\tfrac1{16}\cdot\tfrac1{\sqrt{512}}\), and combining with \ref{app:7.8},
\[\frac{|\Delta_{j,n+1}|}{p_{j,n+1}}\ \le\ \frac{C_3'}{c_5}\,\tau\ =:\ C_4\tau\ \le\ \tfrac12\tag{7.9}\]
by (S1). Lemma~\ref{lem:balance-ratio} now gives \(b_{j,n+1}/r_{j,n+1}\in[\tfrac12,2]\).
Together with (c) and (e), \(\mathcal I_{n+1}\) holds.

This completes the induction and proves items 1, 2, 3. Item 5 follows
from \ref{app:7.7}, the lower bound above, and Lemma~\ref{lem:horizon-arithmetic}, with \(C_2':=4C_2\).

\textbf{(g) Alignment at \(N_1\) (item 4).} By \ref{app:7.3}, \ref{app:7.4}, Lemma~\ref{lem:horizon-arithmetic}
and \(\iota\le2\varepsilon\) (from \ref{app:1.4}, \(\iota\le\iota L\)),
\[B_{j,N_1}=G_{j,N_1}-H_{j,N_1}\ \ge\ \tfrac18\tfrac{\sqrt{p_0}}\varepsilon-\tfrac34\sqrt{p_0}\,\varepsilon-C\iota\cdot2\tfrac{\sqrt{p_0}}\varepsilon\ \ge\ \tfrac1{16}\tfrac{\sqrt{p_0}}\varepsilon\ >\ 0,\]
using (S2). By \ref{app:7.6},
\(\|w^\perp_{j,N_1}\|\le\sqrt{p_0}\bigl(1+2C\iota/\varepsilon\bigr)\le C\sqrt{p_0}\).
Since \(s_j\alpha_{j,N_1}=B_{j,N_1}>0\), we have
\(\theta_{j,N_1}\in[0,\tfrac\pi2)\) and
\[\theta_{j,N_1}\le\tan\theta_{j,N_1}=\frac{\|w^\perp_{j,N_1}\|}{B_{j,N_1}}\le16C\varepsilon=:C_1\varepsilon .\]
\(p_0\) cancels, so \(C_1\) is \(c_0\)-free. Finally
\(Z_{s,N_1}=\max_{j\in S_s}\sin\theta_{j,N_1}\le C_1\varepsilon\) and
\(W_{s,N_1}\le Z_{s,N_1}\).

\textbf{(h) Item 6.}
\(P_{s,N_1}=\frac1{\sqrt m}\sum_{j\in S_s}p_{j,N_1}\) with
\(|S_s|=m/2\), and
\(\frac m{2\sqrt m}p_0\varepsilon^{-2}=\tfrac12c_0\beta\varepsilon^2\);
apply item 5. 
\end{proof}

\subsection{Phase II: mass dynamics}

After Phase I, individual directions are already close to their target
poles. The remaining macroscopic quantity is the mass carried by each
sign class. The next lemma shows that, up to controlled errors, each
class follows a logistic recursion with stable target \(\beta\).

For the remainder of Phase II, we work on \(\mathcal E\) and on iterations \(n\in[N_1,N_2]\)
satisfying the \textbf{Phase-II hypotheses}
\[
\begin{aligned}
\textbf{(H1)}&\quad \Theta_n\le\Theta_\star,
&\textbf{(H2)}&\quad P_{s,n}\le2\beta\quad(s=\pm),\\
\textbf{(H3)}&\quad \operatorname{sign}(a_{j,n})=s_j\quad\forall j,
&\textbf{(H4)}&\quad \frac{|\Delta_{j,n}|}{p_{j,n}}\le C_\Delta\tau L\le\varepsilon\quad\forall j.
\end{aligned}
\tag{H}
\]
(H) implies \(\Pi_n\le4\beta\) and, by Lemma~\ref{lem:balance-ratio},
\(b_{j,n}/r_{j,n}\in[1-\varepsilon,1+\varepsilon]\subseteq[\tfrac12,2]\);
hence \(\Psi_n\in\mathcal R\) and \ref{app:4.1} is available at \(n\). Define
\[F_{j,n}:=s_j\,\mathbb E\bigl[(y-f_n)\sigma(u_{j,n}^\top X)\bigr],\qquad\delta_{s,n}:=\frac{\beta-P_{s,n}}{2\sqrt m}.\]

\begin{lemma}[Phase-II mass dynamics]\label{lem:phase-two-mass}
{Under (H) at iteration \(n\) (indices suppressed):}
\[F_j=\frac{\beta\cos\theta_j}2-\frac1{\sqrt m}\sum_{i=1}^ms_is_jp_iK_{ij},\qquad\Bigl|F_j-\frac{\beta-P_{s_j}}2\Bigr|\le C\beta\Theta_\star^2;\tag{8.1}\label{app:8.1}\]
{sign stability holds at \(n+1\), with \(b_j^+\ge b_j(1-C\tau)>0\)
and \(r_j^+\ge r_j(1-C\tau)>0\); and}
\[p_{j,n+1}=p_{j,n}+\frac{2\eta}{\sqrt m}F_{j,n}\,p_{j,n}+\eta\,p_{j,n}\,\omega_{j,n},\qquad|\omega_{j,n}|\le C\frac\beta{\sqrt m}\bigl(\varepsilon^2+\rho_B+\tau\bigr);\tag{8.2}\label{app:8.2}\]
\[P_{s,n+1}=P_{s,n}+\frac\eta{\sqrt m}P_{s,n}(\beta-P_{s,n})+\eta P_{s,n}\varrho_{s,n},\qquad|\varrho_{s,n}|\le C_5\frac\beta{\sqrt m}\zeta .\tag{8.3}\label{app:8.3}\]
\end{lemma}
\begin{proof} {\ref{app:8.1}.} By Lemma~\ref{lem:gaussian}(i),
\(s_j\mathbb E[y\,\sigma(u_j^\top X)]=s_j\frac\beta2\theta_\star^\top u_j=\frac\beta2\cos\theta_j\);
by \ref{app:1.3} and Lemma~\ref{lem:gaussian}(iv),
\(s_j\mathbb E[f_n\sigma(u_j^\top X)]=\frac1{\sqrt m}\sum_is_is_jp_iK_{ij}\).

For \(i\) with \(s_i=s_j\): both \(u_i,u_j\) lie within angle
\(\Theta_\star\) of \(s_j\theta_\star\), so
\(\psi_{ij}\le\theta_i+\theta_j\le2\Theta_\star\) by the triangle
inequality for the angular metric; Lemma~\ref{lem:kernel-bounds} gives
\(|K_{ij}-\tfrac12|\le\Theta_\star^2\), hence
\[\Bigl|\frac1{\sqrt m}\sum_{i\in S_{s_j}}p_iK_{ij}-\frac{P_{s_j}}2\Bigr|\le\Theta_\star^2\,P_{s_j}\le2\beta\Theta_\star^2 .\]
For \(i\) with \(s_i=-s_j\):
\(\psi_{ij}\ge\pi-\theta_i-\theta_j\ge\pi-2\Theta_\star\), so
\(0\le K_{ij}\le\frac{(2\Theta_\star)^3}{6\pi}\le\Theta_\star^3\), and
the whole opposite-class contribution is
\(\le\Theta_\star^3P_{-s_j}\le2\beta\Theta_\star^3\le2\beta\Theta_\star^2\).
Finally \(|\cos\theta_j-1|\le\Theta_\star^2/2\). Summing gives \ref{app:8.1}.

\textbf{Step 0 (sign stability) .}
Write \(\widehat V^a_j:=-\widehat g^a_j\),
\(\widehat V^w_j:=-\widehat g^w_j\). By Lemma~\ref{lem:population-gradient-bounds} and \ref{app:4.1},
\(\eta|\widehat V^a_j|\le C\tau r_j\le2C\tau b_j\) and
\(\eta\|\widehat V^w_j\|\le C\tau b_j\le2C\tau r_j\). By (S1)
\(2C\tau<1\), so
\[s_ja_j^+=s_j(a_j+\eta\widehat V^a_j)=b_j+\eta s_j\widehat V^a_j\ \ge\ b_j(1-2C\tau)>0,\qquad r_j^+\ge r_j(1-2C\tau)>0 .\]
Hence \(\operatorname{sign}(a_j^+)=s_j\) and
\(b_j^+=s_ja_j^+=b_j+\eta s_j\widehat V^a_j\) exactly.

\textbf{Step 1 (population radial velocities, exact).} With
\(V^a_j,V^w_j\) the population velocities,
\[
\begin{aligned}
s_jV^a_j
&=\frac{s_j}{\sqrt m}\mathbb E[(y-f)\sigma(w_j^\top X)]
 =\frac{r_j}{\sqrt m}F_j,\\
u_j^\top V^w_j
&=\frac{a_j}{\sqrt m}\mathbb E[(y-f)\mathbf 1_{\{u_j^\top X>0\}}u_j^\top X]
 =\frac{a_js_j}{\sqrt m}F_j
 =\frac{b_j}{\sqrt m}F_j.
\end{aligned}
\tag{8.4} \label{app:8.4}
\]
using \(\mathbf 1_{\{t>0\}}t=\sigma(t)\) and \(a_js_j=b_j\).

\textbf{Step 2 (one-step expansions).} With
\(\varsigma^a_j:=s_j(\widehat V^a_j-V^a_j)\),
\(|\varsigma^a_j|\le\frac{\beta r_j}{\sqrt m}\rho_B\):
\[b_j^+=b_j+\frac{\eta r_j}{\sqrt m}F_j+\eta\varsigma^a_j .\] By Step 0,
\(\eta\|\widehat V^w_j\|\le2C\tau r_j\le r_j/5\) by (S1), so Lemma~\ref{lem:norm-remainder}
gives \(r_j^+=r_j+\eta u_j^\top\widehat V^w_j+\vartheta_j\) with
\(|\vartheta_j|\le3\eta^2\|\widehat V^w_j\|^2/r_j\le C\tau^2r_j\). With
\(\varsigma^w_j:=u_j^\top(\widehat V^w_j-V^w_j)\),
\(|\varsigma^w_j|\le\frac{\beta b_j}{\sqrt m}\rho_B\):
\[r_j^+=r_j+\frac{\eta b_j}{\sqrt m}F_j+\eta\varsigma^w_j+\vartheta_j .\]

\textbf{Step 3 (product).} Multiplying,
\[p_j^+=b_j^+r_j^+=p_j+\eta\frac{r_j^2+b_j^2}{\sqrt m}F_j+\eta\bigl(r_j\varsigma^a_j+b_j\varsigma^w_j\bigr)+b_j\vartheta_j+\Xi_j,\]
where \(\Xi_j\) is the sum of products of two increments, each of size
\(\le C\tau\max(b_j,r_j)\), so \(|\Xi_j|\le C\tau^2p_j\). By Lemma~\ref{lem:balance-ratio}
with \(\varrho=\varepsilon\),
\[\frac{r_j^2+b_j^2}{\sqrt m}F_j=\frac{p_j}{\sqrt m}\Bigl(\frac{r_j}{b_j}+\frac{b_j}{r_j}\Bigr)F_j=\frac{2p_j}{\sqrt m}F_j+\frac{p_j}{\sqrt m}O(\varepsilon^2)F_j,\]
and \(|F_j|\le C\beta\) by \ref{app:8.1} and (H2). Also
\(|r_j\varsigma^a_j+b_j\varsigma^w_j|\le\frac{C\beta\rho_B}{\sqrt m}p_j\)
and \(|b_j\vartheta_j|\le C\tau^2p_j=C\eta\frac\beta{\sqrt m}\tau p_j\).
Collecting gives \ref{app:8.2}.

{\ref{app:8.3}.} Sum \ref{app:8.2} over \(j\in S_s\) and divide by \(\sqrt m\);
insert \ref{app:8.1}. The \(O(\beta\Theta_\star^2)\) error contributes
\(\eta P_s\cdot C\frac{\beta\Theta_\star^2}{\sqrt m}\), and
\(\max_j|\omega_j|\le C\frac\beta{\sqrt m}(\varepsilon^2+\rho_B+\tau)\);
since \(\varepsilon^2\le\Theta_\star^2\) (Lemma~\ref{lem:smallness-budget} gives
\(\Theta_\star\ge A\varepsilon\ge\varepsilon\)), both are
\(\le C\frac\beta{\sqrt m}\zeta\). 
\end{proof}

\begin{remark}[Role of the angular error]
The imbalance enters \ref{app:8.2} only
{quadratically}, since \(x+x^{-1}-2=(x-1)^2/x\); in the angular
transverse recursion it enters {linearly}. This asymmetry is why the mass
dynamics tolerates a much larger imbalance than the angular dynamics.
\end{remark}

\subsection{Logistic comparison}

\begin{lemma}[Logistic comparison]\label{lem:logistic-comparison}
{Assume (H) at every iteration of \([N_1,n]\) with \(n\le N_2-1\),
and \(\zeta\le c_\zeta\) (see (S3)). Assume also, for a fixed universal
\(c_{\rm in}>0\), that for each \(s\in\{\pm1\}\)
\[
c_{\rm in}c_0\beta\varepsilon^2\le P_{s,N_1}\le\frac\beta2,
\qquad
\log\frac1{2c_{\rm in}c_0}\le L.
\]
Put \(q:=\eta/\sqrt m\) (so
\(q\beta=\tau\)), \(T_B(p):=p+qp(B-p)\), and
\(\beta_\pm:=\beta(1\pm C_5\zeta)\) with \(C_5\) from \ref{app:8.3}. Then for
each \(s\in\{\pm1\}\):} 1. {(sandwich)
\(P^-_k\le P_{s,k}\le P^+_k\) for \(N_1\le k\le n+1\), where
\(P^\pm_{k+1}=T_{\beta_\pm}(P^\pm_k)\) and \(P^\pm_{N_1}=P_{s,N_1}\);}
2. {(no overshoot) \(P_{s,k}\le\beta_+\le\tfrac32\beta\) for
\(N_1\le k\le n+1\);} 3. {(time to half) \(P^-_k\ge\beta/2\) for
some \(k\le N_1+C_6\tau^{-1}L\) with \(C_6:=25\), provided \(n+1\) is at
least that index;} 4. {(endpoint accuracy) if \(C_L\ge C_7:=C_6+2\)
and \(n+1=N_2\), then
\(|P_{s,N_2}-\beta|\le\beta(\bar\varepsilon+C_5\zeta)\le C\beta\bar\varepsilon\).}
\end{lemma}
\begin{proof} By \ref{app:8.3}, \(P_{k+1}=T_{\kappa_k}(P_k)\) with
\(\kappa_k:=\beta+\sqrt m\varrho_{s,k}\in[\beta_-,\beta_+]\).

{(1)--(2).} For \(p\in[0,3\beta]\),
\(\partial_pT_B(p)=1+q(B-2p)\ge1-6q\beta=1-6\tau\ge\tfrac12\) by (S1);
so \(T_B\) is strictly increasing in \(p\) on \([0,3\beta]\), and it is
increasing in \(B\) for \(p>0\). If \(0\le P\le\beta_+\) then
\(\beta_+-T_{\beta_+}(P)=(\beta_+-P)(1-qP)\ge0\) since \(qP\le2\tau<1\);
so \([0,\beta_+]\) is forward invariant for \(T_{\beta_+}\). As
\(P^+_{N_1}=P_{s,N_1}\le\beta/2\le\beta_+\), we get
\(P^+_k\le\beta_+\) for all \(k\). Now induct: if
\(P^-_k\le P_k\le P^+_k\) with all three in \([0,3\beta]\), then by the
two monotonicities
\(P_{k+1}=T_{\kappa_k}(P_k)\in[T_{\beta_-}(P^-_k),T_{\beta_+}(P^+_k)]=[P^-_{k+1},P^+_{k+1}]\),
and \(P^+_{k+1}\le\beta_+\le\tfrac32\beta\), \(P^-_{k+1}\ge0\). (Here
\(\beta_+\le\tfrac32\beta\) by (S3).)

{(3).} While \(P^-_k\le\beta/2\), and \(\beta_-\ge\tfrac34\beta\)
by (S3),
\[P^-_{k+1}\ge P^-_k\bigl(1+q(\beta_--\tfrac\beta2)\bigr)\ge P^-_k\bigl(1+\tfrac\tau4\bigr).\]
From \(P^-_{N_1}=P_{s,N_1}\ge c_{\rm in}c_0\beta\varepsilon^2\), the number of
steps needed to exceed \(\beta/2\) is at most
\[\frac{\log\bigl(\tfrac1{2c_{\rm in}c_0\varepsilon^2}\bigr)}{\log(1+\tau/4)}\le\frac8\tau\Bigl(2L+\log\tfrac1{2c_{\rm in}c_0}\Bigr)\le\frac{24L}\tau,\]
using \(\log(1+x)\ge x/2\) on \([0,1]\) and
the displayed assumption. So \(C_6=25\)
suffices.

{(4).} Let \(k_0\) be the first index \(\ge N_1\) with
\(P^-_{k_0}\ge\beta/2\); by (3), \(k_0\le N_1+C_6\tau^{-1}L\). If $k_0=N_1$ then $P^-_{k_0}=\beta/2\le\beta_-$. Otherwise $P^-_{k_0-1}<\beta/2$ and \(T_{\beta_-}\) is increasing in \(p\),
\[P^-_{k_0}=T_{\beta_-}(P^-_{k_0-1})\le T_{\beta_-}(\tfrac\beta2)=\tfrac\beta2+q\tfrac\beta2\bigl(\beta_--\tfrac\beta2\bigr)\le\tfrac\beta2(1+\tau)\le\beta_-,\]
by (S1) and (S3); hence \(P^-_{k_0}\in[\tfrac\beta2,\beta_-]\). This
interval is forward invariant for \(T_{\beta_-}\): for \(P\) in it,
\(T_{\beta_-}(P)\ge P\ge\tfrac\beta2\) (since \(\beta_-\ge P\)) and
\(T_{\beta_-}(P)\le\beta_-\) (by the identity in (2)). Put
\(D_k:=\beta_--P^-_k\ge0\) for \(k\ge k_0\); then
\[D_{k+1}=\beta_--P^-_k-qP^-_k(\beta_--P^-_k)=D_k\bigl(1-qP^-_k\bigr)\le D_k\bigl(1-\tfrac\tau2\bigr),\]
so \(D_k\le\beta e^{-\tau(k-k_0)/2}\), and
\(D_k\le\beta\bar\varepsilon\) as soon as
\(k-k_0\ge\frac2\tau\log\frac1{\bar\varepsilon}\). Since
\(\log\frac1{\bar\varepsilon}=L-\log L\le L\), it suffices that
\(k-k_0\ge2\tau^{-1}L\), which holds at \(k=N_2\) when \(C_L\ge C_6+2\).
Therefore
\[\beta(1-C_5\zeta)-\beta\bar\varepsilon\ \le\ P^-_{N_2}\ \le\ P_{s,N_2}\ \le\ \beta(1+C_5\zeta),\]
i.e.~\(|P_{s,N_2}-\beta|\le\beta(\bar\varepsilon+C_5\zeta)\), and
\(\zeta\le3\bar\varepsilon\) by \ref{app:1.4}. 
\end{proof}

\begin{corollary}[No mass collapse]\label{cor:no-mass-collapse}
{Under the same
hypotheses, \(p_{j,k}\le e^{C(C_L+1)\zeta L}p_{j,n+1}\le2\,p_{j,n+1}\)
for \(N_1\le k\le n+1\).}
\end{corollary}
\begin{proof} By \ref{app:8.1}--\ref{app:8.2} and part (2) of Lemma~\ref{lem:logistic-comparison},
\(\beta-P_{s,k}\ge-C_5\beta\zeta\), so
\(p_{j,k+1}\ge p_{j,k}(1-C\tau\zeta)\). By (S3), \(C\tau\zeta\le1/2\).
Hence, with \(x=C\tau\zeta\), the elementary inequality
\((1-x)^{-1}\le e^{2x}\) gives
\[p_{j,k}\le p_{j,n+1}\exp\!\bigl(C\tau\zeta\,(n+1-k)\bigr).\] There are
at most \(N_2-N_1\le C_L\tau^{-1}L+1\) Phase-II steps, so after
enlarging the universal constant \(C\) we obtain
\(p_{j,k}\le e^{C(C_L+1)\zeta L}p_{j,n+1}\). The final bound by
\(2p_{j,n+1}\) is (S5). 
\end{proof}

\subsection{Transverse recursion}

Controlling only the largest angle would discard the cancellation
created by same-class interactions. We therefore first track the
weighted mean-square transverse error \(V_s\) (equivalently
\(W_s=\sqrt{V_s}\)), where the gate-overlap matrix is positive
semidefinite, and then use this averaged control to bound the worst
neuron through \(Z_s\). The following exact recursion is the key input.

Fix \(s\) and \(j\in S_s\); suppress \(n\); write
\(P^\perp:=I-\theta_\star\theta_\star^\top\).

\begin{lemma}[Transverse recursion]\label{lem:transverse-recursion}
{Under (H) at iteration \(n\), on \(\mathcal E\):}
\[
\begin{aligned}
\xi_j^+-\xi_j&=\eta v_j+\eta\tilde e_j,\\
v_j&=-\Lambda_j\xi_j-\frac1m\sum_{i\in S_s}p_iq_{ij}\xi_i,\\
\Lambda_j&=\frac{\beta\cos\theta_j}{2\sqrt m}
-\frac1m\sum_{i\in S_s}p_iq_{ij}\langle u_i,u_j\rangle .
\end{aligned}
\tag{10.1} \label{app:10.1}
\]
\[\|\tilde e_j\|\le e_\star:=C_9\frac\beta{\sqrt m}\zeta;\tag{10.2}\label{app:10.2}\]
\[\delta_s-\frac{\beta\Theta_\star^2}{4\sqrt m}\ \le\ \Lambda_j\ \le\ \frac\beta{2\sqrt m};\tag{10.3} \label{app:10.3}\]
\[\|v_j\|\le C\frac\beta{\sqrt m}\bigl(\|\xi_j\|+W_s\bigr),\qquad\frac1m\sum_{i\in S_s}p_iq_{ij}\|\xi_i\|\le\frac{P_s}{2\sqrt m}W_s;\tag{10.4} \label{app:10.4}\]
\[\sum_{i,j\in S_s}p_ip_j\,q_{ij}\,\langle\xi_i,\xi_j\rangle\ \ge\ 0.\tag{10.5} \label{app:10.5}\]
\end{lemma}
\begin{proof}
{\textbf{Step 1 (normalization, explicit remainder).}
\(w_j^+=w_j+\eta\widehat V^w_j\) with
\(\widehat V^w_j=-\widehat g^w_j\). Put
\(\varpi_j:=\eta\widehat V^w_j/r_j\); by Lemma~\ref{lem:population-gradient-bounds}, \ref{app:4.1} and
\(b_j\le2r_j\), \(\|\varpi_j\|\le C\tau\le\tfrac15\) by (S1). Lemma~\ref{lem:normalization-remainder}
gives
\[u_j^+-u_j=(I-u_ju_j^\top)\varpi_j+E^{\rm norm}_j,\qquad\|E^{\rm norm}_j\|\le14\|\varpi_j\|^2\le C\eta\tfrac\beta{\sqrt m}\tau .\]
Splitting \(\widehat V^w_j=V^w_j+(\widehat V^w_j-V^w_j)\) with
\(\|\widehat V^w_j-V^w_j\|\le\frac{\beta b_j}{\sqrt m}\rho_B\le\frac{2\beta r_j}{\sqrt m}\rho_B\),
\[u_j^+-u_j=\frac\eta{r_j}(I-u_ju_j^\top)V^w_j+\eta\varrho_j,\qquad\|\varrho_j\|\le C\frac\beta{\sqrt m}(\rho_B+\tau).\tag{10.6}\label{app:10.6}\]

\textbf{Step 2 (population direction field).}
\(V^w_j=\frac{a_j}{\sqrt m}B_j\) with
\(B_j:=\frac\beta2\theta_\star-\mathbb E[f_n\mathbf 1_{\{u_j^\top X>0\}}X]\),
by Lemma~\ref{lem:gaussian}(ii). By \ref{app:1.3} and Lemma~\ref{lem:gaussian}(iii),
\[\mathbb E\bigl[f_n\mathbf 1_{\{u_j^\top X>0\}}X\bigr]=\frac1{\sqrt m}\sum_{i=1}^ms_ip_i\Bigl(q_{ij}u_i+\frac{\sin\psi_{ij}}{2\pi}u_j\Bigr).\]
The second family of terms is parallel to \(u_j\) and is therefore
projected to zero by \(I-u_ju_j^\top\):
\[(I-u_ju_j^\top)V^w_j=\frac{a_j}{\sqrt m}(I-u_ju_j^\top)\Bigl[\frac\beta2\theta_\star-\frac1{\sqrt m}\sum_is_ip_iq_{ij}u_i\Bigr].\]

\textbf{Step 3 (imbalance factor).}
\(\frac{a_j}{r_j}=s_j\frac{b_j}{r_j}=s_j(1+\epsilon_j)\) with
\(|\epsilon_j|\le\varepsilon\), by (H4) and Lemma~\ref{lem:balance-ratio}.

\textbf{Step 4 (projection onto \(\theta_\star^\perp\)).} Since
\(\langle u_j,\theta_\star\rangle=s_j\cos\theta_j\),
\[P^\perp(I-u_ju_j^\top)\theta_\star=P^\perp\bigl(\theta_\star-s_j\cos\theta_j\,u_j\bigr)=-s_j\cos\theta_j\,\xi_j,\qquad P^\perp(I-u_ju_j^\top)u_i=\xi_i-\langle u_i,u_j\rangle\xi_j .\]
Hence, with \(\sigma_{ij}:=s_is_j\) (the outer \(s_j\) from Step 3
combining with \(s_i\) inside the sum),
\[\xi_j^+-\xi_j=\eta(1+\epsilon_j)\Bigl[-\frac{\beta\cos\theta_j}{2\sqrt m}\xi_j-\frac1m\sum_{i=1}^m\sigma_{ij}p_iq_{ij}\bigl(\xi_i-\langle u_i,u_j\rangle\xi_j\bigr)\Bigr]+\eta P^\perp\varrho_j .\tag{10.7} \label{app:10.7}\]

\textbf{Step 5 (class split).} The terms with \(i\in S_s\)
(\(\sigma_{ij}=1\)) are exactly \(v_j\) of \ref{app:10.1}. For \(i\notin S_s\):
\(\pi-\psi_{ij}\le\theta_i+\theta_j\le2\Theta_\star\), so
\(q_{ij}\le\Theta_\star/\pi\) by Lemma~\ref{lem:kernel-bounds}, and using
\(\|\xi_i\|,\|\xi_j\|\le\Theta_\star\) and
\(|\langle u_i,u_j\rangle|\le1\) the opposite-class vector has norm at
most
\[\frac1m\sum_{i\notin S_s}p_i\frac{\Theta_\star}\pi\bigl(\|\xi_i\|+\|\xi_j\|\bigr)\le\frac{2\Theta_\star^2}\pi\cdot\frac{D_{-s}}m=\frac{2\Theta_\star^2}\pi\cdot\frac{P_{-s}}{\sqrt m}\le C\frac{\beta\Theta_\star^2}{\sqrt m}.\]

\textbf{Step 6 (proof of \ref{app:10.4}).} For \(i,j\in S_s\) we have
\(\psi_{ij}\le2\Theta_\star\le\tfrac15\), so
\(\langle u_i,u_j\rangle=\cos\psi_{ij}\in[0,1]\), hence
\(0\le q_{ij}\langle u_i,u_j\rangle\le\tfrac12\) and
\(|\Lambda_j|\le\frac\beta{2\sqrt m}+\frac{P_s}{2\sqrt m}\le C\frac\beta{\sqrt m}\).
By weighted Cauchy--Schwarz,
\[\frac1m\sum_{i\in S_s}p_iq_{ij}\|\xi_i\|\le\frac1{2m}\sum_{i\in S_s}p_i\|\xi_i\|\le\frac{D_s}{2m}\Bigl(\frac1{D_s}\sum_{i\in S_s}p_i\|\xi_i\|^2\Bigr)^{1/2}=\frac{P_s}{2\sqrt m}W_s .\]
Combining the two displays gives
\(\|v_j\|\le C\frac\beta{\sqrt m}(\|\xi_j\|+W_s)\).

\textbf{Step 7 (proof of \ref{app:10.2}).} From \ref{app:10.7},
\(\tilde e_j=\epsilon_jv_j+(1+\epsilon_j)\cdot(\text{opposite-class term})+P^\perp\varrho_j\).
By \ref{app:10.4}, \(\|\xi_j\|,W_s\le\Theta_\star\), and
\(\varepsilon\le\Theta_\star\),
\[|\epsilon_j|\,\|v_j\|\le\varepsilon\cdot C\frac\beta{\sqrt m}\cdot2\Theta_\star\le C\frac{\beta\Theta_\star^2}{\sqrt m}.\]
Together with Step 5 and \ref{app:10.6},
\(\|\tilde e_j\|\le C_9\frac\beta{\sqrt m}(\Theta_\star^2+\rho_B+\tau)=e_\star\).

\textbf{Step 8 (proof of \ref{app:10.3}).} Since
\(\frac{P_s}{2\sqrt m}=\frac{D_s}{2m}=\frac1m\sum_{i\in S_s}p_i\cdot\frac12\),
\[\Lambda_j-\delta_s=\frac{\beta(\cos\theta_j-1)}{2\sqrt m}+\frac1m\sum_{i\in S_s}p_i\Bigl(\tfrac12-q_{ij}\langle u_i,u_j\rangle\Bigr).\tag{10.8}\]
The second sum is \(\ge0\) by Step 6; the first is
\(\ge-\frac{\beta\Theta_\star^2}{4\sqrt m}\) since
\(1-\cos\theta_j\le\theta_j^2/2\le\Theta_\star^2/2\). This gives the
lower bound. The upper bound follows from \ref{app:10.1} by discarding the
subtracted sum, which is nonnegative:
\(\Lambda_j\le\frac{\beta\cos\theta_j}{2\sqrt m}\le\frac\beta{2\sqrt m}\).

\textbf{Step 9 (proof of \ref{app:10.5}).)}
\(q_{ij}=\bigl\langle\mathbf 1_{\{u_i^\top\cdot>0\}},\mathbf 1_{\{u_j^\top\cdot>0\}}\bigr\rangle_{L^2(\gamma_d)}\),
so \((q_{ij})_{i,j\in S_s}\) is a Gram matrix, hence positive
semidefinite. Applying this coordinatewise to
\(z_i:=p_i\xi_i\in\mathbb R^d\),
\[\sum_{i,j\in S_s}q_{ij}\langle z_i,z_j\rangle=\sum_{k=1}^d\ \sum_{i,j\in S_s}q_{ij}z_i^{(k)}z_j^{(k)}\ \ge\ 0 .\]}
\end{proof}

\begin{remark}[Sign of the transverse coefficient]
\label{rem:transverse-sign}
In (10.8), the first term is nonpositive and bounded in
magnitude by \(\beta\Theta_\star^2/(4\sqrt m)\), while the
second term is nonnegative and can be of order
\(\beta\Theta_\star/\sqrt m\). Thus we cannot conclude that
\(\Lambda_j-\delta_s\) is nonnegative or that its absolute
value is \(O(\beta\Theta_\star^2/\sqrt m)\).
The proof uses only the lower bound
\[
\Lambda_j\ge
\delta_s-\frac{\beta\Theta_\star^2}{4\sqrt m},
\]
established in \eqref{app:10.3}.
\end{remark}

\subsection{Angular recursions}

\begin{lemma}[Angular recursions]\label{lem:angular-recursions}
{Assume (H) at iteration \(n\) and \(\zeta\le c_\zeta\). There is a
universal \(C_D\ge1\) (chosen after \(C_5,C_8,C_9\)) such that, with}
\[a_{s,n}:=\delta_{s,n}-C_D\frac{\beta\zeta}{\sqrt m},\qquad b_\star:=C_D\frac\beta{\sqrt m}\zeta,\]
{one has \(\eta|a_{s,n}|\le C\tau\le\tfrac12\) and}
\[W_{s,n+1}\le\bigl(1-\eta a_{s,n}\bigr)W_{s,n}+\eta b_\star,\tag{11.1} \label{app:11.1}\]
\[Z_{s,n+1}\le\bigl(1-\eta a_{s,n}\bigr)Z_{s,n}+\eta\frac{P_{s,n}}{2\sqrt m}W_{s,n}+\eta b_\star .\tag{11.2} \label{app:11.2}\]
\end{lemma}
\begin{proof} Write \(N_s:=\sum_{j\in S_s}p_j\|\xi_j\|^2\),
\(D_s:=\sum_{j\in S_s}p_j\), \(V_s=N_s/D_s\). By Lemma~\ref{lem:phase-two-mass},
\(p_j^+=p_j(1+\eta\lambda_j)\) with
\[\lambda_j=\frac2{\sqrt m}F_j+\omega_j=\lambda_s+\mu_j,\qquad\lambda_s:=\frac{\beta-P_s}{\sqrt m},\qquad|\mu_j|\le C\frac{\beta\zeta}{\sqrt m},\qquad|\lambda_j|\le C\frac\beta{\sqrt m},\tag{11.3}\]
and by Lemma~\ref{lem:transverse-recursion}, \(\xi_j^+=\xi_j+\eta v_j+\eta\tilde e_j\) with
\(\|\tilde e_j\|\le e_\star\).

\textbf{Step 1 (expansion of \(N_s\)).}
\[N_s^+-N_s=\underbrace{2\eta\sum_jp_j\langle\xi_j,v_j\rangle}_{\rm(I)}+\underbrace{2\eta\sum_jp_j\langle\xi_j,\tilde e_j\rangle}_{\rm(II)}+\underbrace{\eta\sum_jp_j\lambda_j\|\xi_j\|^2}_{\rm(III)}+\mathcal E_N,\]
\[\mathcal E_N=\eta^2\sum_jp_j(1+\eta\lambda_j)\|v_j+\tilde e_j\|^2+2\eta^2\sum_jp_j\lambda_j\langle\xi_j,v_j+\tilde e_j\rangle,\]
all sums over \(j\in S_s\).

\textbf{Step 2 (second-order terms).} By
\ref{app:10.4} and \(\sum_jp_j(\|\xi_j\|^2+W_s^2)=2D_sV_s\),
\[\sum_jp_j\|v_j\|^2\le C\frac{\beta^2}m\sum_jp_j(\|\xi_j\|+W_s)^2\le2C\frac{\beta^2}m\sum_jp_j\bigl(\|\xi_j\|^2+W_s^2\bigr)=4C\frac{\beta^2}mD_sV_s .\]
Hence, using \(1+\eta\lambda_j\le2\) and
\(\|x+y\|^2\le2\|x\|^2+2\|y\|^2\),
\[\eta^2\sum_jp_j(1+\eta\lambda_j)\|v_j+\tilde e_j\|^2\le C\eta^2\frac{\beta^2}mD_sV_s+4\eta^2D_se_\star^2 .\]
For the cross term, \(|\lambda_j|\le C\beta/\sqrt m\),
\(\sum_jp_j\|\xi_j\|(\|\xi_j\|+W_s)\le2D_sV_s\) (Cauchy--Schwarz for the
second piece) and \(\sum_jp_j\|\xi_j\|\le D_sW_s\), so
\[2\eta^2\sum_jp_j|\lambda_j|\,\|\xi_j\|\bigl(\|v_j\|+\|\tilde e_j\|\bigr)\le C\eta^2\frac{\beta^2}mD_sV_s+C\eta^2\frac\beta{\sqrt m}e_\star D_sW_s .\]
Since \(\eta^2\beta^2/m=\eta\tau\frac\beta{\sqrt m}\) and
\(\eta^2\beta/\sqrt m=\eta\tau\),
\[|\mathcal E_N|\ \le\ C_8\,\eta\tau\frac\beta{\sqrt m}D_sV_s\ +\ C_8\,\eta\tau\,e_\star D_sW_s\ +\ 4\eta^2D_se_\star^2 .\tag{11.4}\]
The first term is \textbf{multiplicative in \(V_s\)} (it perturbs the
rate by \(C_8\tau\beta/\sqrt m\le C_8\zeta\beta/\sqrt m\)); the second
is \(\le\eta e_\star D_sW_s\) once \(C_8\tau\le1\), which is (S1). The second-order contribution is multiplicative in the current squared
angular error, apart from terms controlled by the small forcing. It can
therefore be absorbed in the recursion below.

\textbf{Step 3 (quotient rule).} Exactly,
\[V_s^+-V_s=\rho\cdot\frac1{D_s}\Bigl[(N_s^+-N_s)-V_s(D_s^+-D_s)\Bigr],\qquad\rho:=\frac{D_s}{D_s^+},\quad|\rho-1|\le C\tau,\]
where \(D_s^+=D_s+\eta\sum_jp_j\lambda_j\). Since
\(\sum_jp_j(\|\xi_j\|^2-V_s)=0\) annihilates the class-common part
\(\lambda_s\),
\[\text{(III)}-V_s(D_s^+-D_s)=\eta\sum_jp_j\mu_j\bigl(\|\xi_j\|^2-V_s\bigr),\qquad\Bigl|\frac1{D_s}\bigl[\cdots\bigr]\Bigr|\le2\eta\max_j|\mu_j|\,V_s\le C\eta\frac{\beta\zeta}{\sqrt m}V_s.\tag{11.5}\]

\textbf{Step 4 (drift term).} By \ref{app:10.1}, \ref{app:10.3} and \ref{app:10.5},
\[\frac{\rm(I)}{D_s}=-\frac{2\eta}{D_s}\sum_jp_j\Lambda_j\|\xi_j\|^2-\frac{2\eta}{mD_s}\sum_{i,j\in S_s}p_ip_jq_{ij}\langle\xi_i,\xi_j\rangle\ \le\ -2\eta\Bigl(\delta_s-\frac{\beta\Theta_\star^2}{4\sqrt m}\Bigr)V_s.\tag{11.6}\]

\textbf{Step 5 (forcing term).}
\(\dfrac{|\rm(II)|}{D_s}\le2\eta e_\star\dfrac1{D_s}\sum_jp_j\|\xi_j\|\le2\eta e_\star W_s\).

\textbf{Step 6 (assembling).} By Steps 2--5, with \(C_8\tau\le1\),
\[
\begin{aligned}
\frac1{D_s}\bigl[(N_s^+-N_s)-V_s(D_s^+-D_s)\bigr]
&\le-2\eta a'V_s+3\eta e_\star W_s+4\eta^2e_\star^2,\\
a'&:=\delta_s-\frac\beta{\sqrt m}
\Bigl(\frac{\Theta_\star^2}4+C\zeta+\frac{C_8\tau}2\Bigr).
\end{aligned}
\]
Multiplying by \(\rho\): the nonnegative terms gain a factor
\(\le1+C\tau\le2\), while
\[-2\eta\rho a'V_s=-2\eta a'V_s+2\eta(1-\rho)a'V_s\le-2\eta\Bigl(a'-C\tau\tfrac\beta{\sqrt m}\Bigr)V_s,\]
using \(|1-\rho|\le C\tau\) and \(|a'|\le C\beta/\sqrt m\). Hence, since
\(\tau\le\zeta\) and \(\Theta_\star^2\le\zeta\), for \(C_D\) large
enough,
\[V_s^+\ \le\ V_s-2\eta a_sV_s+6\eta e_\star W_s+8\eta^2e_\star^2 .\]
Apply Lemma~\ref{lem:v-to-w} with \(x=W_s\), \(x_+=W_s^+\), \(b:=3e_\star\) (so
\(6\eta e_\star W_s=2\eta bW_s\) and \(8\eta^2e_\star^2\le4\eta^2b^2\))
and \(\eta|a_s|\le C\tau\le\tfrac12\) by (S1):
\(W_s^+\le(1-\eta a_s)W_s+9\eta e_\star\). Finally enlarge \(C_D\) so
that \(9e_\star\le b_\star\), i.e.~\(C_D\ge9C_9\); this yields \ref{app:11.1}.

\textbf{Step 7 (the max, valid also when \(\Lambda_j<0\)).} Fix
\(j\in S_s\). By \ref{app:10.3},
\(\eta\Lambda_j\le\eta\frac\beta{2\sqrt m}=\tau/2<1\), so
\(1-\eta\Lambda_j>0\) and
\(\|\xi_j-\eta\Lambda_j\xi_j\|=(1-\eta\Lambda_j)\|\xi_j\|\). (No lower
bound on \(\eta\Lambda_j\) is claimed: \(\Lambda_j\) may be negative
when \(P_s>\beta\).) Since
\(\Lambda_j\ge\delta_s-\frac{\beta\Theta_\star^2}{4\sqrt m}\ge a_s\) and
\(1-\eta a_s\ge0\),
\[(1-\eta\Lambda_j)\|\xi_j\|\le(1-\eta a_s)\|\xi_j\|\le(1-\eta a_s)Z_s .\]
By the triangle inequality applied to \ref{app:10.1}, together with \ref{app:10.4} and
\ref{app:10.2},
\[\|\xi_j^+\|\le(1-\eta a_s)Z_s+\eta\frac{P_s}{2\sqrt m}W_s+\eta e_\star .\]
Taking the maximum over \(j\in S_s\) and using \(e_\star\le b_\star\)
gives \ref{app:11.2}. 
\end{proof}

\subsection{Closing the angular bootstrap}

\begin{proposition}[Angular bootstrap]\label{prop:angular-bootstrap}
{There are universal constants $C_W,C_\Theta\ge1$ (independent of $A$)
such that the following holds. Work on $\mathcal E$, assume that
$(H)$ holds at every iteration of $[N_1,n]$, with $n\le N_2-1$,
and assume the initial class-mass hypotheses of
Lemma~\ref{lem:logistic-comparison}. Suppose also that, for a universal
$C_1$,
\[
W_{s,N_1}\le Z_{s,N_1}\le C_1\varepsilon,
\qquad P_{s,N_1}<\frac\beta2,
\]
then for \(s=\pm1\)}
\[\max_{N_1\le k\le n+1}W_{s,k}\le C_W\,\varepsilon,\qquad\max_{N_1\le k\le n+1}Z_{s,k}\le C_\Theta\,\bar\varepsilon .\tag{12.1} \label{app:12.1}\]
\end{proposition}
\begin{proof} Fix \(s\) and use the assumed initial angular bound. Define
\[n_s:=\min\{k\in[N_1,n+1]:\ P_{s,k}>\beta/2\},\qquad\min\emptyset:=\infty .\]
Since \(P_{s,N_1}<\beta/2\), we have
\(n_s>N_1\).

\textbf{Regime I: \(N_1\le k<\min(n_s,n+1)\).} Here
\(P_{s,k}\le\beta/2\), so \(\delta_{s,k}\ge\frac\beta{4\sqrt m}\) and,
by (S3),
\[a_{s,k}=\delta_{s,k}-C_D\frac{\beta\zeta}{\sqrt m}\ \ge\ \frac\beta{8\sqrt m},\]
hence \(\eta a_{s,k}\ge\tau/8>0\); and
\(\eta a_{s,k}\le\eta\delta_{s,k}\le\eta\frac\beta{2\sqrt m}=\tau/2\le1\).
Part (1) of Lemma~\ref{lem:scalar-recursions} applied to \ref{app:11.1} with \(\lambda=\tau/8\),
\(g=\eta b_\star=\tau C_D\zeta\) gives, for
\(N_1\le k\le\min(n_s,n+1)\),
\[W_{s,k}\ \le\ \max\bigl\{C_1\varepsilon,\ 8C_D\zeta\bigr\}\ \le\ (C_1+24C_D)\,\varepsilon\ =:\ C_W'\varepsilon,\tag{12.2}\]
using \(\zeta\le3\varepsilon\) from \ref{app:1.4}. Feeding (12.2) and
\(P_{s,k}\le\beta/2\) into \ref{app:11.2},
\[Z_{s,k+1}\le(1-\eta a_{s,k})Z_{s,k}+\tau\Bigl(\tfrac14C_W'\varepsilon+C_D\zeta\Bigr),\]
and part (1) of Lemma~\ref{lem:scalar-recursions} gives
\(Z_{s,k}\le\max\{C_1\varepsilon,\ 8(\tfrac14C_W'+3C_D)\varepsilon\}=:C_Z'\varepsilon\)
on the same range. In particular, if \(n_s\le n+1\) then
\(W_{s,n_s}\le C_W'\varepsilon\) and \(Z_{s,n_s}\le C_Z'\varepsilon\).

{If \(n_s=\infty\)}, Regime II below is empty and (\ref{app:12.1}) already
follows from (12.2) and \(C_Z'\varepsilon\le C_Z'\bar\varepsilon\).

\textbf{Regime II: \(n_s\le k\le n+1\).} By part (2) of Lemma~\ref{lem:logistic-comparison},
\(P_{s,k}\le\beta(1+C_5\zeta)\), so
\(\delta_{s,k}\ge-\frac{C_5\beta\zeta}{2\sqrt m}\) and
\[-\eta a_{s,k}\ \le\ \eta\frac\beta{\sqrt m}\Bigl(\tfrac{C_5}2+C_D\Bigr)\zeta\ =:\ \mu\ \ge0 .\]
Also \(k-n_s\le N_2-N_1\le C_L\tau^{-1}L+1\). part (2) of Lemma~\ref{lem:scalar-recursions} applied to
\ref{app:11.1}:
\[
\begin{aligned}
W_{s,k}
&\le e^{\mu(k-n_s)}\bigl(W_{s,n_s}+\eta b_\star(k-n_s)\bigr)\\
&\le e^{C(C_L+1)\zeta L}
\Bigl(C_W'\varepsilon+C_D\zeta(C_LL+\tau)\Bigr)\\
&\le2\bigl(C_W'+3C_D(C_L+1)\bigr)\varepsilon
=:C_W\varepsilon.
\end{aligned}
\tag{12.3} \label{app:12.3}
\]
using \(\zeta L\le3\varepsilon\) and \(\zeta\le3\varepsilon\) from
\ref{app:1.4}, \(\tau\le1\), and \(e^{C(C_L+1)\zeta L}\le2\) from (S5). Since
Regime I already gave \(W_{s,k}\le C_W'\varepsilon\le C_W\varepsilon\),
the first half of (\ref{app:12.1}) holds throughout.

Now \ref{app:11.2} in Regime II, with \(P_{s,k}\le2\beta\) and \ref{app:12.3}:
\[Z_{s,k+1}\le(1+\mu)Z_{s,k}+\tau\bigl(C_W\varepsilon+C_D\zeta\bigr).\]
Part (2) of Lemma~\ref{lem:scalar-recursions} gives
\[Z_{s,k}\le e^{\mu(k-n_s)}\Bigl(Z_{s,n_s}+\tau(C_W\varepsilon+C_D\zeta)(k-n_s)\Bigr)\le2\Bigl(C_Z'\varepsilon+(C_W\varepsilon+3C_D\varepsilon)(C_LL+\tau)\Bigr)\le C_\Theta\bar\varepsilon,\]
\[C_\Theta:=2\bigl(C_Z'+(C_W+3C_D)(C_L+1)\bigr),\tag{12.4}\label{12.4}\] using
\(\varepsilon L=\bar\varepsilon\), \(\varepsilon\le\bar\varepsilon\) (as
\(L\ge1\)) and \(\tau\le1\). Regime I gave
\(Z_{s,k}\le C_Z'\varepsilon\le C_\Theta\bar\varepsilon\).

\end{proof}

\begin{lemma}[Angle comparison]\label{lem:angle-comparison}{If \(\theta\in[0,\tfrac\pi2]\) and
\(\sin\theta\le z\), then \(\theta\le\tfrac\pi2z\le2z\).}
\end{lemma}
\begin{proof} \(\sin\theta\ge\tfrac2\pi\theta\) on \([0,\tfrac\pi2]\)
(concavity of \(\sin\)). 
\end{proof}

\begin{lemma}[No jump to the far hemisphere]\label{lem:no-hemisphere-jump}
On $\mathcal E$, if $\Psi_k\in\mathcal R$ then
\[
\|u_{j,k+1}-u_{j,k}\|\le C\tau,
\qquad
|\theta_{j,k+1}-\theta_{j,k}|\le\tfrac\pi2 C\tau.
\]
Consequently, under the smallness conditions
$C\tau\le\tfrac1{100}$ {\rm (S1)} and
$\Theta_\star\le\tfrac1{10}$, if
$\theta_{j,k}\le\Theta_\star$, then
\[
\theta_{j,k+1}
\le \Theta_\star+\frac1{50}
<\frac\pi2.
\]
\end{lemma}
\begin{proof} Write \(w_{j,k+1}=w_{j,k}+\eta\widehat V^w_{j,k}\), where
\(\widehat V^w_{j,k}:=-\widehat g^w_{j,k}\), and put
\(\varpi_j:=\eta\widehat V^w_{j,k}/r_{j,k}\). Because
\(\Psi_k\in\mathcal R\), Lemma~\ref{lem:population-gradient-bounds} and the concentration bound \ref{app:4.1}
give
\(\|\widehat V^w_{j,k}\|\le C\beta b_{j,k}/\sqrt m\le2C\beta r_{j,k}/\sqrt m\).
Therefore \(\|\varpi_j\|\le C\tau\le1/5\) by (S1). Lemma~\ref{lem:normalization-remainder} now yields
\[\|u_{j,k+1}-u_{j,k}\|\le\|\varpi_j\|+14\|\varpi_j\|^2\le C\tau.\] For
unit vectors,
\(\|u-u'\|=2\sin(\angle(u,u')/2)\ge\tfrac2\pi\angle(u,u')\), so
\(\angle(u_{j,k+1},u_{j,k})\le\tfrac\pi2C\tau\). Since
\(u\mapsto\angle(u,s_j\theta_\star)\) is \(1\)-Lipschitz in the angular
metric, the same bound holds for \(|\theta_{j,k+1}-\theta_{j,k}|\). At
the first exit from \(\theta\le\Theta_\star\), the previous angle is at
most \(\Theta_\star\), so the one-step bound gives
\(\theta_{j,k}\le\Theta_\star+1/50<\pi/2\). 
\end{proof}

\subsection{Closing the balanced baseline}

\subsubsection{Convention on constants}

The named constants below are chosen sequentially, each in terms of
constants fixed earlier; no choice depends on a later one. The generic
symbols \(C\) and \(c\) continue to denote universal constants that may
change between occurrences. Separately, the proof uses finitely many
{smallness inequalities}, listed as (S1)--(S7); these are imposed
only after the constants they involve have been fixed. Since
\(\tau\le c_\eta\varepsilon/L\le\varepsilon_0\),
\(\rho_B\le c_1\varepsilon/L\le\varepsilon_0\) and
\(\zeta\le3\varepsilon\le3\varepsilon_0\) (Lemma~\ref{lem:smallness-budget}, with
\(c_1,c_\eta\le1\)), every occurrence of ``for \(\tau\) small'', ``for
\(\zeta\) small'', etc., in the preceding lemmas refers to one of these.

\subsubsection{Order of choice}

\begin{enumerate}
\def\labelenumi{\arabic{enumi}.}
\item
  First fix the constants in Proposition~\ref{prop:phase-one},
  Corollary~\ref{cor:balance-drift}, and Lemma~\ref{lem:concentration}.
  They are determined by the Phase-I, concentration, and balance estimates
  and do not depend on \(c_0\).
\item
  Choose a universal
  \(c_0\in(0,\min\{c_0^{\max},(2C_6')^{-1}\}]\), where
  \(c_0^{\max}=1/(8C_2)\). Appendix~\ref{app:robust-init} may reduce
  this fixed constant and then reduce \(\varepsilon_0\) accordingly.
\item
  \(C_6:=25\), \(\ C_L:=C_7=C_6+2=27\) (Lemma~\ref{lem:logistic-comparison}).
\item
  \(C_5\) (Lemma~\ref{lem:phase-two-mass}); \(C_9\) (Lemma~\ref{lem:transverse-recursion}); \(C_8\) (Lemma~\ref{lem:angular-recursions},
  Step 2).
\item
  \(C_D\) (Lemma~\ref{lem:angular-recursions}, Step 6): an explicit expression in
  \(C_5,C_8,C_9\) and universal constants, satisfying in particular
  \(C_D\ge9C_9\).
\item
  \(C_\Delta\) (Step 3 of the proof below): an explicit expression in
  \(C_\sharp,C_4,C_L\).
\item
  \(c_\eta:=\min\{1,\ 1/C_\Delta\}\) and \(c_1:=1\);
  \(\ C_B:=(2C_\star/c_1)^2\).
\item
  \(C_W':=C_1+24C_D\), \(\ C_Z':=\max\{C_1,\ 2C_W'+24C_D\}\),
  \(\ C_W:=2(C_W'+3C_D(C_L+1))\),
  \(\ C_\Theta:=2(C_Z'+(C_W+3C_D)(C_L+1))\) (Proposition~\ref{prop:angular-bootstrap}).
  Appendix~\ref{app:robust-init} may enlarge \(C_1\); in that case these
  four derived constants are recomputed in the displayed order before
  \(A\) is chosen. They remain independent of \(A\).
\item
  Choose \(A\ge 4C_\Theta+2C_1+1\). Appendix~\ref{app:robust-init}
  may enlarge this universal constant when passing to the full initialization
  class in Assumption~\ref{assumption:initialization}.
\item
  \(\varepsilon_0\): a sufficiently small positive value satisfying (S1)--(S7) below.
\item
  \(C_T:=C_L+6\).
\end{enumerate}

\subsubsection{Smallness conditions}

With \(c_\tau,c_\iota,c_\zeta\) denoting explicit constants determined
by items 1--8:

\begin{itemize}
\item
  \textbf{(S1)} \(\tau\le c_\tau\), where \(c_\tau\) is small enough
  for: \(2C\tau<1\) and \(2C\tau\le\tfrac15\)
  (Lemma~\ref{lem:phase-two-mass}, Step 0;
  Lemma~\ref{lem:transverse-recursion}, Step 1; and
  Proposition~\ref{prop:phase-one} (c)); \(C_3'\tau\le\tfrac1{512}\) and
  \(C_4\tau\le\tfrac12\) (Proposition~\ref{prop:phase-one} (f)); \(\tau\le\tfrac1{12}\) and
  \(\tau\le\tfrac12\) (Lemma~\ref{lem:logistic-comparison}); \(C_8\tau\le1\) and
  \(C\tau\le\tfrac12\) (Lemma~\ref{lem:angular-recursions});
  \(C\tau\le\tfrac1{100}\) (Lemma~\ref{lem:no-hemisphere-jump}).
  Guaranteed by \(\varepsilon_0\le c_\tau\) (since
  \(\tau\le c_\eta\varepsilon_0\le\varepsilon_0\)).
\item
  \textbf{(S2)} \(\iota L\le c_\iota\), where \(c_\iota\) is small
  enough for: \(e^{3C\iota L}\le2\) \ref{app:7.2}; \(3C\iota L\le\tfrac18\)
  (Proposition~\ref{prop:phase-one} (a)); \(C\iota\le\tfrac1{16}\) (Proposition~\ref{prop:phase-one} (c), (g)).
  Guaranteed by \(\varepsilon_0\le c_\iota/2\) (since
  \(\iota L\le2\varepsilon\le2\varepsilon_0\) by \ref{app:1.4}).
\item
  \textbf{(S3)} \(\zeta\le c_\zeta\), where \(c_\zeta\) is small enough
  for: \(C_5\zeta\le\tfrac14\) (so \(\beta_-\ge\tfrac34\beta\),
  \(\beta_+\le\tfrac32\beta\), Lemma~\ref{lem:logistic-comparison}); \(C_D\zeta\le\tfrac18\)
  (Proposition~\ref{prop:angular-bootstrap} Regime I); and \(C\zeta\le\tfrac12\), which together with
  \(\tau\le1\) gives \(C\tau\zeta\le\tfrac12\) in Corollary~\ref{cor:no-mass-collapse}.
  Guaranteed by \(\varepsilon_0\le c_\zeta/3\).
\item
  \textbf{(S4)} \(\varepsilon_0\le2c_6c_0\) (part (3) of Lemma~\ref{lem:logistic-comparison}).
\item
  \textbf{(S5)} \(e^{3C(C_L+1)\varepsilon_0}\le2\)
  (Corollary~\ref{cor:no-mass-collapse} and the Regime-II estimate in
  Proposition~\ref{prop:angular-bootstrap}, using
  \(\zeta L\le3\varepsilon_0\)).
\item
  \textbf{(S6)} \(\varepsilon_0\le10^{-2}\) and
  \(A^2\varepsilon_0\log^3(1/\varepsilon_0)\le1\) (Lemma~\ref{lem:smallness-budget}).
\item
  \textbf{(S7)} \(c_\eta\le1/C_\Delta\) (so
  \(C_\Delta\tau L\le\varepsilon\), giving the second inequality in
  (H4)).
\end{itemize}

All of (S1)--(S6) are upper bounds on \(\varepsilon_0\) in terms of
constants defined in items 1--9; (S7) is item 7. Hence the choice is
consistent and all constants are universal.

\begin{proof}[Proof for the balanced equal-magnitude initialization]

Apply Lemma~\ref{lem:concentration} with \(N=N_\star\); let \(\mathcal E\) be the
resulting event, \(\mathbb P(\mathcal E)\ge1-\delta\). By Lemma~\ref{lem:batch-consequence},
\(\rho_B\le c_1\varepsilon/L\); by Lemma~\ref{lem:smallness-budget}, \ref{app:1.4} holds. All
statements below are deterministic consequences valid pointwise on
\(\mathcal E\).

\textbf{Phase I.} Proposition~\ref{prop:phase-one} gives: \(\Psi_n\in\mathcal R\) and
\(\|f_n\|_2\le\beta\varepsilon^2\) for \(0\le n\le N_1\); sign stability
and non-vanishing; \(|\Delta_{j,n}|/p_{j,n}\le C_4\tau\);
\(\theta_{j,N_1}\le C_1\varepsilon\);
\(p_{j,N_1}\asymp p_0\varepsilon^{-2}\);
\(P_{s,N_1}\asymp c_0\beta\varepsilon^2\). Since \(A\ge2C_1+1\) and
\(L\ge1\),
\[\Theta_{N_1}\le C_1\varepsilon<\tfrac12A\varepsilon L=\tfrac12\Theta_\star,\]
so Phase II starts strictly inside the bootstrap region.
More precisely, Proposition~\ref{prop:phase-one} item 6 supplies the
hypotheses of Lemma~\ref{lem:logistic-comparison} with
\(c_{\rm in}=c_6\): the upper bound is at most \(\beta/2\) after the
universal smallness choices, while (S4) gives
\(\log(1/(2c_6c_0))\le L\).

\textbf{Phase II: first-exit induction.} Define
\[n^\dagger:=\min\bigl\{n\in[N_1,N_2]:\ \text{(H) fails at }n\bigr\},\qquad\min\emptyset:=\infty .\]
(H) holds at \(n=N_1\): (H1) by the display above; (H2) since
\(P_{s,N_1}\le\beta\); (H3), (H4) by Proposition~\ref{prop:phase-one}, using
\(C_4\tau\le C_\Delta\tau L\) (as \(C_\Delta\ge C_4\) and \(L\ge1\)).
Hence \(n^\dagger>N_1\).

Suppose, for contradiction, that \(n^\dagger\le N_2\). Then (H) holds at
every \(n\in[N_1,n^\dagger-1]\), so Lemmas~\ref{lem:phase-two-mass}, \ref{lem:transverse-recursion}, and~\ref{lem:angular-recursions} apply at
every such \(n\), and Lemma~\ref{lem:logistic-comparison}, Corollary~\ref{cor:no-mass-collapse} and Proposition~\ref{prop:angular-bootstrap}
apply with \(n=n^\dagger-1\le N_2-1\). 

To rule out the first exit, it suffices to verify at $n^\dagger$ the
four bootstrap conditions: sign stability and non-vanishing (H3),
class-mass control (H2), layer-balance control (H4), and angular
control (H1).

\textbf{Step 1 -- (H3) and non-vanishing.} Lemma~\ref{lem:phase-two-mass} Step 0, applied at
\(n=n^\dagger-1\), gives \(\operatorname{sign}(a_{j,n^\dagger})=s_j\)
and \(b_{j,n^\dagger},r_{j,n^\dagger}>0\) for every \(j\).

\textbf{Step 2 -- (H2).} Part (2) of Lemma~\ref{lem:logistic-comparison} with \(n+1=n^\dagger\) gives
\(P_{s,n^\dagger}\le\tfrac32\beta<2\beta\).

\textbf{Step 3 -- (H4).} By Corollary~\ref{cor:no-mass-collapse}, \(p_{j,k}\le2p_{j,n^\dagger}\)
for \(N_1\le k\le n^\dagger\). Telescoping Corollary~\ref{cor:balance-drift} from \(N_1\)
(legitimate, since \(\Psi_k\in\mathcal R\) for all \(k<n^\dagger\)) and
using Proposition~\ref{prop:phase-one} item 3 together with
\(n^\dagger-N_1\le C_L\tau^{-1}L+1\),
\[|\Delta_{j,n^\dagger}|\le|\Delta_{j,N_1}|+C_\sharp\tau^2\!\!\sum_{N_1\le k<n^\dagger}\!\!p_{j,k}\le2C_4\tau\,p_{j,n^\dagger}+2C_\sharp\tau^2\bigl(C_L\tau^{-1}L+1\bigr)p_{j,n^\dagger}\le C_\Delta\,\tau L\,p_{j,n^\dagger},\]
with \(C_\Delta:=2C_4+2C_\sharp(C_L+1)\) (using \(\tau\le1\le L\)). By
(S7), \(C_\Delta\tau L\le\varepsilon\).

\textbf{Step 4 -- (H1); the contradiction.} Proposition~\ref{prop:angular-bootstrap} with
\(n=n^\dagger-1\) gives \(Z_{s,n^\dagger}\le C_\Theta\bar\varepsilon\)
for both \(s\). By Lemma~\ref{lem:no-hemisphere-jump} (applicable since \(\Psi_k\in\mathcal R\)
for \(k<n^\dagger\), \(C\tau\le\tfrac1{100}\) by (S1), and
\(\Theta_\star\le\tfrac1{10}\) by Lemma~\ref{lem:smallness-budget}), every
\(\theta_{j,n^\dagger}<\tfrac\pi2\), so Lemma~\ref{lem:angle-comparison} applies:
  \[\Theta_{n^\dagger}=\max_s\Theta_{s,n^\dagger}\le2\max_sZ_{s,n^\dagger}\le2C_\Theta\bar\varepsilon
  <(4C_\Theta+2C_1+1)\bar\varepsilon\le A\bar\varepsilon=\Theta_\star .\]

Steps 1--4 show that (H) holds at \(n^\dagger\), contradicting its
definition. Hence \(n^\dagger=\infty\), i.e.~\textbf{(H) holds at every
\(n\in[N_1,N_2]\)}, and therefore all conclusions of Lemmas~\ref{lem:phase-two-mass}--\ref{lem:angular-recursions} and Propositions~\ref{prop:phase-one} and~\ref{prop:angular-bootstrap} are valid on
\([N_1,N_2]\).

\textbf{Conclusions.} Sign stability, non-vanishing, and the balance bound follow on Phase I from Proposition~\ref{prop:phase-one}, and on Phase II from Steps 1 and 3 above. Proposition~\ref{prop:angular-bootstrap}, followed by Lemmas~\ref{lem:angle-comparison} and~\ref{lem:no-hemisphere-jump}, gives \(\Theta_{N_\star}\le 2C_\Theta\bar\varepsilon\). Part (4) of Lemma~\ref{lem:logistic-comparison} gives \(|P_{s,N_\star}-\beta|\le C\beta\bar\varepsilon\). The sharper endpoint and population-fitting bounds stated in Theorem~\ref{thm:main-sgd} follow from Proposition~\ref{prop:sharp-fit}.

{Horizon.} By Lemma~\ref{lem:horizon-arithmetic}, \(N_1\le4L/\tau+1\), and
\(N_2-N_1\le C_LL/\tau+1\); since \(\tau\le1\le L\),
\(N_\star\le(C_L+6)L/\tau=C_T\frac{\sqrt m}{\eta\beta}\log\frac1\varepsilon\).

All of the above holds on \(\mathcal E\), of probability at least
\(1-\delta\). 
\end{proof}

\section{Independently varying initialization magnitudes}\label{app:robust-init}

The technical proof in Appendix~\ref{app:technical} treats the exactly
balanced, equal-magnitude special case. We now extend the argument to the
initialization in Assumption~\ref{assumption:initialization}. The
equal-magnitude proof used positivity of both mode coefficients, which is
no longer guaranteed. We instead compare each neuron with an explicit
teacher-only trajectory that stays uniformly inside the layer-ratio bounds.

No training update is changed. The symbols
$p_0,L,\bar\varepsilon,\tau,h=\tau/2,N_1,N_\star$ retain their meanings
from the previous proof; $p_0$ is now a reference scale rather than every
neuron's exact initial mass.

\begin{proposition}[Initialization robustness]\label{prop:robust-init}
Under the step, horizon, and fresh-batch conditions of Appendix~\ref{app:technical}, the finite-time alignment and fitting conclusions hold, after adjusting
universal constants, for every integer $m\ge2$ under
\begin{equation}\label{eq:robust-init}
\begin{gathered}
 \frac m4\le |S_s|\le\frac{3m}4\quad(s=\pm1),\qquad
 a_{j,0}=s_jb_{j,0},\quad w_{j,0}=r_{j,0}u_j^0,\\
 b_{j,0},r_{j,0}\in[\sqrt{p_0},\tfrac54\sqrt{p_0}],\quad
 \|u_j^0\|=1,\quad s_j\langle u_j^0,\theta_\star\rangle\ge-\tfrac12.
\end{gathered}
\end{equation}
In particular, neither $b_{j,0}=r_{j,0}$, equal neuron masses, nor equal
sign-class cardinalities are required. The step and batch orders remain
\[
 \eta\le c_\eta\frac{\sqrt m\,\varepsilon}{\beta L},\qquad
 B\ge C_B\frac{L^2}{\varepsilon^2}
 \left(d+\log\frac{16mN_\star}{\delta}\right),\qquad
 N_\star=N_1+\left\lceil C_L L/\tau\right\rceil.
\]
All output signs remain unchanged and every neuron is nonzero on
$[0,N_\star]$. The balance conclusions become
\begin{align}
 \frac{|\Delta_{j,n}|}{p_{j,n}}
 &\le C\big((1+h)^{-2n}+\tau\big),&&0\le n\le N_1,
 \label{eq:robust-balance-early}\\
 \frac{|\Delta_{j,n}|}{p_{j,n}}
 &\le C\big(\varepsilon^2+\tau L\big)\le\varepsilon,
 &&N_1\le n\le N_\star.
 \label{eq:robust-balance-late}
\end{align}
The final inequality is secured by the universal smallness choices.
There is no assertion of $O(\varepsilon)$ balance at initialization.
\end{proposition}

\begin{proof}
We give the changes to Phase I, then verify the inputs to Phase II.
The conditional concentration event of Lemma~\ref{lem:concentration} depends only on the
working region $\mathcal R$, not on initialization equalities. Its
probability is unchanged. Work throughout on this event and put
$\iota=\varepsilon^2+\rho_B$, $q_n=(1+h)^n$ and $\ell_n=(1-h)^n$.
Constants below are universal.

\paragraph{A teacher-only comparison.}
For a fixed neuron suppress $j$, and set
$A_n=s_ja_{j,n}$, $B_n=s_j\theta_\star^\top w_{j,n}$,
$G_0=(b_0+B_0)/2$, $H_0=(b_0-B_0)/2$.
Define the exact unperturbed discrete dynamics by
\[
 \widetilde A_n=G_0q_n+H_0\ell_n,\qquad
 \widetilde B_n=G_0q_n-H_0\ell_n,\qquad
 \widetilde w_n=s_j\widetilde B_n\theta_\star+Q_\star w_0,
 \qquad \widetilde r_n=\|\widetilde w_n\|.
\]
First suppose $B_0<0$. The one-sided condition then implies $|B_0|\le r_0/2$, and the initialization gives
\[
 \frac3{16}\sqrt{p_0}\le G_0,H_0\le\frac{15}{16}\sqrt{p_0}.
\]
The elementary horizon bounds in Lemma~\ref{lem:horizon-arithmetic} give
$hN_1\le3L$, $q_{N_1}\in[\varepsilon^{-1},2\varepsilon^{-1}]$,
and $\ell_n\le q_n^{-1}$.
Thus $\widetilde A_n\ge(3/16)\sqrt{p_0}q_n$.
Also $\|Q_\star w_0\|\ge(\sqrt3/2)\sqrt{p_0}$.
If $q_n\le8$ this implies
$\widetilde r_n\ge(\sqrt3/16)\sqrt{p_0}q_n$.
If $q_n>8$, use
$\widetilde B_n\ge\sqrt{p_0}[(3/16)q_n-(15/16)q_n^{-1}]
\ge(1/10)\sqrt{p_0}q_n$.
Consequently both magnitudes are bounded above and below by universal
multiples of $\sqrt{p_0}q_n$ on $[0,N_1]$.

We also need a ratio bound with a strict margin inside $[1/2,2]$.
Let $T=\|Q_\star w_0\|^2$ and $D_0=b_0^2-B_0^2>0$.
Writing $x=b_0/r_0\in[4/5,5/4]$ and
$z=B_0/r_0\in[-1/2,1/2]$ gives
\[
 \frac{13}{25}\le\frac{D_0}{T}
 =\frac{x^2-z^2}{1-z^2}\le\frac74.
\]
Since $\widetilde A_n^2-\widetilde B_n^2=D_0(1-h^2)^n$ and
$N_1h^2\le3hL$, reducing the universal smallness constants ensures
$(1-h^2)^n\ge25/26$. Therefore
\[
 \frac12\le\frac{D_0(1-h^2)^n}{T}\le\frac74,
 \qquad
 \frac1{\sqrt2}\le
 \frac{\widetilde A_n}{\widetilde r_n}
 \le\frac{\sqrt7}{2}.
\]
The last ratio squared lies between $1$ and
$D_0(1-h^2)^n/T$, since its numerator and denominator have the same
additional term $\widetilde B_n^2$.

If $B_0\ge0$, $H_0$ need not be positive, so we instead compare with
\[
 A_n^c=G_0q_n+H_0/q_n,\qquad B_n^c=G_0q_n-H_0/q_n,
 \qquad r_n^c=\sqrt{(B_n^c)^2+\|Q_\star w_0\|^2}.
\]
These equal $A_n^c=b_0\cosh(\log q_n)+B_0\sinh(\log q_n)$ and
$B_n^c=B_0\cosh(\log q_n)+b_0\sinh(\log q_n)$. Both $A_n^c,B_n^c$ are nonnegative and nondecreasing, and
$(A_n^c)^2-(r_n^c)^2=b_0^2-r_0^2$. Since $r_n^c\ge r_0$,
\[
 \min(1,b_0/r_0)\le A_n^c/r_n^c\le\max(1,b_0/r_0),
\]
so the ratio is in $[4/5,5/4]$. Also $A_n^c\ge(b_0/2)q_n$: if $H_0\ge0$ use $G_0\ge b_0/2$, and otherwise use $H_0/q_n\ge H_0q_n$. Thus both layer magnitudes are comparable to $\sqrt{p_0}q_n$.
Since $\ell_n=q_n^{-1}(1-h^2)^n$,
\[
 |\ell_n-q_n^{-1}|\le nh^2q_n^{-1}\le3hLq_n^{-1}.
\]
The discrete comparison differs from $(A_n^c,B_n^c)$ by at most $ChL\sqrt{p_0}q_n^{-1}$. For sufficiently small $hL$, it therefore has the same magnitude bounds and a strict ratio margin inside $[1/2,2]$. This also covers a direction exactly at its selected teacher pole, where the transverse component is zero. In both cases $G_0\ge c\sqrt{p_0}$ and $|H_0|\le C\sqrt{p_0}$; these are the mode bounds needed below.

\paragraph{Closing Phase I without initial balance.}
Use the same first-exit induction as Proposition~\ref{prop:phase-one}, on sign stability,
nonzero magnitudes, $\mathcal R$, and $\|f_n\|_2\le\beta\varepsilon^2$.
Before a possible exit, Lemma~\ref{lem:phase-one-dynamics} applies; its growth bound now starts
from $b_0+r_0\le(5/2)\sqrt{p_0}$. Hence
$b_n+r_n\le C\sqrt{p_0}q_n$.
Variation of constants in the $G,H$ recursions gives, including a
potential first exiting step,
\begin{align*}
 |G_n-G_0q_n|&\le C\iota hn\sqrt{p_0}q_n,\\
 |H_n-H_0\ell_n|&\le C\iota\sqrt{p_0}q_n,\\
 \|Q_\star w_n-Q_\star w_0\|&\le C\iota\sqrt{p_0}q_n.
\end{align*}
These follow respectively by dividing the growing-mode recursion by
$q_n$, summing the geometric convolution with $\ell_n$, and summing
the transverse increments. Thus
\[
 |A_n-\widetilde A_n|+\|w_n-\widetilde w_n\|
 \le C\iota L\sqrt{p_0}q_n.
\]
Choose $\varepsilon_0$ small enough that $C\iota L$ is smaller than
the fixed margins in the preceding magnitude and ratio estimates.
Then $A_n,r_n>0$, $b_n/r_n\in(1/2,2)$, and
$cp_0q_n^2\le p_n\le Cp_0q_n^2$.
Summing over all neurons gives
$\Pi_n\le C\sqrt m p_0\varepsilon^{-2}=Cc_0\beta\varepsilon^2$.
Choosing $c_0$ small closes $\|f_n\|_2\le\Pi_n/\sqrt2<\beta\varepsilon^2$
and $\Pi_n<6\beta$. This prevents the first exit and proves the estimates
through $N_1$.

The sharper transverse bound gives
$\|Q_\star w_{N_1}\|\le C\sqrt{p_0}$, using
$\iota/\varepsilon\le C$.
The growing and decaying modes give
$B_{N_1}\ge c\sqrt{p_0}/\varepsilon>0$.
Hence every endpoint angle is at most $C\varepsilon$.
Since each sign class contains between $m/4$ and $3m/4$ neurons,
$P_{s,N_1}\asymp c_0\beta\varepsilon^2$ for both signs, with universal
constants. No equality of the two masses is used. Indeed, summing $cp_0\varepsilon^{-2}\le p_{j,N_1}\le Cp_0\varepsilon^{-2}$ over any class with $m/4\le|S_s|\le3m/4$ gives these bounds for every integer width. The later weighted recursions use class masses, not equal counts. The count condition is feasible for every $m\ge2$ by choosing $\lfloor m/2\rfloor$ and $\lceil m/2\rceil$ neurons.

\paragraph{The inherited balance defect.}
The exact samplewise cancellation still holds, but Lemma~\ref{lem:balance-propagation} must be
replaced by
\[
 |\Delta_{j,n}|\le |\Delta_{j,0}|+
 C_\sharp\tau^2\sum_{k<n}p_{j,k},
 \qquad |\Delta_{j,0}|\le\frac9{16}p_0.
\]
The geometric mass bounds and
$\sum_{k<n}q_k^2\le q_n^2/(2h)$ prove
\eqref{eq:robust-balance-early}. At $N_1$ this is
$C(\varepsilon^2+\tau)$.

\paragraph{Phase II and the full horizon.}
Let $c_{\rm I},C_{\rm I},C_{\rm ang}>0$ be universal constants such that
the Phase-I estimates above give
\begin{equation}
\label{eq:robust-phase-one-endpoint}
\begin{gathered}
 c_{\rm I}c_0\beta\varepsilon^2
 \le P_{s,N_1}\le
 C_{\rm I}c_0\beta\varepsilon^2,
 \\
 \Theta_{N_1}\le C_{\rm ang}\varepsilon,
 \qquad
 \frac{|\Delta_{j,N_1}|}{p_{j,N_1}}
 \le C_{\rm I}(\varepsilon^2+\tau).
\end{gathered}
\end{equation}
Reduce the fixed universal constant $c_0$, if necessary, so that the
upper class-mass bound in \eqref{eq:robust-phase-one-endpoint} is at most
$\beta/2$. Reduce
$\varepsilon_0$ so that
$\log(1/(2c_{\rm I}c_0))\le L$. With these substitutions,
Lemma~\ref{lem:logistic-comparison} applies with
\(c_{\rm in}=c_{\rm I}\) to each sign class separately:
the sandwich and no-overshoot conclusions are unchanged, each lower
comparison reaches $\beta/2$ within $25L/\tau$ steps, and the remaining
$2L/\tau$ steps give the same endpoint accuracy. Unequal class sizes and
unequal values of $P_{+,N_1}$ and $P_{-,N_1}$ are therefore harmless.

The Phase-II first-exit argument is unchanged except for the inherited
layer imbalance. Replace (H4) by
\begin{equation}
\label{eq:robust-phase-two-balance}
 \textbf{(H4$'$)}\qquad
 \frac{|\Delta_{j,n}|}{p_{j,n}}
 \le C_\Delta^{\rm rob}(\varepsilon^2+\tau L)\le\varepsilon
 \qquad\text{for every }j.
\end{equation}

The proofs of Lemmas~\ref{lem:phase-two-mass},
\ref{lem:transverse-recursion}, and~\ref{lem:angular-recursions} use (H4)
only through the consequence $|\Delta_j|/p_j\le\varepsilon$.
Hence all Phase-II lemmas and Proposition~\ref{prop:angular-bootstrap}
remain valid with (H4$'$).

Enlarge $C_1$ so that $C_1\ge C_{\rm ang}$ and recompute
$C_W',C_Z',C_W,C_\Theta$ in the order of the master constant list.
Then~\eqref{eq:robust-phase-one-endpoint}, together with the class-mass
bounds above, supplies the same Phase-II initial conditions as in
Appendix~\ref{app:technical}. Unequal class sizes enter only through
these class masses, so no further modification is needed.

It remains only to verify the modified balance condition at a possible
first exit. Let $n^\dagger$ be the first failure time of
(H1)--(H3) and (H4$'$). Before $n^\dagger$,
Corollary~\ref{cor:no-mass-collapse} gives
$p_{j,k}\le2p_{j,n^\dagger}$ for
$N_1\le k\le n^\dagger$. Therefore
\[
\begin{aligned}
 |\Delta_{j,n^\dagger}|
 &\le |\Delta_{j,N_1}|+
 C_\sharp\tau^2\sum_{N_1\le k<n^\dagger}p_{j,k}\\
 &\le 2C_{\rm I}(\varepsilon^2+\tau)p_{j,n^\dagger}
 +2C_\sharp\tau^2(C_LL/\tau+1)p_{j,n^\dagger}\\
 &\le C_\Delta^{\rm rob}(\varepsilon^2+\tau L)
 p_{j,n^\dagger}.
\end{aligned}
\]
Since $\tau L\le c_\eta\varepsilon$, choosing
$c_\eta\le1/(2C_\Delta^{\rm rob})$ and
$\varepsilon_0\le1/(2C_\Delta^{\rm rob})$ gives
$C_\Delta^{\rm rob}(\varepsilon^2+\tau L)\le\varepsilon$.
Thus (H4$'$) also holds at $n^\dagger$.

The other three exit conditions are closed exactly as in the balanced
proof: Lemma~\ref{lem:phase-two-mass} gives sign stability and
nonvanishing, Lemma~\ref{lem:logistic-comparison} gives the class-mass
bound, and Proposition~\ref{prop:angular-bootstrap} together with
Lemmas~\ref{lem:no-hemisphere-jump} and~\ref{lem:angle-comparison}
gives the strict angular bound. Hence the same first-exit contradiction
shows that (H1)--(H3) and (H4$'$) hold through $N_\star$.

The logistic endpoint estimate and
Proposition~\ref{prop:angular-bootstrap} now give the class-mass and
alignment conclusions. Sign stability, nonvanishing, and
\eqref{eq:robust-balance-late} follow from the argument above. The horizon
bound is unchanged. Appendix~\ref{app:sharp-fit} converts these estimates
into the stated $O(\beta\varepsilon)$ population-fitting bound.

The constants are chosen in the same sequential order as in
Appendix~\ref{app:technical}. The robust Phase-I constants and
$C_\Delta^{\rm rob}$ are fixed before $c_\eta$; the angular constants
are then recomputed before choosing $A$; finally, $\varepsilon_0$ is
chosen sufficiently small to satisfy the resulting smallness conditions.
These choices also provide the deterministic constants used in
Appendix~\ref{app:single-example}.
\end{proof}

\begin{corollary}[A teacher-direction-independent random initialization]\label{cor:random-init}
Draw iid Rademacher signs $s_j$, iid uniform directions $u_j^0$ on
$\mathbb S^{d-1}$, and independently draw all $b_{j,0},r_{j,0}$ uniformly
on $[\sqrt{p_0},(5/4)\sqrt{p_0}]$.
The finite-time conclusions hold with probability at least
\[
 1-\delta-2e^{-m/8}-m(5/4)^{-(d-1)/2}.
\]
The lower bound is informative when $m$ and $d$ are sufficiently large;
for every integer width $m\ge2$ the deterministic assumptions remain applicable.
For example, the initialization failure probability is at most $\gamma$
if $m\ge8\log(4/\gamma)$ and
$d\ge1+2\log(4m/\gamma)/\log(5/4)$.
\end{corollary}
\begin{proof}
Hoeffding's inequality bounds failure of the sign-count condition by
$2e^{-m/8}$. For a uniform direction write $u=g/\|g\|$ with standard
Gaussian $g$. The event $|u^\top\theta_\star|>1/2$ is, after rotation,
$3g_1^2>\sum_{k=2}^dg_k^2$. Markov's inequality applied to the exponential
with coefficient $1/8$ bounds its probability by
$2(5/4)^{-(d-1)/2}$. By symmetry the one-sided event $s_j u_j^\top\theta_\star<-1/2$ has half this probability. Union bound over neurons, then combine with the
conditional training guarantee: the fresh-batch result above, or
Theorem~\ref{se:thm:reuse} for fixed-dataset SGD. In the latter case the
initialization is drawn independently of the dataset and indices. No
initialization step uses the teacher.
\end{proof}

\paragraph{Scope.}
The bounded interval is a convenient explicit sufficient condition, not
an optimized range. Independent Gaussian output weights are not covered:
some magnitudes can be arbitrarily small relative to the corresponding
hidden norms, so the uniform growing-mode margin used above can fail.
The small overall scale remains essential to this proof.

\section{Sharper fitting and transverse gradient bounds}\label{app:sharp-fit}

\begin{proposition}[An $O(\varepsilon)$ fitting endpoint]\label{prop:sharp-fit}
For the controlled trajectory established in Appendices~\ref{app:technical}--\ref{app:robust-init},
\[
 |P_{s,N_\star}-\beta|\le C\beta\varepsilon\quad(s=\pm1),\qquad
 \|f_{\Psi_{N_\star}}-y\|_{L^2(\gamma_d)}\le C\beta\varepsilon,
 \qquad \mathcal L(\Psi_{N_\star})\le C\beta^2\varepsilon^2.
\]
Moreover, with $Q_\star=I-\theta_\star\theta_\star^\top$,
\[
 \mathbb E\|Q_\star\nabla f_{\Psi_{N_\star}}(X)\|^2
 \le C\beta^2\varepsilon^2,\qquad
 \sum_{i\ge2}\lambda_i(G_{f_{\Psi_{N_\star}}})\le C\beta^2\varepsilon^2.
\]
These claims hold both for the balanced special case and for
Proposition~\ref{prop:robust-init}.
\end{proposition}
\begin{proof}
Lemma~\ref{lem:logistic-comparison} compares $P_s$ with logistic recursions targeting
$\beta_\pm=\beta(1\pm C_5\zeta)$.
The lower comparison reaches $\beta/2$ by
$k_0\le N_1+C_6 L/\tau$ and subsequently has deficit at most
$\beta\exp[-\tau(n-k_0)/2]$.
Choosing $C_L\ge C_6+2$ gives deficit at most $C\beta\varepsilon$
at $N_\star$; the constant allows integer rounding.
The smallness budget $\zeta L\le3\varepsilon$ therefore yields
$|P_{s,N_\star}-\beta|\le C\beta\varepsilon$.
This retains the exponential estimate instead of weakening it to
$\beta\bar\varepsilon$.

Write $f^\sharp(x)=P_+\sigma(z)-P_-\sigma(-z)$, where
$z=\theta_\star^\top x$, and suppress $N_\star$.
Proposition~\ref{prop:angular-bootstrap} gives the stronger averaged estimate
$W_s^2=(\sum_{j\in S_s}p_j)^{-1}\sum_{j\in S_s}p_j\sin^2\theta_j
\le C\varepsilon^2$.
The angular bootstrap keeps every $\theta_j$ acute, so
$\theta_j\le2\sin\theta_j$.
The ReLU Lipschitz inequality and weighted Cauchy--Schwarz give
\[
 \|f-f^\sharp\|_2\le\frac1{\sqrt m}\sum_jp_j\theta_j
 \le2\sum_sP_sW_s\le C\beta\varepsilon.
\]
Also $\|f^\sharp-y\|_2\le|P_+-\beta|+|P_--\beta|$, proving the fit bound.
Pointwise off gate boundaries,
\[
 \|Q_\star\nabla f(x)\|
 \le\frac1{\sqrt m}\sum_jp_j\sin\theta_j
 \le\sum_sP_sW_s\le C\beta\varepsilon.
\]
Squaring and taking expectations proves the transverse bound.
The spectral tail is at most $\operatorname{tr}(Q_\star G_f)$ by the
variational principle. No post-$N_\star$ training step is used.
\end{proof}

Appendix~\ref{app:single-example} transfers these conclusions to the actual SGD time $T_\star$.

\section{Variational and neuronwise selection}
\label{app:selection-principle}

\subsection{Variational characterization}

\begin{proof}[Proof of Proposition~\ref{prop:variational-characterization}]
For active neurons, let
\(s_j=\operatorname{sign}(a_j)\),
\(u_j=w_j/\|w_j\|\), and
\(\theta_j=\angle(u_j,s_j\theta_\star)\). Define
\[
c(\Psi)=\mathbb E[f_\Psi(X)\theta_\star^\top X],
\qquad
D(\Psi)=\frac1{\sqrt m}\sum_{p_j>0}p_j(1-\cos\theta_j),
\]
and
\[
H(\Psi)=\frac1{2\sqrt m}\sum_j(|a_j|-\|w_j\|)^2.
\]
The Gaussian identity
\(\mathbb E[\sigma(w^\top X)\theta_\star^\top X]
=\theta_\star^\top w/2\) gives
\[
\Pi(\Psi)-2c(\Psi)=D(\Psi)
\]
and direct expansion gives
\[
\frac{S(\Psi)}{2\sqrt m}-2c(\Psi)
=D(\Psi)+H(\Psi)
=\frac1{2\sqrt m}\sum_j\|w_j-a_j\theta_\star\|^2.
\]
At an exact fit, \(c(\Psi)=\beta\). Since \(D,H\ge0\), these identities
imply \(\Pi(\Psi)\ge2\beta\) and
\(S(\Psi)\ge4\beta\sqrt m\).

Equality in the first bound holds exactly when every active neuron has
\(\theta_j=0\), or equivalently \(u_j=s_j\theta_\star\). The represented
function is then
\(P_+\sigma(\theta_\star^\top x)-P_-\sigma(-\theta_\star^\top x)\),
and exact fitting forces \(P_+=P_-=\beta\). Equality in the second bound
also requires \(H=0\), including at inactive neurons, which is equivalent
to \(w_j=a_j\theta_\star\) for every \(j\). Both lower bounds are attained,
for example, by one positive and one negative active neuron with squared
magnitudes \(\beta\sqrt m\), with all remaining raw parameters zero.
\end{proof}

\subsection{Neuronwise mass selection}

\begin{proof}[Proof of Theorem~\ref{thm:explicit-selection-main}]
We work with the controlled trajectory from Appendices~\ref{app:technical}--\ref{app:robust-init}; Appendix~\ref{app:single-example} transfers the conclusions to $T_\star$. All constants are independent
of $m,d,\beta,\varepsilon,\delta$. We retain the error terms in Phase I
rather than only its constant-factor growth conclusions.
Write $h=\tau/2$, $q=(1+h)^{N_1}$ and $\iota=\varepsilon^2+\rho_B$.
The growing-mode bound in Proposition~\ref{prop:phase-one}, part (a) of
its proof, gives
\[
 G_{j,N_1}=qG_j^0+O(\iota L\sqrt{p_0}q).
\]
Proof (b) gives
$H_{j,N_1}=(1-h)^{N_1}H_{j,0}+O(\iota\sqrt{p_0}q)$,
and part (d) of the same proof gives
$\|Q_\star w_{j,N_1}\|\le C\sqrt{p_0}(1+\iota q)$.
Proposition~\ref{prop:robust-init} establishes the same estimates for the
more general initialization used in the main theorem. In both cases
$G_j^0\asymp\sqrt{p_0}$, $|H_{j,0}|\le C\sqrt{p_0}$,
$q\asymp\varepsilon^{-1}$, and $(1-h)^{N_1}\le q^{-1}$.
Consequently
\[
 G_{j,N_1}=qG_j^0(1+O(\iota L)),\quad
 \frac{|H_{j,N_1}|}{qG_j^0}\le C(\varepsilon^2+\iota),\quad
 \frac{\|Q_\star w_{j,N_1}\|}{qG_j^0}\le C(\varepsilon+\iota).
\]
Here $b_j=G_j+H_j$ and $s_j\theta_\star^\top w_j=G_j-H_j$.
Expanding $r_j=\sqrt{(G_j-H_j)^2+\|Q_\star w_j\|^2}$ thus yields
\[
 p_{j,N_1}=q^2(G_j^0)^2(1+e_j),\qquad
 |e_j|\le C(\iota L+\varepsilon^2)\le C\varepsilon.
\]
The final bound uses $L\ge1$, $\iota L\le2\varepsilon$ and sufficiently
small $\varepsilon_0$. Normalizing inside a class gives
$|\log((p_{j,N_1}/\sum_{i\in S_{s_j}}p_{i,N_1})/\lambda_j)|\le C\varepsilon$.

For Phase II, Lemma~\ref{lem:phase-two-mass} gives
\[
 p_{j,n+1}=p_{j,n}(A_{s_j,n}+e_{j,n}),\qquad
 A_{s,n}=1+\tau(1-P_{s,n}/\beta),\qquad |e_{j,n}|\le C\tau\zeta,
\]
where $\zeta=\Theta_\star^2+\rho_B+\tau$.
Put $v_{j,n}=p_{j,n}/\sum_{i\in S_{s_j}}p_{i,n}$ and
$\bar e_{s,n}=\sum_{i\in S_s}v_{i,n}e_{i,n}$.
Then
\[
 \frac{v_{j,n+1}}{v_{j,n}}
 =\frac{A_{s_j,n}+e_{j,n}}{A_{s_j,n}+\bar e_{s_j,n}}.
\]
The class-mass and small-step bounds ensure both numerator and denominator
are at least $1/2$. Since $\log$ is $2$-Lipschitz on $[1/2,\infty)$,
$|\log(v_{j,n+1}/v_{j,n})|\le C\tau\zeta$.
Summing at most $C_LL/\tau+1$ steps and using $\zeta L\le3\varepsilon$
proves the first bound of Theorem~\ref{thm:explicit-selection-main},
uniformly over the entire phase.
This accumulation is $O(\zeta L)$, not $O(L)$: the potentially large
class-common growth cancels exactly.
Appendix~\ref{app:sharp-fit} gives $P_{s,N_\star}=\beta(1+O(\varepsilon))$,
which proves the theorem's endpoint mass bound.

Because $\sum_{j\in S_s}\lambda_j=1$, the selected template has
$P_+=P_-=\beta$ and is a balanced aligned exact representation.
The initialization bounds imply $c/m\le\lambda_j\le C/m$;
the endpoint mass bound proves the asserted individual mass bounds.
For raw parameters, the balance estimate gives
\[
\kappa_\star:=\max_j
\frac{|a_{j,N_\star}^2-\|w_{j,N_\star}\|^2|}{p_{j,N_\star}}
\le C\varepsilon,
\]
and hence
\[
 |b_j-\sqrt{p_j}|+|r_j-\sqrt{p_j}|\le C\kappa_\star\sqrt{p_j}.
\]
For completeness, write $x=b_j/r_j>0$. Then
$|x-x^{-1}|=|b_j^2-r_j^2|/p_j\le\kappa_\star$,
so $x=1+O(\kappa_\star)$ and
$b_j/\sqrt{p_j}=\sqrt{x}$, $r_j/\sqrt{p_j}=x^{-1/2}$.
Combining with the endpoint mass bound shows
$|b_j-\sqrt{p_j^{\rm sel}}|+|r_j-\sqrt{p_j^{\rm sel}}|
\le C\varepsilon\sqrt{p_j}$.
Thus
\[
 |a_j-a_j^{\rm sel}|^2+\|w_j-w_j^{\rm sel}\|^2
 \le Cp_j\bigl(\varepsilon^2+\|u_j-s_j\theta_\star\|^2\bigr).
\]
All endpoint angles are acute and
$\|u_j-s_j\theta_\star\|^2=2(1-\cos\theta_j)\le2\sin^2\theta_j$.
The existing estimate $W_s^2\le C\varepsilon^2$ therefore gives
\[
 \sum_jp_j\|u_j-s_j\theta_\star\|^2
 \le2\sqrt m\sum_sP_sW_s^2\le C\beta\sqrt m\varepsilon^2.
\]
Since $\sum_jp_j=\sqrt m\Pi=O(\beta\sqrt m)$, summing the preceding
bounds proves
\[
\|\Psi_{N_\star}-\Psi^{\rm sel}\|^2
\le C\beta\sqrt m\,\varepsilon^2,
\]
which is the claimed relative-distance estimate.
\end{proof}

\section{Single-example SGD on a fixed dataset}\label{app:single-example}
This appendix proves Theorem~\ref{thm:main-sgd}. The phase analysis in
Appendices~\ref{app:technical}--\ref{app:selection-principle} applies to an
update whose gradient is uniformly close, relative to each neuron's size,
to the population gradient. A single-example gradient does not satisfy such
a bound. We therefore group consecutive iterates into short intervals and
show that the total update over each interval has the required accuracy.
These intervals are used only in the proof: the algorithm updates both
layers after every sampled example.

\subsection{Algorithm and statement}
Retain the Gaussian linear teacher and model normalization from the main
text:
\[
 f_\Psi(x)=m^{-1/2}\sum_j a_j\sigma(w_j^\top x),\quad
 y(x)=\beta\theta_\star^\top x,\quad
 p_0=c_0\beta\varepsilon^4/\sqrt m.
\]
Set
\[
 L=\log(1/\varepsilon),\quad
 D=d+\log\frac{16m}{\delta\varepsilon},\quad
 \omega=c_\omega\varepsilon/L.
\]
Here $\omega$ is the normalized duration of an analysis interval. It is
distinct from the half-step $h=\tau/2$ used in the phase analysis.
For an integer $K$ defined below, use the constant step
\[
 \tau=\omega/K,\qquad \eta=\frac{\sqrt m}{\beta}\tau.
\]
Define analysis indices and actual SGD iteration indices by
\[
 J_1=\left\lceil\frac{L}{\log(1+\omega/2)}\right\rceil,
 \qquad J_\star=J_1+\lceil C_LL/\omega\rceil,
 \qquad T_1=KJ_1,\quad T_\star=KJ_\star.
\]

We first state the fixed-dataset SGD theorem, that is, Theorem \ref{thm:main-sgd} in the main body. Its proof is given at the
end of this appendix, after the concentration and transfer estimates on
which it relies.
\begin{theorem}[Single-example SGD on one fixed dataset]\label{se:thm:reuse}
There are universal constants
$c_\omega,C_K,C_M,C,\varepsilon_0>0$ such that the following holds.
Let $0<\varepsilon\le\varepsilon_0$ and $0<\delta<1/2$, and suppose that
Assumption~\ref{assumption:initialization} holds. Draw once a dataset of
iid Gaussian examples of size
\[
 M=\left\lceil C_M(d+\log(16/\delta))
       \frac{L^2}{\varepsilon^2}\log^2(eL/\varepsilon)\right\rceil
\]
and set
\[
 K=\left\lceil C_K\frac{D^2L^2}{\varepsilon^2}\right\rceil,
 \qquad\tau=\omega/K.
\]
At every iteration draw a single index independently and uniformly from
$\{1,\ldots,M\}$, with replacement, and update both layers on that
example's square loss. With probability at least $1-\delta$ over the dataset
and sampled indices, all output signs remain unchanged and every neuron
remains nonzero through $T_\star$. Moreover,
\[
 \max_{T_1\le n\le T_\star}\max_j
 \angle(u_{j,n},s_j\theta_\star)\le C\varepsilon L,
 \qquad
 \|f_{\Psi_{T_\star}}-y\|_{L^2(\gamma_d)}\le C\beta\varepsilon.
\]

Define
\[
 G_j^0=\tfrac12\bigl(|a_j(0)|+s_j\theta_\star^\top w_j(0)\bigr),
 \qquad
 \lambda_j=\frac{(G_j^0)^2}{\sum_{i\in S_{s_j}}(G_i^0)^2}.
\]
Then, for every $j$ and every $T_1\le n\le T_\star$,
\[
 \left|\log
 \frac{p_{j,n}/\sum_{i\in S_{s_j}}p_{i,n}}{\lambda_j}\right|
 \le C\varepsilon.
\]
At $T_\star$, for $s\in\{-1,+1\}$,
\[
 |P_{s,T_\star}-\beta|\le C\beta\varepsilon,
 \qquad
 \left|\frac{p_{j,T_\star}}{\beta\sqrt m\,\lambda_j}-1\right|
 \le C\varepsilon.
\]
For the exact representation $\Psi^{\rm sel}$ defined by
\[
 a_j^{\rm sel}=s_j\sqrt{\beta\sqrt m\,\lambda_j},
 \qquad
 w_j^{\rm sel}=a_j^{\rm sel}\theta_\star,
\]
we also have
\[
 \frac{\|\Psi_{T_\star}-\Psi^{\rm sel}\|}{\|\Psi^{\rm sel}\|}
 \le C\varepsilon.
\]
Finally,
\[
 \E\|Q_\star\nabla f_{\Psi_{T_\star}}(X)\|^2
 \le C\beta^2\varepsilon^2,
 \qquad
 \sum_{i\ge2}\lambda_i(G_{f_{\Psi_{T_\star}}})
 \le C\beta^2\varepsilon^2.
\]
The parameters satisfy
\[
 \tau\asymp\frac{\varepsilon^3}{D^2L^3},\qquad
 T_\star=O(D^2L^4/\varepsilon^3).
\]
\end{theorem}

Theorem~\ref{se:thm:reuse} contains Theorem~\ref{thm:main-sgd} and the
fixed-dataset conclusions used in Theorem~\ref{thm:explicit-selection-main}
and Corollary~\ref{cor:gradient}. The integer $K$ determines the learning
rate and the deterministic reporting times; it is not an algorithmic batch
size. In particular, the dataset size has no training-horizon factor and no
explicit width dependence. 

\subsection{Normalized dynamics and local control}
Write $s_0=\sqrt{\beta/\sqrt m}$, $a_j=s_0A_j$, $w_j=s_0W_j$, and
$z=((A_j,W_j))_j$. Define
\[
 R_z(x)=m^{-1}\sum_i A_i\sigma(W_i^\top x)-\theta_\star^\top x,
\]
\[
 g_j(z;x)=\bigl(R_z(x)\sigma(W_j^\top x),\;
               R_z(x)A_j\1_{W_j^\top x>0}x\bigr),\qquad F(z)=\E g(z;X).
\]
Write $\mathcal D=(X_1,\ldots,X_M)$ for the fixed dataset. An actual SGD
step is exactly $z_{n+1}=z_n-\tau g(z_n;X_{I_n})$. Within a block, $X_k$
in the sums below denotes the selected example $X_{I_{n_0+k}}$, not a fresh draw.
Let $\mathsf M(z)=m^{-1}\sum_j|A_j|\|W_j\|$ and
\[
 \mathcal R_0=\{\mathsf M\le6,\ 1/2\le |A_j|/\|W_j\|\le2\ \forall j\}.
\]
This is the manuscript's working region in the normalized variables.
At the start of an interval, denote the state by $z^0\in\mathcal R_0$
and put $t_j=|A_j^0|+\|W_j^0\|$. Define
\[
 e(z,z^0)=\max_j\frac{|A_j-A_j^0|+\|W_j-W_j^0\|}{t_j}.
\]
On $e\le1/32$, the output signs are fixed, all raw factors are nonzero,
$\mathsf M(z)\le12$, and the layer ratios are between $1/4$ and $4$.
Also $t_j$ is comparable, by universal factors, to each initial layer norm.

\begin{lemma}[Neuron-relative population regularity]\label{se:lem:relative}
On this local region,
\[
 |F_j^a(z)|+\|F_j^w(z)\|\le C t_j,
\qquad
 |F_j^a(z)-F_j^a(z^0)|+\|F_j^w(z)-F_j^w(z^0)\|
 \le C t_j e(z,z^0).
\]
The constants do not depend on the smallest neuron norm, width, or dimension.
\end{lemma}
\begin{proof}
Gaussian Cauchy--Schwarz and $\|R_z\|_2\le1+\mathsf M(z)$ give the first bound.
For the second, use
\[
 \E[\sigma(W_i^\top X)\1_{u_j^\top X>0}X]
 =r_i\left(q(\psi_{ij})u_i+\frac{\sin\psi_{ij}}{2\pi}u_j\right),
 \quad r_i=\|W_i\|,\quad q(\psi)=\frac{\pi-\psi}{2\pi}.
\]
Angular distance is Lipschitz in either direction, relative to Euclidean
distance on the unit sphere, with constant $\pi/2$. Both $q$ and $\sin$
are Lipschitz on $[0,\pi]$. Therefore the expression in parentheses is
uniformly Lipschitz in $(u_i,u_j)$.
A relative parameter perturbation of size $e$ changes each unit direction
by at most $Ce$, each radius by at most $Ce r_i^0$, and each output
coefficient by at most $Ce|A_i^0|$.
Summing changes in the displayed feature formula is bounded by
$Ce\,m^{-1}\sum_i|A_i^0|r_i^0\le6Ce$.
The prefactor $A_j$ in $F_j^w$ changes by $Ce|A_j^0|$.
Its teacher contribution is $-A_j\theta_\star/2$, which does not depend
on the gate direction. These facts give the hidden-gradient bound.
For $F_j^a$, the analogous formula uses
$r_jr_i(\sin\psi+(\pi-\psi)\cos\psi)/(2\pi)$ and teacher term
$-\theta_\star^\top W_j/2$. The same calculation applies.
\end{proof}

Data reuse creates two errors: the empirical gradient differs from the
population gradient, and the sampled gradient differs from the empirical
gradient. A uniform covariance event controls the first for all parameter
states. Conditional martingale bounds control the second along the
adaptive trajectory.

\subsection{Probability bounds on a fixed dataset}

We first record the conditional Bernstein inequality used for the sampled
indices. It is stated in scalar form; a standard net argument will give the
corresponding vector estimate.

\begin{lemma}[Conditional Bernstein inequality]\label{se:lem:bernstein}
Suppose that the centered adapted increments $Y_k$ satisfy
\[
 \E[e^{sY_k}\mid\mathcal F_k]\le
 \exp\left(\frac{s^2v}{2(1-b|s|)}\right),\qquad |s|<1/b.
\]
Then, for every $q\ge1$ and $u>0$,
\[
 \Pr\left(\left|\sum_{k<q}Y_k\right|>
           C(\sqrt{qvu}+bu)\right)\le2e^{-u}.
\]
If $\|Y_k\|_{\psi_1\mid\mathcal F_k}\le C_0$, the hypothesis holds with
$v,b$ depending only on $C_0$. If instead $|Y_k|\le b_0$ and
$\E[Y_k^2\mid\mathcal F_k]\le v_0$, it holds with $v=Cv_0$ and $b=Cb_0$.
\end{lemma}
\begin{proof}
Multiplying the conditional moment generating functions and applying the
Chernoff bound with
$s\asymp\min\{\sqrt{u/(qv)},1/b\}$ proves the two tails. For a conditionally
sub-exponential increment, expand the exponential and use
$\E[|Y_k|^r\mid\mathcal F_k]\le r!C^r$. For a bounded increment, use
$\E[|Y_k|^r\mid\mathcal F_k]\le b_0^{r-2}v_0$ for $r\ge2$.
\end{proof}

We apply Lemma~\ref{se:lem:bernstein} to an output coordinate and to a
fixed $1/2$-net
of the unit sphere for each hidden-gradient block. The net has at most
$5^d$ elements and recovers vector norm up to a factor two.
A union bound over neurons, intervals, and prefixes adds only
$C[d+\log(16mKJ_\star/\delta)]$ to $u$.
For the parameter choices above, after fixing the universal constants and
reducing $\varepsilon_0$ if necessary,
\begin{equation}\label{se:eq:logs}
 d+\log(16mKJ_\star/\delta)\le C D,
 \qquad
 d+\log(16M/\delta)\le C D.
\end{equation}
These inequalities follow by substituting the polynomial expressions for
$K,M,J_\star$ and using $\log D\le D$ and $\log L\le L\le D$.

We next invoke the uniform covariance estimate that controls the difference
between empirical and population gradients throughout the working region.
\begin{lemma}[Uniform gated covariance; \citep{tinaz2026both}, Lemma 10]
\label{se:lem:tinaz-uniform}
There are universal constants $C,c>0$ such that, for $q\ge2$,
$0<\nu<1/2$, and iid $X_i\sim N(0,I_q)$, if
$M\ge Cq\nu^{-2}\log^2(1/\nu)$, then with probability at least
$1-3e^{-cq}$,
\[
 \sup_{u,v\in\mathbb S^{q-1}}
 \left\|\frac1M\sum_{i=1}^M X_iX_i^\top
 \1_{u^\top X_i\ge0}\1_{v^\top X_i\ge0}
 -\E[XX^\top\1_{u^\top X\ge0}\1_{v^\top X\ge0}]\right\|_{\rm op}
 \le\nu.
\]
\end{lemma}
\begin{proof}
This is Lemma 10 of \citep{tinaz2026both}; its proof is given in their
Appendix C.6.
\end{proof}

Choose $\nu=\varepsilon/(1024L)$ and apply the lemma after the
following dimension augmentation.
To obtain failure $\delta/4$ in the observed dimension $d$, append independent
Gaussian coordinates to dimension at least
$\max\{d,c^{-1}\log(12/\delta)\}$. Restrict the event to directions with
zero appended coordinates and the upper-left covariance block. The
restricted event depends only on the observed coordinates. The full
augmented event implies it, so its marginal probability is at least
$1-\delta/4$.

Let $\Sigma_M=M^{-1}\sum_iX_iX_i^\top$ and define
\[
 K_M^\ge(u,v)=M^{-1}\sum_iX_iX_i^\top
          \1_{u^\top X_i\ge0}\1_{v^\top X_i\ge0}.
\]
Let $K_M(u,v)$ use strict inequalities in both indicators, and let
$H_M(u)=M^{-1}\sum_i X_iX_i^\top\1_{u^\top X_i>0}$.
For one fixed direction $e$, almost surely
$\Sigma_M=K_M^\ge(e,e)+K_M^\ge(-e,-e)$, hence $\|\Sigma_M-I\|\le2\nu$.
For arbitrary, data-dependent directions, the exact identities
\[
 H_M(u)=\Sigma_M-K_M^\ge(-u,-u),
\]
\[
 K_M(u,v)=\Sigma_M-K_M^\ge(-u,-u)-K_M^\ge(-v,-v)+K_M^\ge(-u,-v)
\]
handle strict gates, including zero preactivations. Thus $H_M$ and $K_M$
have errors at most $3\nu$ and $5\nu$. The arbitrary-width gradient formula
\[
 F^w_{M,j}=A_j\left(m^{-1}\sum_i A_iK_M(u_j,u_i)W_i
                              -H_M(u_j)\theta_\star\right)
\]
and its scalar version
\[
 F^a_{M,j}=(W_j)^\top\left(m^{-1}\sum_i A_iK_M(u_j,u_i)W_i
                              -H_M(u_j)\theta_\star\right)
\]
give, for every state with $\mathsf M\le12$, after subtracting their population
counterparts and summing $m^{-1}\sum_i|A_i|\|W_i\|=\mathsf M$,
\begin{equation}\label{se:eq:empbias}
 |F^a_{M,j}-F^a_j|\le63\nu\|W_j\|,
 \qquad \|F^w_{M,j}-F^w_j\|\le63\nu|A_j|.
\end{equation}
The event is uniform in the parameters. We also use the Gaussian event
$\max_{i\le M}\|X_i\|^2\le C D$, whose failure probability is at most
$\delta/4$.

Fix an interval starting at $z^0\in\mathcal R_0$ and let
\[
 \kappa=\min\bigl(\{q\in\{1,\ldots,K\}:e(z_q,z^0)>1/32\}
                    \cup\{K+1\}\bigr).
\]
In the martingale sums below, multiply the increment at time $k$ by the
predictable indicator $\1_{\{k<\kappa\}}$ and freeze the stopped trajectory
after time $\kappa$. For an interval whose initial state is outside
$\mathcal R_0$, set all corresponding increments to zero. We suppress these
indicators in the notation. Conditional
on the dataset and the past sampled indices, the mean of $g(z;X_I)$ is
$F_M(z)$.
Since $\|\Sigma_M\|\le2$ and $\mathsf M\le12$, the empirical residual norm is bounded:
\[
 \left(M^{-1}\sum_i R_z(X_i)^2\right)^{1/2}
 \le\sqrt2(1+\mathsf M)\le13\sqrt2.
\]
Cauchy--Schwarz and $\|\Sigma_M\|\le2$ also give
$|F^a_{M,j}|+\|F^w_{M,j}\|\le Ct_j$ on the local region.
The normalized scalar gradient components used in the net argument are
bounded by $CD$ and have conditional variance at most $CD$.
Indeed their square is at most $CD R_z(X_I)^2$, and the last display
bounds its empirical expectation. Centering changes only universal constants.
Lemma~\ref{se:lem:bernstein}, followed by the net and union bounds described
above, gives simultaneously for all neurons, intervals, and prefixes
$q\le K$,
\begin{equation}\label{se:eq:reusedmart}
 \left|\sum_{k<q}(g_j^a-F_{M,j}^a)/t_j\right|
 +\left\|\sum_{k<q}(g_j^w-F_{M,j}^w)/t_j\right\|
 \le C(D\sqrt K+D^2),\qquad q\le K.
\end{equation}
Indeed, the variance contribution is
$\sqrt{K\cdot D\cdot(CD)}=O(D\sqrt K)$, and the linear Bernstein term is
$O(D^2)$. A union bound gives \eqref{se:eq:reusedmart} with failure
probability at most $\delta/2$. The conditional mean is $F_M(z)$ because
the indices are sampled independently with replacement.

\subsection{Closing each interval and transferring the phase proof}
\begin{lemma}[Interval estimates]\label{se:lem:interval}
On the events just constructed, every interval starting in $\mathcal R_0$
satisfies the following estimates. Denote its endpoint by $z^+$ and write
\[
 \overline g=K^{-1}\sum_{k<K}g(z_k;X_k),\qquad z^+=z^0-\omega\overline g.
\]
Then, with universal constants,
\begin{align}
 \max_{q\le K}e(z_q,z^0)&\le C\omega,\label{se:eq:local}\\
 |\overline g_j^a-F_j^a(z^0)|
 &\le \|W_j^0\|\,\rho,\nonumber\\
 \|\overline g_j^w-F_j^w(z^0)\|
 &\le |A_j^0|\,\rho,\label{se:eq:blockerror}\\
 |\Delta_j^+-\Delta_j^0|&\le C_b\omega^2p_j^0.\label{se:eq:blockbalance}
\end{align}
Here $\Delta_j=A_j^2-\|W_j\|^2$, $p_j^0=|A_j^0|\|W_j^0\|$, and
\[
 \rho\le C\left(\omega+D/\sqrt K+D^2/K\right)+72\nu
 \le \frac{\varepsilon}{2L}.
\]
The constants $c_\omega$ and $C_K$ can be chosen so that the last inequality
holds and $C_b$ is no larger than the balance-drift constant used in
Appendix~\ref{app:technical}.
\end{lemma}

\begin{proof}
Before and including a possible local exiting update, the accumulated
conditional-mean displacement is at most $CK\tau t_j=C\omega t_j$.
The stopped noise sum in \eqref{se:eq:reusedmart}, multiplied by $\tau$,
is also at most $C\omega t_j$ because $K\ge C_KD^2L^2/\varepsilon^2$.
For sufficiently small $c_\omega$ and $\varepsilon_0$, a first exiting
iterate would therefore satisfy $e\le C\omega<1/32$, a contradiction.
This proves \eqref{se:eq:local}; the stopped process includes the first
candidate exiting update.

For the effective gradient, divide the terminal martingale sum by $K$.
Lemma~\ref{se:lem:relative} and \eqref{se:eq:local} bound the difference between
the average population mean and $F(z^0)$ by $C\omega t_j$.
The uniform empirical bias \eqref{se:eq:empbias} contributes at most
$72\nu$ times the relevant starting layer norm, because
\eqref{se:eq:local} keeps the layer magnitudes within a factor $9/8$ of
their initial values. Converting $t_j$ to either starting layer norm costs
only a universal factor.
This proves \eqref{se:eq:blockerror} and the displayed bounds on $\rho$.
Choose $c_\omega$ to make the $C\omega$ contribution at most
$\varepsilon/(8L)$; choose $C_K$ so the sampling contribution is at most
$\varepsilon/(4L)$. The fixed choice $\nu=\varepsilon/(1024L)$ gives
$72\nu<\varepsilon/(8L)$.

Because the gradients within an analysis interval are evaluated at different parameter states, their average need not satisfy the samplewise balance identity. We therefore telescope the exact balance identity over the individual SGD updates.
\[
 \Delta_j^+-\Delta_j^0
 =\tau^2\sum_{k<K}\bigl[(g_j^a(z_k;X_k))^2
                              -\|g_j^w(z_k;X_k)\|^2\bigr].
\]
On the dataset event $\max_{i\le M}\|X_i\|^2\le CD$, local layer
comparability gives the pointwise estimate
$(g_j^a)^2+\|g_j^w\|^2\le CD^2p_j^0$. It follows that the balance defect is
at most
$C\omega^2(D^2/K)p_j^0$ after summing over the interval. Since
$D^2/K\le\varepsilon^2/(C_KL^2)$, increasing $C_K$ places this bound within
the balance-drift budget of Appendix~\ref{app:technical}. This proves
\eqref{se:eq:blockbalance}.
\end{proof}

\begin{proposition}[Transfer to the phase analysis]\label{se:prop:transfer}
Let $\widetilde\Psi_l=\Psi_{Kl}$ and regard $\omega$ as the normalized step
of this endpoint sequence. On the event of Lemma~\ref{se:lem:interval},
the endpoint sequence satisfies the deterministic update bounds stated
after Lemma~\ref{lem:balance-propagation}, with normalized step $\omega$
and relative gradient error at most
$\varepsilon/(2L)$. Consequently, the conclusions of
Proposition~\ref{prop:phase-one}, Proposition~\ref{prop:robust-init},
Proposition~\ref{prop:angular-bootstrap}, and
Proposition~\ref{prop:sharp-fit} hold at the corresponding endpoint indices
$J_1$ and $J_\star$. The deterministic share and parameter-distance
estimates derived in Appendix~\ref{app:selection-principle} hold at the same
endpoints.
\end{proposition}
\begin{proof}
Undoing the normalization in \eqref{se:eq:blockerror} gives the
per-neuron relative gradient bound used in the Phase-I growing, decaying,
and transverse recursions. Those recursions use only the decomposition of
the effective update into the population gradient and a bounded error; they
do not require the effective update to be unbiased.

Equation~\eqref{se:eq:blockbalance} gives the one-step balance-drift bound
with $\omega$ in place of $\tau$. In the balanced case it is telescoped from
zero, and under Assumption~\ref{assumption:initialization} it is telescoped
from the initial defect as in Proposition~\ref{prop:robust-init}. The
Phase-II radial expansion uses the population radial velocities, the
relative error bound, layer comparability, and the normalization remainder.
The transverse recursion uses the population Gaussian feature formula and
the current balance estimate. Thus the logistic and angular recursions, and
the first-exit argument built from them, apply without changing their
hypotheses.

In raw coordinates the effective step is $K\eta=\sqrt m\,\omega/\beta$;
the scaling $a_j=s_0A_j$, $w_j=s_0W_j$ converts
\eqref{se:eq:blockerror} into (4.1) with $\rho_B$ replaced by $\rho$.
Likewise, multiplying \eqref{se:eq:blockbalance} by $s_0^2$ gives the
raw balance bound $C_b\omega^2p_j$. Thus both required bounds hold
with the same normalization. The deterministic allowance is
$\varepsilon/L$, since $c_1=1$ in Appendix~\ref{app:technical}.
Lemma~\ref{se:lem:interval} gives $\rho\le\varepsilon/(2L)$, leaving room
for the fixed constants in the normalization estimates. Finally,
Proposition~\ref{prop:sharp-fit} and the selection argument use only the
endpoint recursions just obtained. This proves the stated transfer.
\end{proof}

\begin{proof}[Proof of Theorem~\ref{se:thm:reuse}]\label{proof thm1}
The uniform covariance event and the input-norm event fail with probability
at most $\delta/2$ in total. Conditional on these dataset events, the union
bound leading to \eqref{se:eq:reusedmart} fails with probability at most
$\delta/2$. We henceforth work on their intersection.

Apply Proposition~\ref{se:prop:transfer} by induction over the intervals.
At an endpoint already known to lie in $\mathcal R_0$,
Lemma~\ref{se:lem:interval} controls the entire next interval and its
endpoint. The deterministic first-exit argument then keeps the next endpoint
in $\mathcal R_0$. Starting from the admissible initialization, the induction
closes Phase I through $J_1$ and Phase II through $J_\star$.

Within an interval, \eqref{se:eq:local} changes each layer magnitude by a
relative $O(\omega)$ and each unit direction by an angle $O(\omega)$.
It therefore preserves signs and nonvanishing, and changes each effective
mass and normalized within-class share by a relative $O(\omega)$. Since
$\omega=c_\omega\varepsilon/L$, the Phase-II conclusions extend from the
endpoints to every actual SGD iterate. The fitting-time conclusions hold
directly because $T_\star=KJ_\star$ is an endpoint.

Choose the constants in the following order: first retain the universal
constants from the deterministic phase analysis; next choose $c_\omega$
small enough for the local tube and step-size conditions; then choose
$C_M$ and $C_K$ large enough for the covariance, martingale, and balance
bounds; finally reduce $\varepsilon_0$ to satisfy the remaining smallness
conditions and \eqref{se:eq:logs}. These choices are independent of
$m,d,\beta,\varepsilon,$ and $\delta$.

Since $J_\star=O(L/\omega)=O(L^2/\varepsilon)$ and
$K=O(D^2L^2/\varepsilon^2)$,
\[
 T_\star=KJ_\star=O(D^2L^4/\varepsilon^3),
 \qquad
 \tau=\omega/K\asymp\varepsilon^3/(D^2L^3).
\]
\end{proof}

\section{Exact representations and local degeneracy}
\label{app:local-geometry}

This section proves the zero-set statements from Section~\ref{sec:zeroset}
and the local obstruction cited in Section~\ref{sec:discussion}.

Using \(\sigma(t)=(t+|t|)/2\), every nondegenerate parameter vector satisfies
\begin{equation}
\begin{aligned}
f_\Psi(x)&=\tfrac12\langle V_\Psi,x\rangle+\tfrac12H_\Psi(x),\\
V_\Psi&=\frac1{\sqrt m}\sum_j s_jp_ju_j,\\
H_\Psi(x)&=\frac1{\sqrt m}\sum_j s_jp_j|u_j^\top x|.
\end{aligned}
\label{eq:odd-even-decomposition}
\end{equation}
The first term is odd and the second is even, so they are orthogonal in
\(L^2(\gamma_d)\).

\begin{lemma}[Independence of absolute-value features]
\label{lem:absolute-value-independence}
Let \(\nu_1,\ldots,\nu_K\in\mathbb S^{d-1}\), \(d\ge2\), span pairwise
distinct lines. If
\(\sum_{k=1}^K c_k|\nu_k^\top x|=0\) almost everywhere, then
\(c_1=\cdots=c_K=0\).
\end{lemma}

\begin{proof}
The identity holds everywhere by continuity. The distributional Hessian of
\(|\nu^\top x|\) is a matrix-valued measure supported on \(\nu^\perp\):
\[
\bigl\langle\partial^2|\nu^\top\!\cdot|,\varphi\bigr\rangle
=2\nu\nu^\top\int_{\nu^\perp}\varphi\,d\mathcal H^{d-1}.
\]
For each \(k\), the hyperplane \(\nu_k^\perp\) contains a point outside
the union of all other such hyperplanes. Testing in a small neighborhood of
that point isolates \(c_k\), so \(c_k=0\). Repeating this for every \(k\)
proves the claim.
\end{proof}

\begin{proof}[Proof of Proposition~\ref{prop:zeroset}]
By \eqref{eq:odd-even-decomposition}, \(f_\Psi=y\) if and only if its odd
part equals the teacher and its even part vanishes. Equality of the odd
parts is exactly (Z1). Since
\(|u^\top x|=|(-u)^\top x|\), group the neurons by the line spanned by
\(u_j\). The coefficient of the resulting absolute-value feature on a line
\(\ell\) is
\[
\frac1{\sqrt m}\sum_{j:\,u_j\in\ell}s_jp_j.
\]
Lemma~\ref{lem:absolute-value-independence} shows that the even part vanishes
if and only if every such coefficient is zero, which is (Z2).
\end{proof}

\begin{proof}[Proof of Corollary~\ref{cor:width}]
For \(m=2\), condition (Z2) forces the two neurons to lie on the same line
with equal masses. They cannot have the same orientation, because then the
left side of (Z1) vanishes. Hence \(u_2=-u_1\), and (Z1) gives
\(u_1=\theta_\star\) and \(p_1=p_2=\sqrt2\,\beta\).

For any integer $m\ge4$, put
$\nu_\pm=\cos\gamma\,\theta_\star\pm\sin\gamma\,e$ and
$p=\beta\sqrt m/(2\cos\gamma)$, with $e\perp\theta_\star$ a unit vector.
Take a positive/negative antipodal pair on each line $\nu_\pm$, each member with mass $p$. Their combined output is
$p(\nu_++\nu_-)^\top x/\sqrt m=y(x)$.
If $m>4$, repeatedly split a neuron into two with the same sign and direction and positive masses summing to its former mass. This gives exactly $m$ active neurons without changing the function. Set $a_j=s_j\sqrt{p_j}$ and $w_j=\sqrt{p_j}u_j$. Every neuron is balanced, the maximum angle is $\gamma$, and the two lines are distinct since $0<\gamma<\pi/2$.
\end{proof}

Because an \(L^2(\gamma_d)\) equality between these continuous functions is
pointwise, every exact representation has zero residual on every sample.
Its mini-batch gradient is therefore identically zero, proving the fixed-point
claim in the main text.

\paragraph{Width three.}
Every occupied line must contain at least two active neurons by (Z2). Thus a nondegenerate width-three exact fit has only one hidden line, which (Z1) forces to be the teacher line. This does not force output-sign alignment: for any $q>0$, take
\[
(s_j,u_j,p_j)=(+1,\theta_\star,\beta\sqrt3),\quad
(+1,-\theta_\star,q),\quad
(-1,-\theta_\star,\beta\sqrt3+q).
\]
The output is exactly $y$, but the second neuron is at the wrong sign-selected pole. This distinguishes teacher-line recovery from the stronger neuronwise conclusion of Theorem~\ref{thm:main-sgd}.

Thus width three does not yet admit multiple exact hidden lines, but the
next proposition shows that cubic local degeneracy already occurs at
width three.

\begin{proposition}[Cubic degeneracy]
\label{prop:cubic-degeneracy}
For every integer $m\ge3$ and $d\ge2$, there is a balanced family
$\Psi^\flat(t)$ approaching a balanced aligned two-pole fit, with
$P_+=P_-=\beta$ and maximum angle $t$, such that
\[
\mathcal L(\Psi^\flat(t))
=
\beta^2\left(
\frac{t^3}{6\pi}
+\frac{t^4}{16}
+O(t^5)
\right),
\qquad
\frac{\|\nabla\mathcal L(\Psi^\flat(t))\|^2}
     {2\mathcal L(\Psi^\flat(t))}
\le
C\frac{\beta t}{\sqrt m}.
\]
Consequently, no neighborhood of $\Psi^\flat(0)$ admits a constant
$\mu>0$ such that
$\|\nabla\mathcal L(\Psi)\|^2\ge 2\mu\mathcal L(\Psi)$
throughout that neighborhood. Thus a uniform local
Polyak--\L{}ojasiewicz (PL) inequality fails for the population loss.
\end{proposition}

\begin{proof}
We first construct the family at width three and then obtain every
$m\ge3$ by scaling and splitting neurons.

\paragraph{Width three.}
Fix a unit vector $e\perp\theta_\star$ and write
$\nu(\alpha)=\cos\alpha\,\theta_\star+\sin\alpha\,e$.
Take
$(s_1,u_1,p_1)=(+1,\nu(t),\beta\sqrt3/2)$,
$(s_2,u_2,p_2)=(+1,\nu(-t),\beta\sqrt3/2)$, and
$(s_3,u_3,p_3)=(-1,-\theta_\star,\beta\sqrt3)$.
Balance every neuron by setting
$a_j=s_j\sqrt{p_j}$ and $w_j=\sqrt{p_j}\,u_j$.

Then
$P_+=(p_1+p_2)/\sqrt3=\beta$ and
$P_-=p_3/\sqrt3=\beta$, while the maximum angle to the
sign-selected teacher pole is $t$. At $t=0$ this is a balanced aligned
exact representation.

Write $z=\theta_\star^\top x$ and $r=e^\top x$. The residual is
\[
R_t(x)
=
\frac{\beta}{2}
\left\{
\sigma(\nu(t)^\top x)+\sigma(\nu(-t)^\top x)
\right\}
-\beta\sigma(z).
\]
Using the kernel $K$ from Lemma~\ref{lem:gaussian},
$\mathcal L=\frac{\beta^2}{2}
[\frac34+\frac12K(2t)-2K(t)]$, and hence
\[
\mathcal L
=
\beta^2\left(
\frac{t^3}{6\pi}
+\frac{t^4}{16}
+O(t^5)
\right).
\]

It remains to bound the gradient. Using
$\sigma(q)=(q+|q|)/2$ and
$(|a+b|+|a-b|)/2=\max\{|a|,|b|\}$, we can write
\[
R_t
=
\frac{\beta}{2}
\left[
(\cos t-1)z
+\max\{\cos t\,|z|,\sin t\,|r|\}
-|z|
\right].
\]
For $0<t\le1/10$, define
$W=\{|z|\le\tan t\,|r|\}$.
Polar coordinates in the $(\theta_\star,e)$-plane give
$\mathbb P(W)\le Ct$ and
$\mathbb E[\mathbf 1_W(v^\top X)^2]\le Ct$
for every unit vector $v$.

For directions outside the $(\theta_\star,e)$-plane, decompose $v$
into its planar and orthogonal components. The latter Gaussian
component is independent of the wedge event, so the same bound holds
with a constant independent of $d$.

On $W^c$,
$\max\{\cos t\,|z|,\sin t\,|r|\}=\cos t\,|z|$,
and therefore
$\|R_t\mathbf 1_{W^c}\|_2\le C\beta t^2$.
On $W$, for each of the three neuron directions,
$|\sigma(u_j^\top X)|\le Ct|r|$, so
$\|\sigma(u_j^\top\cdot)\mathbf 1_W\|_2\le Ct^{3/2}$.
Since the loss expansion gives
$\|R_t\|_2\asymp\beta t^{3/2}$, for every neuron $j$ and unit vector $v$,
\[
\left|
\mathbb E\!\left[
R_t\mathbf 1_{\{u_j^\top X>0\}}v^\top X
\right]
\right|
\le C\beta t^2,
\qquad
\left|
\mathbb E\!\left[
R_t\sigma(u_j^\top X)
\right]
\right|
\le C\beta t^2.
\]
Therefore
$\|\nabla\mathcal L(\Psi^\flat_3(t))\|^2
\le \frac{C\beta^2t^4}{3}
\sum_{j=1}^3(a_j^2+\|w_j\|^2)$.
Since
$\sum_{j=1}^3(a_j^2+\|w_j\|^2)
=2\sum_{j=1}^3p_j=4\beta\sqrt3$,
we obtain
$\|\nabla\mathcal L(\Psi^\flat_3(t))\|^2
\le C\beta^3t^4/\sqrt3$.

\paragraph{General width.}
Fix any integer $m\ge3$ and choose positive integers
$r_1,r_2,r_3$ with $r_1+r_2+r_3=m$.
Starting from the width-three construction above, put
$c=(m/3)^{1/4}$ and replace base neuron $j$ by $r_j$ identical copies
with raw parameters
$a_{j,\ell}^{(m)}=c\,a_j^{(3)}/\sqrt{r_j}$ and
$w_{j,\ell}^{(m)}=c\,w_j^{(3)}/\sqrt{r_j}$,
for $1\le\ell\le r_j$.

Each copy remains balanced. By positive homogeneity of ReLU,
\[
\frac1{\sqrt m}
\sum_{\ell=1}^{r_j}
a_{j,\ell}^{(m)}
\sigma\!\left((w_{j,\ell}^{(m)})^\top x\right)
=
\frac1{\sqrt3}
a_j^{(3)}
\sigma\!\left((w_j^{(3)})^\top x\right).
\]
Thus the width-$m$ network represents exactly the same function as the
width-three network. In particular, its loss is unchanged, its class
masses remain $P_+=P_-=\beta$, and its maximum angle remains $t$.

The corresponding raw gradient blocks satisfy
$\nabla_{a_{j,\ell}^{(m)}}\mathcal L_m
=(3/m)^{1/4}\nabla_{a_j^{(3)}}\mathcal L_3/\sqrt{r_j}$
and
$\nabla_{w_{j,\ell}^{(m)}}\mathcal L_m
=(3/m)^{1/4}\nabla_{w_j^{(3)}}\mathcal L_3/\sqrt{r_j}$.
Summing their squared norms over all copies gives
$\|\nabla\mathcal L_m\|^2
=\sqrt{3/m}\,\|\nabla\mathcal L_3\|^2
\le C\beta^3t^4/\sqrt m$.

Since
$2\mathcal L_m
=\beta^2t^3(1+O(t))/(3\pi)$, we obtain
\[
\frac{\|\nabla\mathcal L_m\|^2}
     {2\mathcal L_m}
\le
C\frac{\beta t}{\sqrt m}.
\]
Letting $t\downarrow0$ rules out any positive state-independent PL
constant on a full neighborhood of the aligned fit.
\end{proof}

\section{Experimental Details}
\label{app:two-phase-experiment}

\paragraph{SGD first aligns then fits.}
We use $d=2$, $m=1000$, $\beta=1$, batch size $B=1$, and random
seed 17. Both layers are updated simultaneously on one example
sampled uniformly with replacement from a fixed dataset of
$M=2000$ Gaussian examples, drawn once before training.
There are $m/2$ neurons of each initial output-weight sign.
Writing $\alpha=10^{-2}$, we set
\[
p_0=\frac{\beta\alpha^2}{\sqrt m},
\qquad
|a_j(0)|=\|w_j(0)\|=\sqrt{p_0}.
\]
The initial directions $u_j(0)$ are sampled independently and
uniformly from the unit circle, conditional on
$s_j\langle u_j(0),\theta_\star\rangle\geq-1/2$,
as required by Assumption~\ref{assumption:initialization}. The two layers are initialized with
balanced magnitudes. For the directional-trajectory visualization, we display
30 randomly selected neurons satisfying
$|\langle u_j(0),\theta_\star\rangle|\leq1/2$,
so that the plotted neurons start away from both teacher poles
and their subsequent alignment is visually apparent.
In the theorem's parameterization,
$\alpha=\sqrt{c_0}\,\varepsilon^2$.

The numerical parameters illustrate the mechanism; we do not claim
that they satisfy the theorem's unspecified smallness constants.
The dimensionless step size is
$\tau=\eta\beta/\sqrt m=0.003$, equivalently
$\eta=0.003\sqrt m/\beta$.
We run 12,000 SGD steps and record the trajectory every 20 steps.
No weight decay, clipping, normalization, or explicit rebalancing
is used. The alignment time is the first recorded iterate at which
every neuron is within $0.05$ radians of its sign-selected teacher
pole. The population loss falls below $10^{-8}$.

The reported loss is the exact population loss under the Gaussian
input distribution, rather than an empirical estimate on a test set.
At each recorded iterate, we evaluate
\[
L(\Psi)=\frac12\mathbb E\left[
\left(
\frac1{\sqrt m}\sum_{j=1}^{m}a_j\sigma(w_j^\top X)
-\beta\theta_\star^\top X
\right)^2
\right]
\]
in closed form. Expanding the square leaves only pairwise ReLU
expectations, ReLU--teacher cross terms, and the teacher second
moment. For nonzero $w_i,w_j$, writing
$\varphi_{ij}=\angle(w_i,w_j)$, we use
\[
\mathbb E[\sigma(w_i^\top X)\sigma(w_j^\top X)]
=
\frac{\|w_i\|\|w_j\|}{2\pi}
\left(
\sin\varphi_{ij}
+(\pi-\varphi_{ij})\cos\varphi_{ij}
\right),
\]
together with
\[
\mathbb E[\sigma(w_j^\top X)\theta_\star^\top X]
=\frac12w_j^\top\theta_\star,
\qquad
\mathbb E[(\theta_\star^\top X)^2]=1.
\]

\begin{figure}[!htbp]
    \centering
    \includegraphics[width=\linewidth]{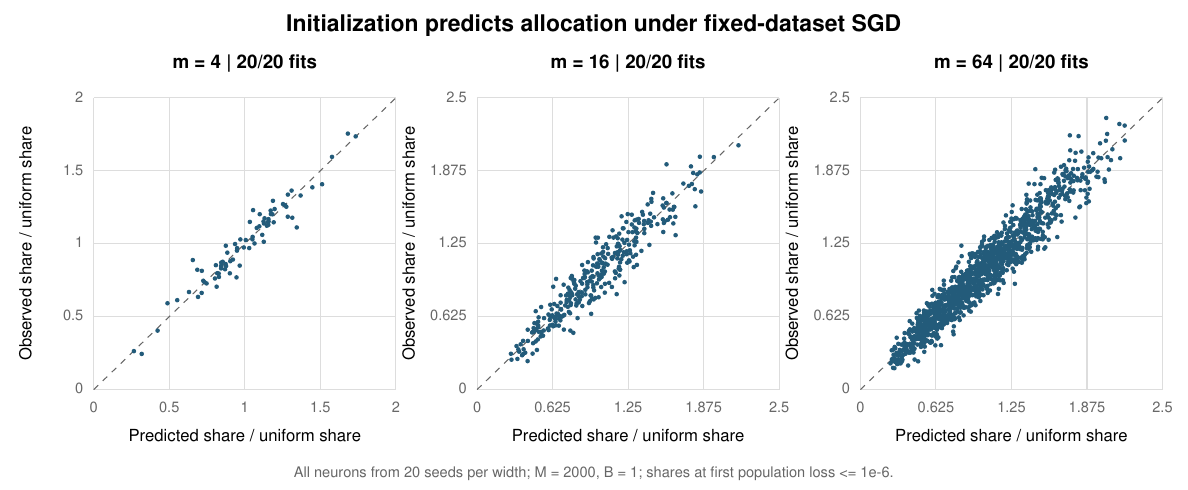}
    \caption{\textbf{Initialization predicts each neuron's share.}
    Share predicted before training against share observed at fitting,
    both measured relative to uniform allocation. Each panel shows all
    neurons from $20$ independent runs at one width.
    The dashed diagonal corresponds to exact agreement between
    predicted and observed shares. Although all neurons start with
    equal shares, their fitting-time shares are nonuniform and closely
    match the allocation predicted before training.
    Training uses batch-size-one SGD sampled uniformly with replacement
    from one fixed dataset of $M=2000$ Gaussian examples per run.
    All $20$ runs at each width reach the fitting threshold.}
    \label{fig:mass-selection-experiment}
\end{figure}

\paragraph{Neuronwise-allocation experiment.}
For Figure~\ref{fig:mass-selection-experiment}, we use the same
values of $\alpha$, $\beta$, $B$, and $\tau$.
Each run uses one fixed dataset of $M=2000$ Gaussian examples,
sampled uniformly with replacement.
We set $d=20$, vary $m\in\{4,16,64\}$, and run 20 independent
seeds at each width. Every run is trained for at most 20,000 steps.
The initial magnitudes and sign-class sizes are as above;
directions are sampled uniformly on the sphere conditional on
$|\langle u_j(0),\theta_\star\rangle|\leq1/2$.
For each neuron, define its initial growing mode by
\[
G_j^0
=
\frac12\left(
|a_j(0)|+s_j\theta_\star^\top w_j(0)
\right),
\qquad
s_j=\operatorname{sign}(a_j(0)).
\]
At the first monitored fitting time, we record
\[
\lambda_j
=
\frac{(G_j^0)^2}
{\sum_{i\in S_{s_j}}(G_i^0)^2},
\qquad
v_{j,n_{\mathrm{fit}}}
=
\frac{p_{j,n_{\mathrm{fit}}}}
{\sum_{i\in S_{s_j}}p_{i,n_{\mathrm{fit}}}}.
\]
Here $n_{\mathrm{fit}}$ denotes the first monitored iterate at which
the exact population loss satisfies
$L(\Psi_{n_{\mathrm{fit}}})\leq10^{-6}$;
the loss is evaluated every 20 SGD steps.
The plotted coordinates are
$(m\lambda_j/2,mv_{j,n_{\mathrm{fit}}}/2)$, so a value of one
corresponds to uniform allocation within a sign class.
A point on the diagonal $y=x$ corresponds to
$v_{j,n_{\mathrm{fit}}}=\lambda_j$, meaning that the neuron's
observed share of its sign-class mass at fitting exactly matches
the share predicted from its initialization. Thus, closeness to
the diagonal measures the accuracy of the initialization-dependent
allocation prediction.

\section{Additional Experiments}
\label{app:additional-experiments}

\paragraph{Protocol.}
Unless stated otherwise, we use the allocation initialization of
Appendix~\ref{app:two-phase-experiment}, $d=20$, $m=16$,
$\alpha=0.01$, $\beta=1$, $\tau=0.003$, and batch-size-one SGD
on one fixed dataset of $M=2000$ examples, with 20 independent
seeds per condition. The allocation error $E_{\rm pool}$ is the
median of $|\log(v_j/\lambda_j)|$ over all neurons of all fitted
runs, measured at the first monitored iterate with population
loss at most $10^{-6}$. As a reference, predicting uniform shares
within each sign class gives an error of approximately
$0.18$--$0.30$ in the baseline setting. Angles are measured
relative to the initial sign-selected teacher poles unless stated
otherwise. The experiments below that leave our assumptions are
empirical; they are not covered by Theorems~\ref{thm:main-sgd} and~\ref{thm:explicit-selection-main}.

\paragraph{Online sampling.}
With a fresh Gaussian example at every update, all 20 runs at each
width $m\in\{4,16,64\}$ again align and then fit. The allocation
errors are $0.0474$, $0.0589$, and $0.0732$, compared with
$0.0383$, $0.0853$, and $0.0858$ on a fixed dataset. Reusing
$M=2000$ examples therefore changes the picture little.

\paragraph{Initialization scale and learning rate.}
We sweep $\alpha\in\{0.001,0.003,0.01,0.03,0.1,0.3,1\}$ with a
budget of 30,000 updates. For $\alpha\leq0.03$ every run fits,
and alignment precedes or coincides with fitting in all runs
except one at $\alpha=0.03$. The median maximum angle at fitting
grows with the scale: $0.003$, $0.005$, $0.014$, and $0.037$
radians. For $\alpha\geq0.1$, no run reaches the fitting threshold
within the budget; the median maximum angles at the final
checkpoint are $0.11$, $0.29$, and $0.76$ radians. These results
illustrate the importance of small initialization for alignment
before fitting.

\begin{figure}[!htbp]
\centering
\includegraphics[width=\linewidth]{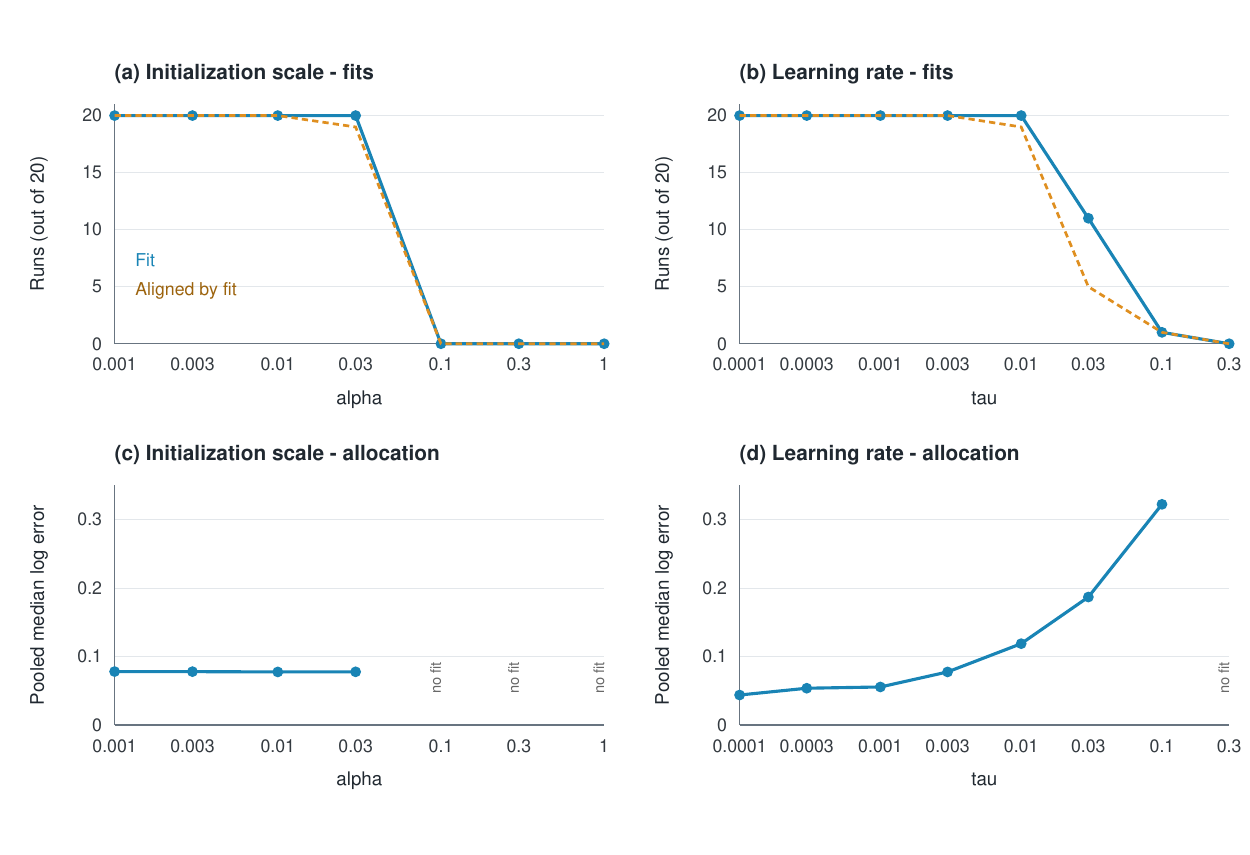}
\caption{Initialization-scale and learning-rate sweeps for
fixed-dataset SGD ($M=2000$, $m=16$, $B=1$, 20 seeds per setting).
Top: counts of runs that fit and align by fitting.
Bottom: pooled median absolute log-share error over fitted runs
only; missing points indicate no fit. The scale sweep uses
30,000 updates and $\tau=0.003$; the learning-rate sweep uses
$90/\tau$ updates and $\alpha=0.01$. All runs diverge at $\tau=0.3$.}
\label{fig:h-scale-rate}
\end{figure}

The learning-rate sweep covers
$\tau\in\{10^{-4},3\times10^{-4},10^{-3},0.003,0.01,0.03,0.1,0.3\}$.
A smaller learning rate needs proportionally more updates to make
the same progress, so each run uses $90/\tau$ updates, well beyond
what it needs to fit. All runs fit for
$\tau\leq0.01$; at larger rates SGD becomes unstable, with 11/20
runs fitting at $\tau=0.03$, 1/20 at $0.1$, and all diverging at
$0.3$. As the learning rate decreases, the allocation prediction
becomes more accurate: the error falls from $0.119$ at $\tau=0.01$
to $0.044$ at $\tau=10^{-4}$.
Figure~\ref{fig:h-scale-rate} summarizes both sweeps.

\paragraph{Dataset size.}
At $\tau=3\times10^{-4}$, we vary
$M\in\{500,2000,8000,32000\}$, using at most
$\lfloor80/\tau\rfloor$ updates and monitoring every 200 updates.
All 20 runs fit in each condition. The allocation errors are
$0.0857$, $0.0459$, $0.0269$, and $0.0237$, approaching the online
value $0.0189$. The prediction therefore becomes more accurate
as the learning rate decreases and the dataset grows, consistent
with the small-step, large-sample regime of Theorems~\ref{thm:main-sgd} and~\ref{thm:explicit-selection-main}.
The online comparison and dataset-size sweep are in Table~\ref{tab:h-sampling}.

\begin{table}[!htbp]
\centering
\small
\setlength{\tabcolsep}{4pt}
\begin{tabular}{rlrrrrr}
\toprule
$m$ & Sampling & $M$ & $\tau$ & Fits
    & $E_{\rm pool}$ & Log corr. \\
\midrule
4  & Online & --   & 0.003 & 20/20 & 0.0474 & 0.9746 \\
4  & Fixed  & 2000 & 0.003 & 20/20 & 0.0383 & 0.9668 \\
16 & Online & --   & 0.003 & 20/20 & 0.0589 & 0.9718 \\
16 & Fixed  & 2000 & 0.003 & 20/20 & 0.0853 & 0.9488 \\
64 & Online & --   & 0.003 & 20/20 & 0.0732 & 0.9665 \\
64 & Fixed  & 2000 & 0.003 & 20/20 & 0.0858 & 0.9555 \\
\midrule
16 & Online & --    & 0.0003 & 20/20 & 0.0189 & 0.9970 \\
16 & Fixed  & 500   & 0.0003 & 20/20 & 0.0857 & 0.9485 \\
16 & Fixed  & 2000  & 0.0003 & 20/20 & 0.0459 & 0.9831 \\
16 & Fixed  & 8000  & 0.0003 & 20/20 & 0.0269 & 0.9931 \\
16 & Fixed  & 32000 & 0.0003 & 20/20 & 0.0237 & 0.9955 \\
\bottomrule
\end{tabular}
\caption{Online and fixed-dataset SGD ($B=1$, 20 seeds per row).
Allocation is evaluated at the first monitored loss at most
$10^{-6}$. $E_{\rm pool}$ pools absolute log-share errors over
neurons and fitted runs. The lower block is the separate
dataset-size sweep at a smaller learning rate.}
\label{tab:h-sampling}
\end{table}

\paragraph{Separating the initialization assumptions.}
Assumption~\ref{assumption:initialization} combines a small scale,
an angular condition $s_j\langle u_j(0),\theta_\star\rangle\geq-1/2$,
and comparable but not necessarily equal layer magnitudes.
We examine these ingredients separately and include combined
Gaussian and Xavier variants ($M=8192$, 15,000 updates).
Unlike the allocation initialization in
Appendix~\ref{app:two-phase-experiment}, this sweep's baseline
directions are conditioned on
$s_j\langle u_j(0),\theta_\star\rangle\geq-1/2$.
The allocation error here is the median across runs of the
within-run mean of $|\log(v_j/\lambda_j)|$; uniform shares give
$0.36$ for the baseline.

\begin{itemize}
\item \textit{Scale.}
Between $0.25\times$ and $5\times$ the baseline scale, every run
fits and the allocation error stays at $0.08$--$0.10$. At
$20\times$ and $100\times$, no run fits within the budget and the
maximum angle at its end is $0.21$ and $0.80$ radians. Under Xavier
normal initialization, no run fits and the neurons do not align
(maximum angle $1.83$ radians), consistent with
\citet{tinaz2026both}.

\item \textit{Angular condition.}
Dropping the condition changes nothing ($0.0905$). Even when $25\%$
or all neurons start with selected-pole cosine in $[-0.9,-0.6]$,
every run aligns and fits. The allocation error rises to $0.13$ and
$0.19$, still far below the uniform reference ($0.93$ and $0.60$).

\item \textit{Layer magnitudes.}
Independent magnitudes in $[\sqrt{p_0},1.25\sqrt{p_0}]$ leave the
error at $0.0866$. We also test wide layer ratios, i.i.d.\ Gaussian
output weights without sign balancing, and small Gaussian weights
in both layers. All runs fit. In these conditions a neuron that
starts with a very small output weight can change sign: typically
one neuron per run, carrying under $0.3\%$ of the mass. The mass-weighted angle remains small
($0.006$--$0.007$ radians). For neurons retaining their initial
sign at fitting, the allocation error is $0.14$--$0.18$,
compared with $1.0$--$1.5$ for uniform shares.
\end{itemize}

Scale has the clearest effect on reaching the fitting threshold
in this sweep. All tested angular and magnitude variants fit,
although their neuronwise alignment and allocation accuracy differ.
Table~\ref{tab:h-initialization} lists all conditions.

\begin{table}[!htbp]
\centering
\small
\setlength{\tabcolsep}{4pt}
\begin{tabular}{lrrrr}
\toprule
Initialization & Fits & \shortstack{Max. angle\\(rad)}
    & \shortstack{Allocation\\error} & Uniform \\
\midrule
Baseline
    & 20/20 & 0.013 & 0.0902 & 0.36 \\
Scale $0.25\times$
    & 20/20 & 0.004 & 0.0906 & 0.32 \\
Scale $0.5\times$
    & 20/20 & 0.007 & 0.1016 & 0.35 \\
Scale $2\times$
    & 20/20 & 0.024 & 0.0855 & 0.35 \\
Scale $5\times$
    & 20/20 & 0.059 & 0.0812 & 0.33 \\
Scale $20\times$
    & 0/20 & 0.21$^{*}$ & -- & -- \\
Scale $100\times$
    & 0/20 & 0.80$^{*}$ & -- & -- \\
Xavier normal
    & 0/20 & 1.83$^{*}$ & -- & -- \\
\midrule
Unrestricted directions
    & 20/20 & 0.014 & 0.0905 & 0.34 \\
25\% violating angle condition
    & 20/20 & 0.025 & 0.1319 & 0.93 \\
100\% violating angle condition
    & 20/20 & 0.010 & 0.1888 & 0.60 \\
\midrule
Bounded independent magnitudes
    & 20/20 & 0.016 & 0.0866 & 0.36 \\
Wide layer ratios
    & 20/20 & 0.19$^{\dagger}$ & 0.135$^{\dagger}$
    & 0.99$^{\dagger}$ \\
i.i.d.\ Gaussian $a$
    & 20/20 & 0.12$^{\dagger}$ & 0.177$^{\dagger}$
    & 1.45$^{\dagger}$ \\
i.i.d.\ Gaussian $a$, unrestricted directions
    & 20/20 & 0.18$^{\dagger}$ & 0.165$^{\dagger}$
    & 1.28$^{\dagger}$ \\
Small Gaussian $a,w$
    & 20/20 & 0.18$^{\dagger}$ & 0.140$^{\dagger}$
    & 1.39$^{\dagger}$ \\
\bottomrule
\end{tabular}
\caption{Initialization ablations ($M=8192$, 20 seeds per condition,
15,000 updates, monitored every 100). Angles are medians across runs
of the maximum neuron angle at fitting; $^{*}$at the end of the budget.
Xavier normal: $w_j\sim N(0,2I_d/(d+m))$, $a_j/\sqrt m\sim N(0,2/(m+1))$.
Wide layer ratios: $|a_j(0)|=\sqrt{p_0}e^{z_j}$,
$\|w_j(0)\|=\sqrt{p_0}e^{-z_j}$, $z_j\sim N(0,0.7^2)$.
Small Gaussian: $a_j=\sqrt{p_0}g_j$, $w_j=\sqrt{p_0}z_j/\sqrt d$.
$^{\dagger}$Angles use the current output sign; allocation errors
include only neurons that keep their sign, with shares renormalized
within each class. Dashes indicate no fit.}
\label{tab:h-initialization}
\end{table}

\paragraph{Non-Gaussian isotropic inputs.}
We test independent Laplace coordinates with scale $1/\sqrt2$
and uniform coordinates on $[-\sqrt3,\sqrt3]$, both with covariance
$I_d$, and versions of each product distribution composed with a
fixed random rotation. For non-Gaussian inputs, loss is estimated
on an independent test set of 20,000 examples. Monitoring is every
20 updates for $m=16$ and every 40 for $m=64$.
For $m\in\{16,64\}$ and 15,000 updates, all 20 runs per condition
align and fit. Allocation errors range from $0.0668$ to $0.0934$,
compared with $0.0842$ and $0.0820$ for Gaussian inputs
(Table~\ref{tab:h-nongaussian}).

This is expected for the early teacher term. For centrally symmetric
inputs with finite second moment $\Sigma=\mathbb E[XX^\top]$ and
$\mathbb P(w^\top X=0)=0$,
\[
\mathbb E\!\left[
(\theta_\star^\top X)\mathbf 1_{\{w^\top X>0\}}X
\right]
=\frac12\Sigma\theta_\star.
\]
For $\Sigma=I_d$, this is the population teacher term that drives
early alignment; this term depends on the input law only through
its second moment. Gaussianity is used in the closed-form ReLU
kernel and angular-interaction estimates of Appendix~\ref{app:technical}, and the
gated-covariance concentration of Appendix~\ref{app:single-example}. A proof for more
general inputs would need suitable replacements for these bounds
and the corresponding concentration estimates.

\begin{table}[!htbp]
\centering
\small
\setlength{\tabcolsep}{4pt}
\begin{tabular}{rlrrrr}
\toprule
$m$ & Input law & Fits & Max. angle (rad)
    & $E_{\rm pool}$ & Uniform \\
\midrule
16 & Gaussian        & 20/20 & 0.0132 & 0.0842 & 0.294 \\
16 & Laplace         & 20/20 & 0.0156 & 0.0916 & 0.271 \\
16 & Uniform         & 20/20 & 0.0129 & 0.0668 & 0.275 \\
16 & Rotated Laplace & 20/20 & 0.0144 & 0.0840 & 0.298 \\
16 & Rotated uniform & 20/20 & 0.0133 & 0.0804 & 0.304 \\
64 & Gaussian        & 20/20 & 0.0171 & 0.0820 & 0.324 \\
64 & Laplace         & 20/20 & 0.0195 & 0.0888 & 0.310 \\
64 & Uniform         & 20/20 & 0.0159 & 0.0810 & 0.298 \\
64 & Rotated Laplace & 20/20 & 0.0178 & 0.0934 & 0.316 \\
64 & Rotated uniform & 20/20 & 0.0153 & 0.0830 & 0.315 \\
\bottomrule
\end{tabular}
\caption{Input-law comparison on fixed datasets ($M=2000$, $B=1$,
20 seeds per row). Angles are medians across runs; allocation errors
pool neurons and fitted runs, and ``Uniform'' is the same error for
uniform shares. For non-Gaussian inputs the loss is estimated on an
independent test set. Every run reaches the alignment threshold by
the fitting checkpoint.}
\label{tab:h-nongaussian}
\end{table}

\paragraph{Anisotropic Gaussian inputs.}
We use covariance matrices with eigenvalues logarithmically spaced
between 1 and $\kappa$, for
$\kappa\in\{1,3,10,30,100,300,1000\}$, rescaled so that
$\theta_\star^\top\Sigma\theta_\star=1$.
The unit teacher direction $\theta_\star$ is an eigenvector with
the largest eigenvalue (top), the smallest eigenvalue (bottom),
or the normalized sum of these two orthogonal eigenvectors (mixed).
The default budget is 40,000 updates.

By the identity above, the early teacher force points along
$\Sigma\theta_\star$, which is parallel to $\theta_\star$ only
when $\theta_\star$ is an eigenvector. With the top placement,
all runs align and fit up to $\kappa=1000$.
With the mixed placement, the directions initially move toward
$\pm\Sigma\theta_\star/\|\Sigma\theta_\star\|$ and later rotate
toward $\pm\theta_\star$. All runs fit within the default budget
for $\kappa\leq10$, and within 400,000 updates for $\kappa=30$ and
$100$. Replacing $\theta_\star$ by
$\Sigma\theta_\star/\|\Sigma\theta_\star\|$ in the initial growing
mode improves the allocation prediction: at $\kappa=10$ the error
drops from $0.1753$ to $0.0691$. This is consistent with the early
teacher force influencing the eventual shares.
The bottom placement is the hardest, since all other directions
have larger variance: for $\kappa\geq10$ no run fits within the
default budget, and with $\tau=0.003/\kappa$ and 400,000 updates
14/20 runs fit at $\kappa=10$ (Table~\ref{tab:h-anisotropic}).

\begin{table}[!htbp]
\centering
\small
\setlength{\tabcolsep}{4pt}
\begin{tabular}{lrrrrrr}
\toprule
Teacher & $\kappa$ & $\tau$ & Budget & Fits
    & $E_\theta$ & $E_{\Sigma\theta}$ \\
\midrule
Top & 1    & 0.003 & 40k & 20/20 & 0.0763 & 0.0763 \\
Top & 3    & 0.003 & 40k & 20/20 & 0.0671 & 0.0671 \\
Top & 10   & 0.003 & 40k & 20/20 & 0.0600 & 0.0600 \\
Top & 30   & 0.003 & 40k & 20/20 & 0.0562 & 0.0562 \\
Top & 100  & 0.003 & 40k & 20/20 & 0.0528 & 0.0528 \\
Top & 300  & 0.003 & 40k & 20/20 & 0.0470 & 0.0470 \\
Top & 1000 & 0.003 & 40k & 20/20 & 0.0452 & 0.0452 \\
Bottom & 3    & 0.003 & 40k & 20/20 & 0.0956 & 0.0956 \\
Bottom & 10   & 0.003 & 40k & 0/20 & -- & -- \\
Bottom & 30   & 0.003 & 40k & 0/20 & -- & -- \\
Bottom & 100  & 0.003 & 40k & 0/20 & -- & -- \\
Bottom & 300  & 0.003 & 40k & 0/20 & -- & -- \\
Bottom & 1000 & 0.003 & 40k & 0/20 & -- & -- \\
Mixed & 3    & 0.003 & 40k & 20/20 & 0.1467 & 0.0781 \\
Mixed & 10   & 0.003 & 40k & 20/20 & 0.1753 & 0.0691 \\
Mixed & 30   & 0.003 & 40k & 0/20 & -- & -- \\
Mixed & 100  & 0.003 & 40k & 0/20 & -- & -- \\
Mixed & 300  & 0.003 & 40k & 0/20 & -- & -- \\
Mixed & 1000 & 0.003 & 40k & 0/20 & -- & -- \\
\midrule
Bottom & 10  & 0.0003  & 400k  & 14/20 & 0.0864 & 0.0864 \\
Bottom & 30  & 0.0001  & 1200k & 0/20  & -- & -- \\
Bottom & 100 & 0.00003 & 4000k & 0/20  & -- & -- \\
Mixed & 30   & 0.003 & 400k & 20/20 & 0.1800 & 0.0653 \\
Mixed & 100  & 0.003 & 400k & 20/20 & 0.1917 & 0.0569 \\
Mixed & 1000 & 0.003 & 400k & 0/20 & -- & -- \\
\bottomrule
\end{tabular}
\caption{Anisotropic Gaussian inputs and budget/step-size controls
(20 seeds per row; $1\mathrm{k}=1000$ updates). Top and bottom place
the teacher along eigenvectors of the largest and smallest covariance
eigenvalues; mixed uses their normalized sum. $E_\theta$ and
$E_{\Sigma\theta}$ are pooled median absolute log-share errors at
fitting for the original and covariance-based predictions. The lower
block reports the controls. The two predictions coincide when
$\theta_\star$ is an eigenvector. Dashes indicate no fit.}
\label{tab:h-anisotropic}
\end{table}

\paragraph{Label noise.}
We train with labels $y_i=\theta_\star^\top x_i+\sigma\xi_i$,
$\sigma\in\{0,0.01,0.03,0.1,0.3,1\}$, where $\xi_i\sim N(0,1)$ is
drawn once and reused on every visit. Runs at different noise levels use the same inputs,
initialization, and order of SGD samples, so they differ
only in the labels. We train for 200,000 updates, far beyond the noiseless fitting
time (about 8,000), to observe the effect of fitting the noise.

Shortly after the noiseless fitting time, moderate noise leaves the
selected representation intact. At step 10,000 with $\sigma=0.1$,
the allocation error is $0.072$, the same as without noise, the
maximum angle is $0.03$ radians ($0.015$ without noise), and the
clean population loss is within a small factor of the
least-squares error on the same data. At the same timestep (10,000), larger noise affects the directions before
the allocation. At $\sigma=0.3$ the allocation error is still
$0.083$, while the maximum angle has grown to $0.09$ radians. At
$\sigma=1$, where the noise is as large as the signal, the
allocation error is $0.10$ and the maximum angle $0.38$ radians.

\begin{figure}[!htbp]
\centering
\includegraphics[width=\linewidth]{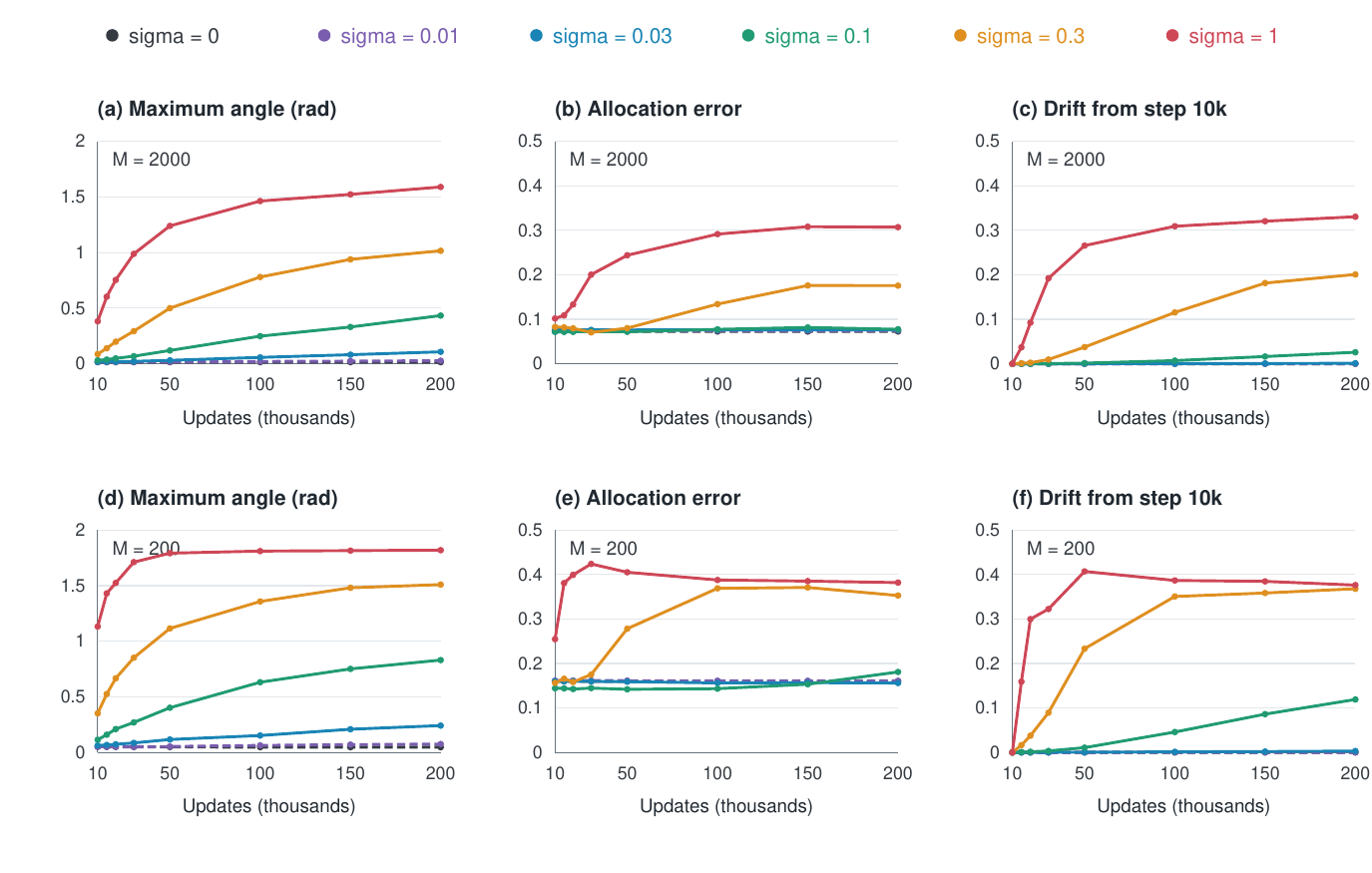}
\caption{Label noise and long training ($m=16$, $B=1$, 20 seeds
per condition; rows $M=2000$ and $M=200$). Curves are medians
across runs at common checkpoints. Angles are measured to the
initial sign-selected poles. Allocation error is the within-run
median of $|\log(v_j/\lambda_j)|$, and drift compares shares with
the same run at step 10,000. For moderate noise the shares stay
close to the prediction; noise fitting appears only after long
training, and earlier for the smaller dataset.}
\label{fig:h-noise}
\end{figure}

Much longer training eventually overfits the noise. At
$\sigma=0.1$, by step 200,000 the training loss falls below the
least-squares value on the same data. The directions are affected
more than the shares: the maximum angle grows to $0.43$ radians,
while the shares move by a median log ratio of only $0.026$.
For $\sigma\le0.03$ the shares stay at their predicted values over
the whole run, while for $\sigma\ge0.3$ both alignment and allocation degrade
substantially by step 200,000. With $M=200$, fewer examples than
the 336 parameters, overfitting starts earlier
(see Figure~\ref{fig:h-noise} for a summary).

\paragraph{Nonlinear odd teachers.}
Keeping the ReLU student, we replace the teacher by
$y(x)=\tanh(k\theta_\star^\top x)/k$, for $k\in\{1,3\}$ and widths
$m\in\{4,16,64\}$. Since the odd part of a bias-free two-layer ReLU network is
linear, the student cannot fit this teacher exactly.
We therefore evaluate at the end of training:
40,000 updates for $k=1$ and 80,000 for $k=3$.
The test loss, estimated on an independent Gaussian test set of
20,000 examples (10,000 for $m=64$), is $0.0143$--$0.0144$ and
$0.0092$--$0.0093$, close to the errors $0.0137$ and $0.0089$ of
the best linear approximation.

The prediction $\lambda_j$ of Theorem~\ref{thm:explicit-selection-main} is used unchanged.
Its error ranges from $0.0390$ to $0.0696$, compared with
$0.1690$ to $0.3008$ for uniform shares
(Table~\ref{tab:h-nonlinear}). For these odd single-index teachers
under Gaussian inputs,
$\mathbb E[y(X)\mathbf 1_{\{w^\top X>0\}}X]
=\frac12\mathbb E[y(X)X]$ for every $w\ne0$, and
$\mathbb E[y(X)X]$ is parallel to $\theta_\star$.
Thus the population teacher term has the same direction as for a
linear teacher. This suggests that the shares are decided early
in training for a broader class of targets than the one covered
by our theorems.

\begin{table}[!htbp]
\centering
\small
\setlength{\tabcolsep}{4pt}
\begin{tabular}{lrrrrrr}
\toprule
$k$ & $m$ & Step & Test loss & Angle (rad)
    & $E_{\rm pool}$ & Uniform \\
\midrule
1 & 4  & 40k & 0.01439 & 0.0751 & 0.0459 & 0.1690 \\
1 & 16 & 40k & 0.01431 & 0.0870 & 0.0696 & 0.2597 \\
1 & 64 & 40k & 0.01429 & 0.1060 & 0.0665 & 0.3008 \\
3 & 4  & 80k & 0.00922 & 0.1966 & 0.0390 & 0.1755 \\
3 & 16 & 80k & 0.00926 & 0.3186 & 0.0601 & 0.2657 \\
3 & 64 & 80k & 0.00916 & 0.4168 & 0.0577 & 0.2977 \\
\bottomrule
\end{tabular}
\caption{Nonlinear teachers $y(x)=\tanh(k\theta_\star^\top x)/k$
with a ReLU student, fixed $M=2000$ and $B=1$
(20 seeds per row). Loss and maximum angle are medians across
runs; allocation errors pool neurons across runs. All measurements
use the final checkpoint, and no run reaches loss $10^{-6}$.}
\label{tab:h-nonlinear}
\end{table}
\end{document}